\documentclass{article}
\usepackage[letterpaper,top=2cm,bottom=2cm,left=2cm,right=1.8cm,marginparwidth=1.75cm]{geometry}

\usepackage{amsthm}
\newtheoremstyle{boldremark}
  {\topsep}                    
  {\topsep}                    
  {\normalfont}                
  {}                           
  {\bfseries}                  
  {.}                          
  {.5em}                       
  {}                           
\theoremstyle{boldremark}
\newtheorem{rmk}{Remark}       

\usepackage[pdfencoding=auto, psdextra]{hyperref}
\usepackage{todonotes}

\usepackage{hyperref}
\usepackage{amssymb,amsmath,amsthm,bm}
\newtheorem{theorem}{Theorem}

\usepackage{amsthm} 
\newtheorem*{corollary}{Corollary} 

\usepackage{authblk}
\usepackage{xcolor}
\usepackage{pifont}
\newcommand{\cmark}{\textcolor{green}{\ding{51}}} 
\newcommand{\xmark}{\textcolor{red}{\ding{55}}}   

\newcommand{\ikt}[1]{\textcolor{magenta}{Comment from Irina: #1}}

\newcommand{\circled}[1]{%
    \tikz[baseline=(char.base)]{
        \node[shape=circle, draw, inner sep=1pt, minimum size=1.3em, line width=0.4pt] (char) {\small #1};}}
        
\usepackage{enumitem}
\usepackage{float}

\usepackage{amsthm}
\theoremstyle{remark}

\usepackage{graphicx}
\usepackage{subcaption}
\usepackage{float}
\usepackage{multirow}
\usepackage{rotating}
\usepackage{pdflscape}

\usepackage{algorithm}
\usepackage{algorithmic}
\usepackage{amsmath}

\usepackage{microtype}
\usepackage{comment}
\usepackage{soul}
\usepackage{todonotes}

\usepackage{hyperref}
\hypersetup{
    colorlinks=true,
    linkcolor=blue,
    citecolor=blue,
    urlcolor=blue
}

\title{Tackling Failure Modes of PINNs and PIKANs Using Conflict-Free Gradients}
\author[1]{Sidharth S. Menon}
\author[2]{Irina Tezaur}
\author[1]{Ameya D. Jagtap\thanks{Corresponding author: Ameya D. Jagtap (ajagtap@wpi.edu, ameyadjagtap@gmail.com)}}

\affil[1]{\textit{\small{Aerospace Engineering Department, Worcester Polytechnic Institute, Worcester, MA 01609, USA.}}}
\affil[2]{\textit{\small{Sandia National Laboratories, 7011 East Ave, Livermore, CA 94550, USA.}}}
\date{}
\begin{document}

\maketitle

\begin{abstract}
Scientific machine learning methods such as physics-informed neural networks (PINNs) increasingly rely on domain decomposition for better scalability while solving partial differential equations (PDEs) over complex geometries, yet the resulting composite loss comprising residual, boundary, and interface terms is highly susceptible to conflicting gradients that degrade training. This work bridges domain decomposition with projection-based gradient surgery to systematically mitigate such conflicts in 2D and 3D settings. We evaluate two existing projection-based algorithms, PCGrad and ConFIG, and identify their performance degradation in specific scenarios such as 3D domains with multiple overlapping interfaces. To address this limitation, we propose Norm-PCGrad, a normalized variant that achieves state-of-the-art accuracy across a range of 2D and 3D domain decomposition problems. Across the benchmarks considered, Norm-PCGrad consistently achieves the lowest relative $L_2$ error compared to training without gradient surgery as well as to existing algorithms such as PCGrad and ConFIG, while incurring negligible additional computational overhead. We further investigate gradient conflict mitigation through curvature-aware optimization by utilizing the quasi-Newton SSBroyden optimizer, either independently or in combination with Adam, and compare the resulting performance gains against Norm-PCGrad. To improve computational efficiency of domain decomposition frameworks such as Extended PINN (XPINN), we propose replacing vanilla PINNs in selected subdomains with separable architectures such as Separable PINN (SPINN), reducing the computational cost from quadratic (or cubic) to linear. We additionally demonstrate that gradient surgery extends to physics-informed Kolmogorov–Arnold Networks (PIKANs), yielding substantial accuracy improvements for 3D domain decomposition and confirming the generality of the proposed approach across network architectures. We also identify a previously unreported PINN failure mode in domain decomposition, where an incorrect choice of interface location leads to a physically inconsistent trivial solution, independent of the gradient surgery employed. The code is available at \href{https://github.com/ParamIntelligence/Conflict-Free-domain decomposition}{https://github.com/ParamIntelligence/Conflict-Free-domain decomposition}. A YouTube video showing the solution to the 2D Poisson test case is available at \href{https://www.youtube.com/watch?v=na3FuxI0XBk}{YouTube Video}.
\end{abstract}

\maketitle
\vspace{0.2cm}
\noindent
\begin{small}Keywords: \textit{Physics-Informed Neural Networks}, \textit{Gradient Conflicts}, \textit{Gradient Surgery}, \textit{Quasi-Newton Optimizers}, \textit{Domain Decomposition}, \textit{Kolmogorov Arnold Networks}.
\end{small}

\section{Introduction}
Solving partial differential equations (PDEs) over complex geometries is a fundamental challenge across science and engineering disciplines. While conventional numerical methods such as the finite element method (FEM) have long served as reliable tools for this purpose, they often require expensive mesh generation and can become computationally prohibitive for large-scale three-dimensional problems. Moreover, these conventional approaches can struggle with inverse and high-dimensional problems, where the computational cost and complexity can increase substantially. The emergence of physics-informed neural networks (PINNs) \cite{raissi2019physics} and their variants \cite{jagtap2022deep,jagtap2020locally,jagtap2020adaptive} has introduced a promising mesh-free alternative by embedding the governing physical laws directly into the neural network loss function. This framework not only avoids the need for explicit mesh generation but also offers a flexible approach for addressing ill-posed inverse and high-dimensional problems that are often challenging for conventional numerical methods; see, for example, \cite{abbasi2025history,jagtap2022physics,shukla2021physics,jagtap2022deepD,jagtap2023important,mhaskar2026approximation,hu2021extended,mao2020physics,abbasi2025challenges,menon2026scientific,menon2026intelligent,maity2026enpinn} for a broad range of applications of PINNs. However, PINNs face their own scalability challenges, particularly when applied to large or geometrically complex domains, motivating the adoption of domain decomposition strategies \cite{jagtap2020extended,cpinn_intro,penwarden2023unified,moseley2023finite,shukla2021parallel,hu2023augmented} that partition the problem into smaller, tractable subdomains. This decomposition, while beneficial for scalability, introduces additional loss terms to enforce solution continuity across subdomain interfaces, making the resulting composite loss function susceptible to conflicting gradients that can degrade training. Addressing these gradient conflicts in the context of domain decomposition, through both gradient surgery and curvature-aware optimization, forms the central focus of this work. While a formal mathematical definition is described later in this paper, a conflict-free gradient can be qualitatively defined as a gradient that has non-negative cosine similarity with all its counter-parts in the loss landscape.
\\\\
\noindent
\textcolor{blue}{\textbf{Gradient Surgery}}: In the realm of deep learning, it is well-known that multi-task learning is harder than single-task learning, although the full picture of this difficulty is not yet well understood. One widely studied cause of this difficulty has been attributed to the destructive interference between task gradients, commonly referred to as \textit{gradient conflict} in the literature. A simple, yet effective approach to overcome this conflict was proposed by Yu et al. \cite{yu2020gradient} in the context of multi-task learning. PCGrad was proposed in their work as a computationally efficient remedy that projects conflicting gradients based on their cosine similarity in the shared parameter space. An earlier work in this direction $-$ GradNorm \cite{chen2018gradnorm} utilized adaptive tuning of the gradient magnitudes to facilitate multi-task learning using deep neural networks. They ensured similar learning speed for individual tasks in a multi-task learning setup to mitigate the imbalance between tasks during model training. Another work, Impartial Multi-task Learning-Gradient (IMTL-G) \cite{liu2021towards}, achieved gradient balance and loss balance simultaneously, with the authors recognizing both as complementary to one another in the context of multi-task learning. Inspired by the aforementioned methods, a more recent work \cite{meng2024ri} utilized simultaneous rescaling and balancing of multi-task gradients to further improve upon the performance of PCGrad on segmentation and depth estimation tasks in computer vision.

With the advent of physics-informed machine learning \cite{raissi2019physics}, the gradient surgery techniques developed for multi-task learning were adopted for balancing the gradients of different loss terms involved while solving differential equations using neural networks. An earlier work \cite{zhou2023generic} in this direction utilized PCGrad to mitigate the gradient imbalance across loss terms while solving multi-state systems using PINNs, showing substantial improvement in accuracy that was otherwise hindered during model training. A more recent work by Liu et al. \cite{liu2025config} proposed ConFIG, inspired by existing algorithms like PCGrad, for mitigating gradient conflicts while solving PDEs using PINNs and demonstrated state-of-the-art performance for linear and non-linear PDEs. Furthermore, their experiments showed that both ConFIG and PCGrad surpassed the performance of methods like Learning Rate Annealing (LRA) \cite{wang2021understanding}, that adaptively mitigates any loss imbalance between different loss terms within the composite loss function, and others like IMTL-G that performs gradient surgery to ensure an equal learning speed across various loss terms before backpropagation.

Separately, Thanasutives et al. \cite{thanasutives2021adversarial} leveraged adversarial training strategies \cite{goodfellow2014generative} in combination with multi-task learning techniques to improve generalization of PINN-like architectures for handling a range of PDEs. Here, adversarial training was used to generate samples that maximize the loss, which in turn enabled the training procedure to focus on reducing these high-loss regions to attain better generalization, as opposed to merely overfitting to a specific type of PDE.
Fundamentally, gradient surgery \cite{yu2020gradient,liu2021towards,liu2025config} and dynamic weighting methods \cite{wang2021understanding} can be considered as two independent strategies used to improve training of physics-informed neural networks. More precisely, the dynamic weighting strategy is meant to tackle the magnitude imbalance of the loss terms, while gradient surgery tackles the directional conflicts between gradients of the loss terms in the loss landscape. Since dynamic weighting scales the loss terms to suppress the dominance of any one term over the others, gradient conflicts can still be present even after mitigating the magnitude imbalance, as was shown by previous works \cite{liu2025config}.
\\\\
\noindent
\textcolor{blue}{\textbf{Domain Decomposition}}: Domain decomposition is a well-developed strategy that is widely adopted to make solving large-scale numerical problems tractable using parallel computing hardware. More recently, this strategy has also been adopted for improving the computational efficiency of architectures like PINNs \cite{raissi2019physics}, which become prohibitively expensive to train for large-scale problems of practical relevance. To address these challenges, the conservative PINN (cPINN) \cite{cpinn_intro} introduced the first domain decomposition-based PINN framework specifically designed for conservation laws. Further, eXtended PINNs (XPINNs) \cite{jagtap2020extended} were proposed as an efficient improvement upon PINNs to solve the problem across decomposed subdomains, potentially leveraging multiple hardware resources to accelerate the training process. Other works \cite{snyder2023domain} in this direction leveraged the Schwarz alternating method to facilitate coupling of PINNs with conventional numerical methods like FEM via domain decomposition. A further direction was pursued by Pan et al. \cite{pan2024domain}, who utilized POD-generated local basis functions within each subdomain, thereby reducing the computational cost while preserving the essential dynamics at the global scale. While the initial works on domain decomposition \cite{jagtap2020extended} ensure continuity across subdomain interfaces by modeling this as an additional loss term in the composite loss function of PINNs, some recent works \cite{moseley2023finite} \cite{chung2026hard} have proposed to model the interface as a hard constraint, which ensures exact satisfaction of interface continuity and thereby avoids the need for an additional loss term. However, it is worth noting that hard constraints often require careful selection of the solution \textit{ansatz} prior to training and may not be applicable for large-scale 3D problems where choosing a suitable \textit{ansatz} becomes non-trivial or non-intuitive. Furthermore, Snyder et al. \cite{snyder2023domain} observed that strong enforcement of boundary conditions via a solution \textit{ansatz} does not always improve convergence and can even hinder training due to the sensitivity of the scaling functions to their hyperparameters, particularly for higher-dimensional or multi-subdomain settings.

Along a complementary direction, DADD-PINN \cite{xiong2026dadd} showcased orders of magnitude improvement in continuity across the interface through an adaptive sampling strategy combined with the utilization of overlapping interfacial domains across the different subdomains. However, their work was limited to 2D domains, and such overlapping strategies \cite{xiong2026dadd,ye2025spatial} can become intractable for 3D domains \cite{gu2024physics} encountered in practice. Other notable works in this direction include D3M \cite{li2019d3m} and DeepDDM \cite{li2020deep}, which utilized deep learning-based domain decomposition to solve elliptic PDEs, and Mosaic Flows \cite{wang2022mosaic}, which proposed a transferable framework for coupling pre-trained subdomain-local PINNs to solve PDEs on unseen domains. It is worth noting that domain decomposition in the context of PINNs can be leveraged in two distinct ways: (i) using domain decomposition to facilitate PINN training offline, where subdomain-local models are trained jointly via interface coupling \cite{jagtap2020extended, snyder2023domain, li2019d3m, li2020deep}, and (ii) coupling pre-trained subdomain-local PINNs online to solve PDEs on unseen or composite domains \cite{wang2022mosaic}. The current work follows the former approach, where domain decomposition is utilized to improve the training of PINNs by decomposing the problem into manageable subdomains that are solved simultaneously with interface conditions ensuring solution continuity.\\

\noindent
\textcolor{blue}{\textbf{Optimizers}}: While most previous discussions focused on gradient projection-based strategies to overcome conflicts between the individual loss terms within the composite loss function during model training, there has also been considerable emphasis on choosing a suitable optimizer to address such conflicts in the context of solving PDEs using PINN-based architectures.

One of the initial works along this line proposed a scale-invariant approach called MultiAdam \cite{yao2023multiadam} that segregates the loss terms (such as the PDE loss and boundary loss) into separate groups, each equipped with its own optimizer state. The gradients for each group are computed independently, and the resulting updates are aggregated before modifying the trainable parameters of the model. The proposed strategy consistently showcased superior performance, especially for cases where projection-based methods like PCGrad were shown to fail. In a related effort, Hwang et al. \cite{hwang2024dual} proposed adjusting the gradients such that the resulting update lies within the dual-cone region, where the inner product between the gradient vectors is known to be non-negative (or \textit{conflict-free}). Beyond first-order methods, some recent works \cite{kiyani2025optimizing} have advocated for quasi-Newton methods such as SSBroyden and SSBFGS for improving the convergence and accuracy of PINN-based architectures. While the SOAP \cite{vyas2025soap} (ShampoO \cite{gupta2018shampoo} with Adam in the Preconditioner's eigenbasis) optimizer has also been explored for mitigating gradient imbalances \cite{wang2026gradient} over first-order optimizers like Adam \cite{kingma2014adam}, recent empirical studies \cite{kiyani2025optimizing} have clearly shown self-scaled variants of quasi-Newton-based optimizers to outperform both SOAP and other first-order optimizers across a range of PDE benchmarks. Optimizers such as SOAP \cite{vyas2025soap} have been analyzed and demonstrated to overcome gradient conflicts through \textit{gradient whitening} which is essentially rescaling the gradients by local curvature information leading to dampening of ill-conditioned components in a multi-objective loss landscape.\\

\noindent
\textcolor{blue}{\textbf{Kolmogorov-Arnold Networks}}: Kolmogorov-Arnold Networks (KANs) \cite{liu2025kan} are a notable recent development in the scientific machine learning community that motivates the design of interpretable architectures as opposed to conventional black-box models based on multi-layer perceptrons (MLPs). Although the fundamental building blocks of this architecture were proposed in the 20th century through the seminal contributions of Kolmogorov \cite{kolmogorov1957representations,kolmogorov1961representation} and Arnold \cite{arnol1960representation,arnold2009representation}, it was only recently that researchers recognized their potential when combined with modern advancements in backpropagation to formulate KANs as an independent architecture. Although the original architecture \cite{liu2025kan} is based on spline basis functions, it has been found to be computationally expensive, and many recent works have proposed alternate formulations such as Chebyshev polynomials \cite{ss2024chebyshev} and radial basis functions \cite{abueidda2025deepokan}, which have been shown to be computationally more efficient. Despite the known numerical instability of Chebyshev polynomials at higher orders, the Chebyshev basis has been identified as a suitable candidate especially for solving PDEs \cite{shukla2024comprehensive}, mainly owing to desirable traits such as orthogonality that enable these basis functions to accurately learn and facilitate internal representations of solutions to many different PDEs. Owing to their interpretability, KANs have also been extensively utilized for solving physical systems \cite{guo2025physics}, high-dimensional PDEs \cite{menon2025anant}, and operator learning \cite{zhang2026bubbleokan}. A more recent work proposing Feature-Enriched KAN \cite{menon2026fekan} (FEKAN) showcased state-of-the-art performance for different basis functions across a range of tasks that include function approximation, solving PDEs, and operator learning.

The primary work on domain decomposition using KANs was proposed by Howard et al. \cite{howard2024finite}, inspired by the Finite Basis PINN (FBPINN) \cite{moseley2023finite} architecture. In their work, they utilized the B-spline basis for their Finite Basis KAN (FBKAN) architecture, with smaller overlapping subdomains that were trained in parallel for multiscale problems of varying frequencies. Another recent work proposed a hybrid architecture $-$ HPKM-PINN \cite{huang2025modified} that utilizes both MLP and KAN simultaneously to capture the global low-frequency and local high-frequency features, respectively, of a given multiscale problem. In both of the aforementioned works \cite{howard2024finite,huang2025modified} on domain decomposition, overlapping subdomains are handled similarly to how they are handled in FBPINNs \cite{moseley2023finite}, and the solution \textit{ansatz} utilizes partition-of-unity basis functions to facilitate continuity across the multiple interfaces. However, it is worthwhile to note that these existing works are strictly limited to 2D domains, since choosing a suitable unity function for the \textit{ansatz} becomes non-trivial for 3D domains, and further investigation is needed to assess the capability of KAN-based domain decomposition to efficiently handle complex 3D geometries.

It is evident that while \textit{gradient surgery} is utilized in the context of PINNs to mitigate \textit{gradient conflicts} between different loss terms present in a composite loss function, the current literature does not provide any clarity on its utilization in scenarios such as domain decomposition \cite{jagtap2020extended}, where multiple subdomains are solved simultaneously. Unlike in a typical PINN \cite{raissi2019physics} problem where the total loss comprises a fixed number of individual loss terms, domain decomposition, on the other hand, involves handling additional loss terms to ensure continuity across the interfaces of the different subdomains. In such a setup, each subdomain needs to be solved simultaneously while minimizing the losses and ensuring continuity across all the interfaces. Consequently, domain decomposition is more prone to \textit{gradient conflicts} due to the interplay between the various loss terms within the composite loss function, and we address this extensively in the current work using gradient surgery. The contributions of this work are as follows:
\begin{itemize}
    \item We recognize gradient imbalance as a critical concern for domain decomposition methods. Accordingly, we investigate the efficacy of some of the existing projection-based methods, demonstrating numerically that these approaches produce \textit{conflict-free} gradients for domain decomposition. Inspired by PCGrad, we also propose a simple yet effective formulation called Norm-PCGrad to ensure \textit{conflict-free} gradients for 2D and 3D domain decomposition. We showcase the effectiveness of the proposed methods through rigorous empirical studies in the results section of this paper.
    \item We propose to assign separable architectures such as SPINN \cite{cho2023separable} to specific subdomains as opposed to naively using PINN \cite{raissi2019physics} in all scenarios. This serves as a computational enhancement for the existing domain decomposition methods by facilitating linear computational complexity for certain specific subdomains. For instance, if an arbitrary domain $\Omega$ can be decomposed into subdomains $\Omega_1$ and $\Omega_2$, and if $\Omega_1$ is Euclidean, then we propose to utilize a SPINN-PINN configuration (denoting the architecture assigned to each subdomain) as opposed to PINN-PINN for improved computational efficiency.
    \item We further tackle gradient conflicts in the context of domain decomposition through curvature-aware optimization using self-scaled variants \cite{oren1974self} of quasi-Newton optimizers (such as SSBroyden), either independently or in combination with first-order optimizers like Adam \cite{kingma2014adam}. Specifically, we study the effectiveness of SSBroyden both as a standalone remedy for gradient conflicts and in conjunction with gradient surgery.
    \item Finally, we extend all the proposed strategies to facilitate \textit{fast}, \textit{conflict-free} 2D and 3D domain decomposition for architectures based on Kolmogorov-Arnold Networks, supporting both non-overlapping interfaces and overlapping interfacial domains. In doing so, we overcome the limitations of existing works on domain decomposition using KANs \cite{howard2024finite} \cite{huang2025modified}, which were demonstrated only for simple 2D domains with overlapping interfaces and are challenging to extend to 3D domains in their current setup.
\end{itemize}

\section{Methodology}
PINNs incorporate physical laws into neural network training by embedding governing equations directly into the loss function. This physics-based regularization constrains the admissible solution space and enables accurate modeling even with limited or noisy data. One of the major concerns with the framework is its immense computational cost incurred while training a model. To this end, an effective alternative based on domain decomposition was proposed to tackle this inefficiency by decomposing a domain into multiple sub-domains and solving them as manageable chunks especially in a resource-constrained environment.

We consider a parametric PDE of the general form,
\begin{align}
\label{eq:pde}
\mathcal{N}(u;\beta) &= f(x),~~~x\in\Omega \subset \mathbb{R}^d,\\
\mathcal{B}_k(u) &= g_k(x),~~~x\in\Gamma_k \subset \partial\Omega.
\end{align}
Here, $k=1,2,\dots, n_b$, and $\mathcal{N}(\cdot)$ denotes a (possibly nonlinear) differential operator and $\mathcal{B}_k(\cdot)$ represents a boundary operator corresponding to Dirichlet, Neumann, or Robin conditions. Furthermore, $u$ is the solution approximated by a neural network surrogate, $\beta$ denotes the physical parameter(s) of the parametric PDE, and $f(x)$ is the forcing term. While the above formulation is generic, it is worthwhile to note that time, $t$, can be incorporated as one of the components of $x$ and initial conditions treated as a specific type of boundary condition for a given spatiotemporal domain.

To enable domain decomposition, the domain $\Omega$ is partitioned into $N_{sd}$ non-overlapping sub-domains, $\Omega = \bigcup_{q=1}^{N_{sd}} \Omega_q$, with shared interfaces $\Gamma_{qq'} = \partial\Omega_q \cap \partial\Omega_{q'}$ between adjacent sub-domains $\Omega_q$ and $\Omega_{q'}$. Within each sub-domain, the unknown solution $u(x)$ is approximated by a separate parametric model $u_{\theta_q}(x)$, where $\theta_q$ denotes its trainable parameters and $\theta = \{\theta_q\}_{q=1}^{N_{sd}}$ collects all parameters. For notational convenience, we define the $q^{th}$-subdomain PDE residual
\begin{align}
\label{eq:residual}
r_{\theta_q}(x) = \mathcal{N}\big(u_{\theta_q}(x);\beta\big) - f(x).
\end{align}
The models are trained jointly by minimizing the XPINN sub-domain loss function of the following form:
\begin{align}
\label{eqn:loss_function}
\mathcal{L}_q(\theta) = \lambda_{\rm PDE}\, \mathcal{L}_{\rm PDE}(\theta) + \lambda_{u}\, \mathcal{L}_{u}(\theta) + \lambda_{u_{\rm avg}}\, \mathcal{L}_{u_{\rm avg}}(\theta) + \lambda_{\mathcal{R}}\, \mathcal{L}_{\mathcal{R}}(\theta),
\end{align}
where the individual terms are defined as
\begin{align}
\label{eq:loss_pde}
\mathcal{L}_{\rm PDE}(\theta) &= \sum_{q=1}^{N_{sd}}
\frac{1}{N_r^{q}} \sum_{i=1}^{N_r^{q}}
\left| r_{\theta_q}(x_r^{q,i}) \right|^2 , \\
\label{eq:loss_u}
\mathcal{L}_{u}(\theta) &= \sum_{q=1}^{N_{sd}}
\frac{1}{N_u^{q}} \sum_{i=1}^{N_u^{q}}
\left| u_{\theta_q}(x_u^{q,i}) - u^{q,i} \right|^2 , \\
\label{eq:loss_uavg}
\mathcal{L}_{u_{\rm avg}}(\theta) &= \sum_{q=1}^{N_{sd}}
\frac{1}{N_I^{q}} \sum_{i=1}^{N_I^{q}}
\left| u_{\theta_q}(x_I^{q,i}) - \big\{\!\big\{ u(x_I^{q,i}) \big\}\!\big\} \right|^2 , \\
\label{eq:loss_R}
\mathcal{L}_{\mathcal{R}}(\theta) &=
\sum_{q=1}^{N_{sd}}
\frac{1}{N_I^{q}} \sum_{i=1}^{N_I^{q}}
\left| r_{\theta_q}(x_I^{q,i}) - r_{\theta_{q^+}}(x_I^{q,i}) \right|^2 .
\end{align}
Here, $\{x_r^{q,i}\}_{i=1}^{N_r^{q}}$ are the residual (collocation) points in the interior of $\Omega_q$; $\{x_u^{q,i}\}_{i=1}^{N_u^{q}}$ are the supervised points with prescribed boundary/initial values or observed data $u^{q,i}$; and $\{x_I^{q,i}\}_{i=1}^{N_I^{q}}$ are the points on the interfaces of $\Omega_q$, with $q^+$ indexing the adjacent sub-domain across the interface. The interface average is defined as $\{\!\{ u \}\!\} = \tfrac{1}{2}\big( u_{\theta_q} + u_{\theta_{q^+}} \big)$.

As in conventional deep neural networks, the trainable parameters $\theta$ are optimized via backpropagation using automatic differentiation. Equation~\eqref{eqn:loss_function} predominantly comprises four components $-$ PDE residual loss ($\mathcal{L}_{\rm PDE}$), data loss ($\mathcal{L}_u$), average solution continuity loss ($\mathcal{L}_{u_{\rm avg}}$), and residual continuity loss ($\mathcal{L}_{\mathcal{R}}$). $\mathcal{L}_{\rm PDE}$ and $\mathcal{L}_u$ are the loss functions responsible for enforcing the PDE and the boundary and/or initial conditions, respectively, within each sub-domain. On the other hand, $\mathcal{L}_{u_{\rm avg}}$ and $\mathcal{L}_{\mathcal{R}}$ impose continuity across the interfaces shared between the sub-domains. Each of these loss components is prioritized based on the weights $-~\lambda_{\rm PDE},~ \lambda_{u},~ \lambda_{u_{\rm avg}}, \text{ and } \lambda_{\mathcal{R}}$, which are typically assigned prior to the model training. 

\begin{rmk}
Similar to PINNs, XPINNs also emphasize the importance of choosing optimal weights, which play a crucial role in the convergence of the loss function. It is worthwhile to note that suboptimal weighting of the loss terms can adversely affect the model training. To this end, dynamically adapting weights during training is a potential strategy that is prevalent in the literature. However, we note that the dynamic weighing strategy comes at an additional computational cost since the weights would be no longer fixed, but are instead trainable parameters. To this end, gradient surgery \cite{yu2020gradient} provides an alternate strategy based on the projection of the gradients for the individual loss terms. Gradient surgery not only ensures conflict-free gradients for the loss terms but also ensures minimization of all the loss terms simultaneously  within a composite loss function irrespective of the fixed weights chosen prior to model training. In our experiments we fix $\lambda_{\rm PDE}= \lambda_{\mathcal{R}} = 1$ and set the data ($\lambda_{u}$) and interface ($\lambda_{u_{\rm avg}}$) weights larger to emphasize continuity similar to the strategy adopted by XPINN \cite{jagtap2020extended}. We note that even if the chosen weights are suboptimal, the model still attains superior accuracy and convergence through gradient surgery which is discussed in our work. A previous work \cite{liu2025config} has also demonstrated performance gains of PCGrad and ConFIG (based on gradient surgery) over LRA \cite{wang2021lra} (based on dynamic weights) while solving a range of different PDEs. 
\end{rmk}


\noindent
\subsection{Respecting the Implicit Coupling at the Interface}
In the context of domain decomposition using methods like XPINN \cite{jagtap2020extended}, the interface is modeled using an interface loss function as shown in Equation (\ref{eqn:loss_function}). Minimizing this loss function is critical for attaining continuity of the solution across the interface spanning any two subdomains. In other words, the interface loss function can be considered as an implicit coupling between any two subdomains that must be respected for maintaining solution continuity.

As discussed earlier, conflicting gradients are prevalent in the context of PINNs since their training involves optimizing a composite loss function comprising individual loss terms for PDE residuals, boundary conditions, and initial conditions. Recent works have shown such conflicts to be detrimental for the training of PINNs.  These conflicts can become more pronounced with an increasing number of loss terms within the composite loss function. The existing methods for gradient surgery (such as PCGrad \cite{yu2020gradient}, ConFIG \cite{liu2025config}) are typically based on gradient projections, which involve recognizing conflicting gradients and projecting them such that their inner product remains non-negative, thereby rendering the gradients conflict-free.

Gradient conflicts are expected to be more pronounced in domain decomposition methods such as XPINN, which involve multiple loss terms across subdomains and interfaces. However, applying gradient surgery in this setting can alter the directions of interface gradients, potentially disrupting the implicit coupling between adjacent subdomains and introducing discontinuities at the interfaces. Thus, while gradient surgery can effectively mitigate conflicts in PINN-like architectures, it may be counterproductive for domain decomposition. In this work, we investigate the effectiveness of existing gradient surgery methods for XPINN and propose alternative strategies that mitigate gradient conflicts while preserving the coupling of interface gradients.

\subsection{Gradient Surgery}
\begin{figure}
\centering
{
\centering
\includegraphics[width=0.7\linewidth, trim={0mm 0mm 0mm 0mm}, clip]{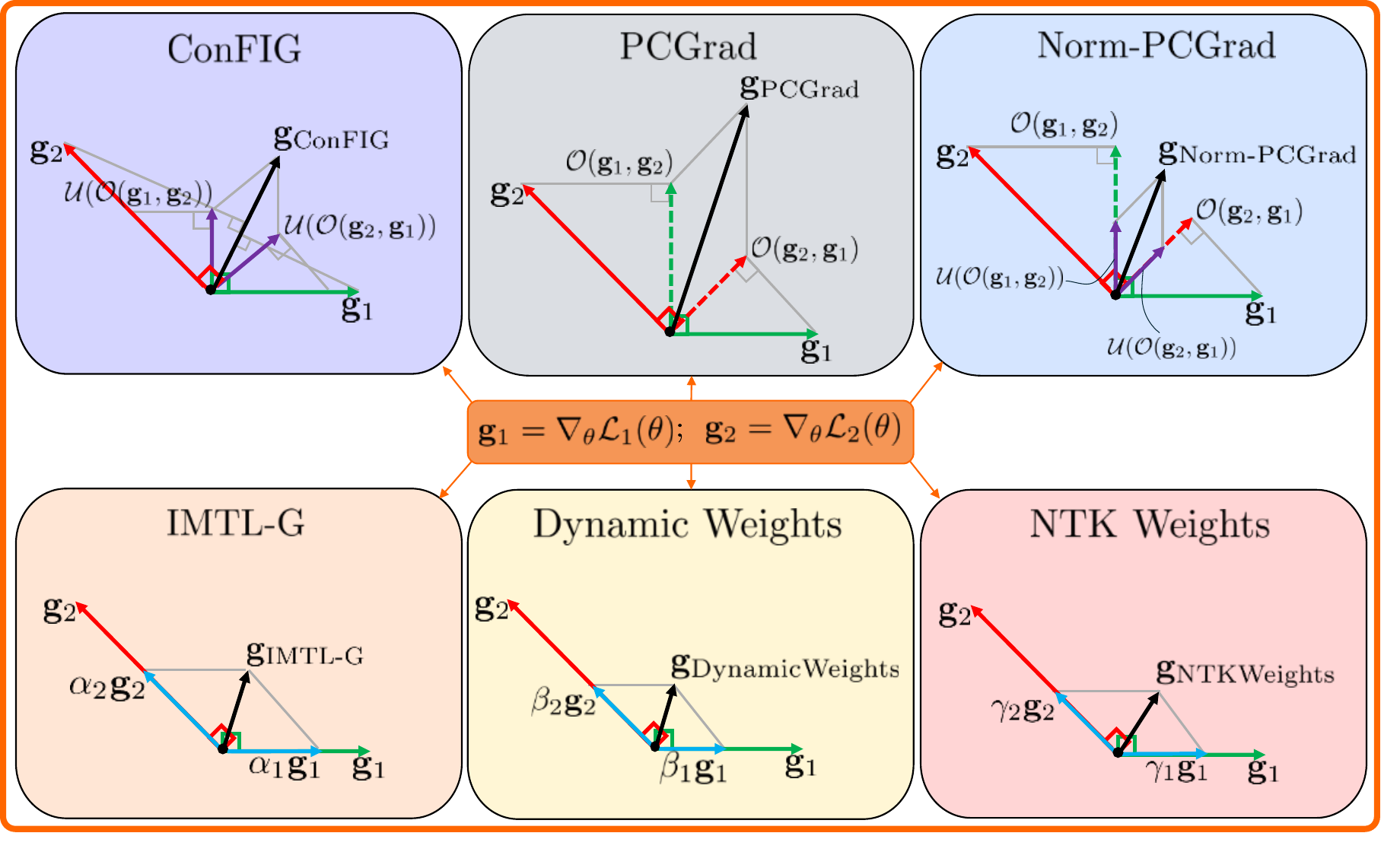}}
\caption{\textbf{Gradient surgery overview.} Comparison of gradient update strategies for two conflicting gradients $\mathbf{g}_1 = \nabla_\theta \mathcal{L}_1(\theta)$ and $\mathbf{g}_2 = \nabla_\theta \mathcal{L}_2(\theta)$. \textit{Top row (gradient surgery):} ConFIG normalizes each orthogonal projection before summing and rescales by the total projection length; PCGrad sums the orthogonal projections $\mathcal{O}(\mathbf{g}_1, \mathbf{g}_2)$ and $\mathcal{O}(\mathbf{g}_2, \mathbf{g}_1)$ at their original magnitudes; Norm-PCGrad normalizes each projection to unit length $\mathcal{U}(\cdot)$ before summing, attaining the same conflict-free direction as ConFIG but with a different magnitude rule. \textit{Bottom row (scaling-based methods):} IMTL-G rescales the original gradients using task-specific weights $\alpha_1, \alpha_2$ to ensure equal decrease across loss terms without explicit projections; dynamic weighting and NTK-based weights scale the gradients using heuristic-dependent factors ($\beta_1, \beta_2$ and $\gamma_1, \gamma_2$, respectively). Unlike gradient surgery, these scaling-based methods do not resolve gradient conflicts; instead, they scale the different components of gradients based on magnitude imbalance.  The scaling of the gradients shown is for illustrative purposes only.}
\label{fig:pcgrad}
\end{figure} 

The current work recognizes that choosing these weights and fixing them prior to model training is sub-optimal and instead explores \textit{gradient surgery} as an effective way of dealing with sub-optimal weights directly in the loss landscape through \textit{conflict-free} gradients. An alternate strategy called as Norm-PCGrad (as shown in Figure \ref{fig:pcgrad}) is proposed in the current work, inspired by the existing PCGrad \cite{yu2020gradient} algorithm that was originally proposed for multi-task learning. In addition to PCGrad, we also test the efficacy of another algorithm for gradient surgery, ConFIG \cite{liu2025config} for domain decomposition. The algorithm proposed in this work, $-$ Norm-PCGrad is described as Algorithm \ref{alg:normpcgrad}.

By inspection of Figure \ref{fig:pcgrad} with two loss terms, it is worthwhile to observe the similarities and differences between the algorithms of gradient surgery. While the algorithms share a similarity due to their shared objective of producing a conflict-free update direction, their differences mainly arise from the specific strategies they adopt to avoid conflicts between the gradients in the loss landscape. In PCGrad, typically a loss-specific gradient is projected onto the normal plane of others and vice versa, while algorithms like IMTL-G \cite{liu2021towards} rescale the loss-specific gradients (using $\alpha_1$ and $\alpha_2$ with reference to Figure \ref{fig:pcgrad}) to ensure an equal decrease of all loss terms without any explicit gradient projections. On the other hand, ConFIG attains a trade-off where it performs orthogonal projections similar to PCGrad while ensuring an equal decrease rate for all loss terms like IMTL-G. The proposed Norm-PCGrad algorithm is an extension of PCGrad where the gradients are orthogonally projected and the individual components are normalized before their accumulation, as shown in Algorithm \ref{alg:normpcgrad}. We demonstrate through an extensive numerical study that Norm-PCGrad attains conflict-free gradients while ensuring an equal decrease rate similar to ConFIG. However, it is worth noting the differences in their formulations that lead to differences in performance while solving PDEs through domain decomposition.

It is also worth noting that, the dynamic weighting methods, including LRA, GradNorm, and NTK-based approaches, rescale loss terms and their gradients without changing their directions. Thus, like IMTL-G, they cannot resolve gradient conflicts. Given the demonstrated advantage of gradient surgery over dynamic weighting for PDEs \cite{liu2025config}, we focus on gradient-surgery methods for obtaining conflict-free gradients in domain-decomposed PDE solvers.

\subsubsection{Conflict-Free Gradient Updates}
\label{sec:conflict_free}
The forthcoming discussion is with reference to Figure \ref{fig:pcgrad} which illustrates the conflict between only two gradients for the purpose of concise discussions. The composite objective in Equation~\eqref{eqn:loss_function} produces one gradient per term, $\mathbf{g}_i = \nabla_\theta \mathcal{L}_i(\theta)$. A plain summation $\sum_i \mathbf{g}_i$ can point along a direction that \emph{increases} an individual loss whenever the terms disagree: an update $\mathbf{g}_c$ conflicts with the decrease of $\mathcal{L}_i$ when $\mathbf{g}_i^\top \mathbf{g}_c < 0$. PCGrad \cite{yu2020gradient} and ConFIG \cite{liu2025config} both resolve such conflicts using the same two operators, which we state first for a consistent presentation. For two loss terms, we denote the corresponding gradients by $\mathbf{g}_1$ and $\mathbf{g}_2$. The normalization and orthogonality operators are defined in the literature as follows:
\begin{align}
\label{eq:operators}
\mathcal{U}(\mathbf{g}) = \frac{\mathbf{g}}{\|\mathbf{g}\| + \varepsilon},
\qquad
\mathcal{O}(\mathbf{g}_i, \mathbf{g}_j) = \mathbf{g}_j - \frac{\mathbf{g}_i^\top \mathbf{g}_j}{\|\mathbf{g}_i\|^2}\, \mathbf{g}_i,
\end{align}
where $\varepsilon$ is a small constant for numerical stability. The operator $\mathcal{O}(\mathbf{g}_i, \mathbf{g}_j)$ returns the component of $\mathbf{g}_j$ orthogonal to $\mathbf{g}_i$, and the cosine similarity between two gradients is written as
\begin{align}
\label{eq:similarity}
\mathcal{S}_c(\mathbf{g}_i, \mathbf{g}_j) = \frac{\mathbf{g}_i^\top \mathbf{g}_j}{(\|\mathbf{g}_i\|\,\|\mathbf{g}_j\| + \varepsilon)}.
\end{align}
\paragraph{PCGrad.}
When the two gradients conflict, i.e.\ $\mathbf{g}_1^\top \mathbf{g}_2 < 0$, PCGrad projects each gradient onto the plane orthogonal to the other and sums the two components,
\begin{align}
\label{eq:pcgrad}
\mathbf{g}_{\rm PCGrad} = \mathcal{O}(\mathbf{g}_1, \mathbf{g}_2) + \mathcal{O}(\mathbf{g}_2, \mathbf{g}_1),
\end{align}
leaving the gradients unchanged when they do not conflict. The resulting direction is conflict-free, but since the orthogonal components are added at their original magnitudes,
$\mathbf{g}_{\mathrm{PCGrad}}$ is biased toward the loss with the larger gradient and its projection lengths onto $\mathbf{g}_1$ and $\mathbf{g}_2$ are unequal.

\paragraph{ConFIG.}
ConFIG builds on the same two orthogonal components but rescales each to unit length before summing. This removes the magnitude bias and fixes the update \emph{direction} so that both losses receive an equal projection, and hence a uniform decrease rate,
\begin{align}
\label{eq:config_dir}
\mathbf{g}_v = \mathcal{U}\!\left[\,\mathcal{U}\big(\mathcal{O}(\mathbf{g}_1, \mathbf{g}_2)\big)
+ \mathcal{U}\big(\mathcal{O}(\mathbf{g}_2, \mathbf{g}_1)\big)\right].
\end{align}
The final update scales this direction by the sum of the projection lengths of both
loss-specific gradients onto it,
\begin{align}
\label{eq:config}
\mathbf{g}_{\mathrm{ConFIG}} = \mathcal{G}(\mathbf{g}_1, \mathbf{g}_2)
= \left(\mathbf{g}_1^\top \mathbf{g}_v + \mathbf{g}_2^\top \mathbf{g}_v\right)\mathbf{g}_v.
\end{align}
The magnitude in Equation \eqref{eq:config} adapts to the level of conflict: it increases when the loss-specific gradients are aligned (a favorable descent direction) and decreases when they oppose one another, which prevents the optimizer from stalling in a local minimum of a single term. 
\paragraph{Norm-PCGrad.}
It is interesting to note that for the two-loss case, PCGrad shares the same conflict-free direction as ConFIG and differs only in their magnitude rule \cite{liu2025config} and this is also depicted in Figure \ref{fig:pcgrad}. To this end, we introduce Norm-PCGrad through the current work which again shares similarities with PCGrad and ConFIG in terms of their resultant directions while differ only in the magnitude rule for the two-loss case as follows:
\begin{align}
\label{eq:normpcgrad}
\mathbf{g}_{\rm Norm-PCGrad} = \mathcal{U}(\mathcal{O}(\mathbf{g}_1, \mathbf{g}_2)) + \mathcal{U}(\mathcal{O}(\mathbf{g}_2, \mathbf{g}_1)).
\end{align}
\noindent
Unlike ConFIG, which shares similarities in its updates only for the two-loss case, Norm-PCGrad on the other hand attains a middle ground between PCGrad and IMTL-G for any number of loss terms. This is because Norm-PCGrad uses an update rule similar to PCGrad with an additional step of normalization that makes it also align with update rule of IMTL-G. The extension of PCGrad to Norm-PCGrad is described as Algorithm \ref{alg:normpcgrad}.

\paragraph{How does Norm-PCGrad minimize the presense of conflict-free gradients for domain decomposition?} As discussed in the previous section, domain decomposition especially using methods like XPINN involves an implicit coupling between the subdomains through an interface loss function, as well as their corresponding gradients, to ensure interface continuity. While the two subdomains share the same magnitude of the interface loss, their gradients can be entirely different and point towards different directions to minimize the same interface loss within their respective subdomains.
\noindent
To this end, we note that the magnitude of the gradients resulting from PCGrad is biased towards the direction of the gradient with the largest magnitude and this bias is localized to a given subdomain. It is also worth noting that the bias in the resulting gradients for one subdomain can be independent of the other, leading to the resulting interface gradients not respecting the coupling at the interface between the two subdomains after the gradient projections. This in turn leads to discontinuities at the interface in spite of using the state-of-the-art algorithms for mitigation of conflicting gradients. 

By explicitly normalizing the projected gradients, as shown in Equation \eqref{eq}, Norm-PCGrad reduces abrupt changes in the resultant gradient direction caused by differences in gradient magnitude. This normalization gives the different directions more balanced contributions while also accounting for the coupling between the subdomains through the interface gradients. We demonstrate their performance enhancement through detailed empirical studies for both 2D and 3D domain decompositions and showcase their difference in performance compared to the existing state-of-the-art algorithms like PCGrad and ConFIG. 

\paragraph{Extension to more than two losses.}
The objective in Equation \eqref{eqn:loss_function} contains four terms, so the two-term form of Equations \eqref{eq:pcgrad}--\eqref{eq:config} is a special case. For $m$ losses, ConFIG replaces the explicit orthogonal sum with the pseudoinverse form:
\begin{align}
\label{eq:config_m}
\mathbf{g}_v = \mathcal{U}\!\left(\big[\mathcal{U}(\mathbf{g}_1), \dots, \mathcal{U}(\mathbf{g}_m)\big]^{-\top}\mathbf{1}_m\right),
\qquad
\mathbf{g}_{\mathrm{ConFIG}} = \left(\sum_{i=1}^{m} \mathbf{g}_i^\top \mathbf{g}_v\right)\mathbf{g}_v,
\end{align}
where $[\cdot]^{-\top}$ is the pseudoinverse of the transposed gradient matrix and $\mathbf{1}_m$ is the all-ones vector; setting the direction weights to $\mathbf{1}_m$ recovers the uniform decrease rate of Equation \eqref{eq:config_dir}. PCGrad and the proposed Norm-PCGrad generalizes instead by looping over losses in random order and projecting out one conflict at a time, which no longer guarantee a conflict-free direction once more than two terms are involved. We adopt Equation \eqref{eq:config_m} for the four-term loss of Equation \eqref{eqn:loss_function}. It is also worthwhile to note that in the current work, we also note that the sampling of the gradients need not always be performed in the random order specifically in the context of solving PDEs.

In the context of PINNs, ConFIG \cite{liu2025config} and PCGrad have demonstrated state-of-the-art performance across a range of PDEs compared with adaptive loss-weighting methods, including linear rate annealing \cite{wang2021lra}, MinMax \cite{liu2021dual}, and ReLoBRaLo \cite{bischof2025multi}. These results motivate our investigation of projection-based methods for mitigating \textit{gradient conflicts} in domain-decomposed PDE solvers. 


\begin{algorithm}[ht]
\caption{\textcolor{blue}{Norm-PCGrad}/PCGrad}
\begin{minipage}{\linewidth}
\begin{algorithmic}[1]
\REQUIRE Model parameters $\theta$, task minibatch $\mathcal{B} = \{\mathcal{T}_k\}$
\STATE $\mathbf{g}_k \leftarrow \nabla_\theta \mathcal{L}_k(\theta) \ \forall k$
\STATE $\mathbf{g}_k^{\text{PC}} \leftarrow \mathbf{g}_k \ \forall k$
\FOR{$\mathcal{T}_i \in \mathcal{B}$}
    \FOR{$\mathcal{T}_j \overset{\text{uniformly}}{\sim} \mathcal{B} \setminus \mathcal{T}_i$ in random order}
        \IF{$\mathbf{g}_i^{\text{PC}} \cdot \mathbf{g}_j < 0$}
            \STATE // Subtract the  projection of $\mathbf{g}_i^{\text{PC}}$ onto $\mathbf{g}_j$
            \STATE Set $\mathbf{g}_i^{\text{PC}} = \mathbf{g}_i^{\text{PC}} - \dfrac{\mathbf{g}_i^{\text{PC}} \cdot \mathbf{g}_j}{\|\mathbf{g}_j\|^2}\mathbf{g}_j$
        \ENDIF
    \ENDFOR
\ENDFOR

\RETURN update $\Delta\theta = \mathbf{g}^{\text{PC}} = \sum_i \mathbf{g}_i^{\text{PC}}$  ~~~\textit{// For PCGrad}

\textcolor{blue}{\STATE Normalize the gradients before update: $\mathbf{g}^{\text{Norm-PC}}_i = \dfrac{\mathbf{g}_i^{\text{PC}}}{\left\|\mathbf{g}_i^{\text{PC}} \right\|}$ ~~~\textit{// For Norm-PCGrad}}
\textcolor{blue}{\RETURN update $\Delta\theta = \mathbf{g}^{\text{Norm-PC}} = \sum_i \mathbf{g}_i^{\text{Norm-PC}}$ ~~~\textit{// For Norm-PCGrad}}
\end{algorithmic}
\end{minipage}
\label{alg:normpcgrad}
\end{algorithm}

\subsection{Curvature-Aware Optimization}
The previous discussions on \textit{gradient surgery} dealt with the alignment of gradients corresponding to different loss terms within a composite loss function to mitigate the conflicting gradients in the loss landscape. However, it is worthwhile to note that gradient conflicts can also be resolved by appropriately choosing a suitable optimizer that can lead to conflict-free updates during the model training.

To this end, several studies on PDEs have demonstrated the benefits of curvature-aware optimizers, such as BFGS, over widely used first-order methods such as Adam. By leveraging Hessian information, curvature-aware methods can achieve faster convergence than optimizers based solely on the gradient. However, Newton-type methods incur prohibitive computational costs due to the evaluation and inversion of dense Hessians and can be unstable for ill-conditioned problems. Quasi-Newton methods provide a practical alternative by approximating the Hessian, reducing computational cost while retaining valuable curvature information.


Based on recent findings \cite{kiyani2025optimizing} \cite{wang2026gradient}, quasi-Newton optimizers in combination with Adam or independently have been shown to improve the convergence and accuracy of PINNs for a range of PDEs. The general formulation of the Broyden family of quasi-Newton methods and its derivation are provided in Appendix \ref{appx:ssbroyden}. It is worth noting that the performance of Broyden class optimizers can deteriorate for ill-conditioned problems. A potential alternative was proposed as self-scaled variants that overcome this limitation by scaling the Hessian approximation using a positive scalar $\tau_k$. The self-scaled Broyden (SSBroyden) optimizer is defined with the Hessian approximation of the following form,
\begin{equation}
\label{eqn:general_quasi_selfscaled}
    \mathrm{B}_{k+1} = \tau_k \biggl[\mathrm{B}_{k} - \frac{\mathrm{B}_{k}s_k s_k^\intercal \mathrm{B}_{k}}{s_k^\intercal \mathrm{B}_{k}s_k} + \theta_k (s_k^\intercal \mathrm{B}_k s_k)w_k w_k^\intercal\biggr] + \frac{y_k y_k^\intercal}{y_k^\intercal s_k}
\end{equation}
with its inverse update given by,
\begin{equation*}
\label{eqn:general_quasi2_selfscaled}
    \mathrm{H}_{k+1} = \tau_k \biggl[\mathrm{H}_{k} - \frac{\mathrm{H}_{k}y_k y_k^\intercal \mathrm{H}_{k}}{y_k^\intercal \mathrm{H}_{k}y_k} + \phi_k (y_k^\intercal \mathrm{H}_k y_k)v_k v_k^\intercal\biggr] + \frac{s_k s_k^\intercal}{y_k^\intercal s_k}
\end{equation*}
where $s_k = x_{k+1} - x_k$ represents the change in parameters, $y_k = \nabla f_{k+1} - \nabla f_{k}$ represents the change in gradients, and $\mathrm{H}_k = \mathrm{B}_k^{-1}$ approximates the inverse Hessian. The remaining terms $v_k$, $\phi_k$, and $\theta_k$ are defined in Appendix \ref{appx:ssbroyden}.

Specifically, the JAX-based optimistix \cite{optimistix2024} library in combination with the SSBroyden \cite{bioli2026selfscaledbroydenfamilyquasinewton} wrapper was used for the implementation of SSBroyden in the current work. While the source implementation of the wrapper \cite{bioli2026selfscaledbroydenfamilyquasinewton} is based on \textit{Equinox}, the implementation in the current work was modified to facilitate JIT-compilation for better speed and compatibility with our existing Flax implementation of the neural networks.

\subsection{Kolmogorov-Arnold Networks}

KANs \cite{liu2025kan} replace the fixed activation functions of conventional MLPs with learnable univariate basis functions on the network edges, enabling function approximation grounded in the Kolmogorov-Arnold representation theorem. Although the original architecture is based on spline basis functions, it is known to struggle with high computational cost. Faster alternatives such as the Chebyshev basis \cite{ss2024chebyshev} have been proposed, though they are shown to be highly unstable \cite{shukla2024comprehensive} in the context of solving PDEs. To this end, the recently proposed Feature-Enriched KAN (FEKAN) \cite{menon2026fekan} overcomes this computational bottleneck while improving stability, convergence, and attaining extreme parametric efficiency compared to KAN. The formulation of FEKAN is defined as follows,
\begin{equation}
\label{eq:kolmogorov-features-explicit}
f\big(\mathbf{x}, u_1(\mathbf{x}), \dots, u_m(\mathbf{x})\big) 
= \sum_{i=1}^{2(n+m)+1} 
\phi_i \Bigg(
\underbrace{\sum_{q=1}^{n} \psi_q(x_q)}_{\text{original inputs}} \;+\; 
\underbrace{\sum_{j=1}^{m} \psi_{n+j}(u_j(\mathbf{x}))}_{\text{additional features}}
\Bigg)
\end{equation}
\noindent
where $n$ and $m$ are the inputs and additional continuous features, respectively. See, \cite{menon2026fekan} for more details.
The additional terms in Equation \eqref{eq:kolmogorov-features-explicit} emerge from the feature enrichment, which can range from higher-order polynomials to Fourier features. Essentially, the representation of KAN can be derived from FEKAN in Equation~\eqref{eq:kolmogorov-features-explicit} by removing the additional features.

Beyond computational efficiency, KANs are distinguished by their \textit{interpretability}, enabled by learnable internal basis functions absent in conventional MLPs. This property has also been leveraged for discovering governing physical laws from data by incorporating symbolic structures into the architecture \cite{faroughi2026symbolic}. For PDEs, physics-informed KANs (PI-KANs) \cite{wang2025kolmogorov,zhao2025pikan,guo2025physics,shukla2024comprehensive} incorporate governing equations into the loss function, analogous to PINNs, with trainable basis-function parameters. PI-FEKAN \cite{menon2026fekan} follows the same framework but augments the KAN with a feature-enrichment layer that maps input coordinates to a richer representation before the KAN backbone, using orthogonal sine-cosine pairs at varying frequencies as Fourier features, with the number of enrichment terms serving as an additional architectural hyperparameter.

In the context of domain decomposition, we observe a lack of relevant studies utilizing KANs. The existing works are limited to 2D domains where continuity is ensured through predefined unity basis functions for the \textit{ansatz}, which can become non-trivial for 3D domains. The current work extends the proposed strategies for overcoming \textit{gradient conflicts} in domain decomposition to KAN-based architectures. Specifically, we utilize the FEKAN architecture for the empirical studies to showcase the efficacy of gradient surgery using the proposed Norm-PCGrad to attain higher accuracy at minimal-to-no increase in computational cost. For a comprehensive discussion of the recent advancements in KAN architectures and practical guidance for practitioners, the reader is referred to the review by Noorizadegan et al. \cite{noorizadegan2026practitioner}.\\
\subsection{Theoretical Analysis of Gradient Surgery}
In this section, we analyze the geometric properties of PCGrad and Norm-PCGrad in the two-loss setting. We first establish the conflict-free property of PCGrad, then characterize its magnitude bias and show how normalization reduces this dependence. We further show that Norm-PCGrad has a max--min optimality property and is equivalent to the two-gradient ConFIG direction. These theoretical results are restricted to two losses.

\begin{theorem}[Conflict-free property of PCGrad with two-gradients]
For two nonzero, non-collinear conflicting gradients, the PCGrad update
\begin{equation*}
\mathbf{g}_{\mathrm{PCGrad}} = \mathbf{p}_1+\mathbf{p}_2
\end{equation*}
satisfies
\begin{align*}
\mathbf{g}_1^\top\mathbf{g}_{\mathrm{PCGrad}}&= \|\mathbf{p}_2\|^2>0,
\\
\mathbf{g}_2^\top\mathbf{g}_{\mathrm{PCGrad}}&=\|\mathbf{p}_1\|^2>0.
\end{align*}
Therefore, $\mathbf{g}_{\mathrm{PCGrad}}$ is a simultaneous
first-order descent direction for both loss functions.
More details on proof of theorem is described in Appendix \ref{thm1:pcgrad_conflict_free}.
\end{theorem}

\begin{theorem} [Magnitude bias of PCGrad and normalization]
Let
\begin{equation*}
\mathcal{S}_c(\mathbf{g_1},\mathbf{g_2}) = \frac{\mathbf{g}_1^\top\mathbf{g}_2}
{\|\mathbf{g}_1\|\,\|\mathbf{g}_2\|}
\end{equation*}
denote the cosine similarity between the two gradients. Then the orthogonalized components satisfy
\begin{align*}
\|\mathbf{p}_1\|
&=
\|\mathbf{g}_2\|\sqrt{1-c^2},
\\
\|\mathbf{p}_2\|
&=
\|\mathbf{g}_1\|\sqrt{1-c^2}.
\end{align*}
Consequently, the relative first-order contribution of the two losses
under PCGrad is
\begin{equation}
\boxed{
\frac{
\mathbf{g}_1^\top\mathbf{g}_{\mathrm{PCGrad}}
}{
\mathbf{g}_2^\top\mathbf{g}_{\mathrm{PCGrad}}
}
=
\frac{\|\mathbf{g}_1\|^2}
{\|\mathbf{g}_2\|^2}.
}
\end{equation}

\noindent
Thus, PCGrad exhibits a quadratic dependence on the relative gradient magnitudes. In contrast, define the normalized projected components
\begin{equation*}
\mathbf{u}_1
=
\frac{\mathbf{p}_1}{\|\mathbf{p}_1\|},
\qquad
\mathbf{u}_2
=
\frac{\mathbf{p}_2}{\|\mathbf{p}_2\|},
\end{equation*}
and the Norm-PCGrad update
\begin{equation*}
\mathbf{g}_{\mathrm{Norm}}
=
\mathbf{u}_1+\mathbf{u}_2.
\end{equation*}
Then
\begin{align*}
\mathbf{g}_1^\top\mathbf{g}_{\mathrm{Norm}}
&=
\|\mathbf{p}_2\|,
\\
\mathbf{g}_2^\top\mathbf{g}_{\mathrm{Norm}}
&=
\|\mathbf{p}_1\|,
\end{align*}
and therefore
\begin{equation}
\boxed{
\frac{
\mathbf{g}_1^\top\mathbf{g}_{\mathrm{Norm}}
}{
\mathbf{g}_2^\top\mathbf{g}_{\mathrm{Norm}}
}
=
\frac{\|\mathbf{g}_1\|}
{\|\mathbf{g}_2\|}.
}
\end{equation}

\noindent
Hence, normalization changes the quadratic magnitude dependence of PCGrad to a linear dependence and removes the explicit weighting of the aggregate direction by the magnitudes of the projected gradients. More details on proof of theorem is described in Appendix \ref{thm2:magnitude_bias}.
\end{theorem}

\begin{theorem} [Max--min optimality of Norm-PCGrad]
Under the assumptions above, let
\begin{equation*}
\mathbf{u}_1
=
\frac{\mathbf{p}_1}{\|\mathbf{p}_1\|},
\qquad
\mathbf{u}_2
=
\frac{\mathbf{p}_2}{\|\mathbf{p}_2\|}.
\end{equation*}
The unit-norm Norm-PCGrad direction
\begin{equation*}
\mathbf{d}_{\mathrm{Norm}}
=
\frac{
\mathbf{u}_1+\mathbf{u}_2
}{
\|\mathbf{u}_1+\mathbf{u}_2\|
}
\end{equation*}
solves the max--min optimization problem
\begin{equation}
\boxed{
\mathbf{d}_{\mathrm{Norm}}
=
\underset{\|\mathbf d\|=1}{\operatorname{arg\,max}}
\;
\min
\left\{
\mathbf d^\top\mathbf u_1,
\mathbf d^\top\mathbf u_2
\right\}.
}
\end{equation}
Therefore, Norm-PCGrad selects the unit-norm direction that maximizes the worst-case alignment with the two normalized orthogonalized gradient components. More details on proof of theorem is described in Appendix \ref{thm3:maxmin_normpcgrad}.
\end{theorem}

\begin{corollary} [Equivalence of Norm-PCGrad and two-gradient ConFIG directions]
For two nonzero, non-collinear conflicting gradients, the
Norm-PCGrad direction is identical to the two-gradient ConFIG
direction:
\begin{equation}
\boxed{
\mathbf d_{\mathrm{Norm}}
=
\mathcal{U}
\left[
\mathcal{U}(\mathbf{p}_1)
+
\mathcal{U}(\mathbf{p}_2)
\right]
=
\mathbf g_v.
}
\end{equation}
Hence, PCGrad, Norm-PCGrad, and ConFIG have the same conflict-free direction in the two-loss case only after accounting for the normalization of the projected components. PCGrad differs through magnitude weighting, whereas Norm-PCGrad and ConFIG generate the same unit direction. More details on proof of theorem is described in Appendix \ref{cor1:normpcgrad_config}.
\end{corollary}

\section{Results}
In the current section, we perform elaborate empirical experiments to demonstrate the benefits of gradient surgery to domain decomposition methods. We specifically study the effectiveness of existing state-of-the-art algorithms like PCGrad and ConFIG for solving different PDEs with varying complexity arising from differences in the number of dimensions and interfaces in the domain decomposition problem. We also compare the performance of these existing algorithms with the newly proposed Norm-PCGrad. We utilize a learning rate of $5 \times 10^{-4}$ and Glorot \cite{glorot2010understanding} initialization for all empirical studies unless stated otherwise.

It is worth noting that the loss weights $\lambda_{\rm PDE}, \lambda_{u}, \lambda_{u_{\rm avg}}$ and $\lambda_{\mathcal{R}}$ in Equation \eqref{eqn:loss_function} are fixed rather than individually optimized, and we rely on gradient surgery using a suitable algorithm to handle the resulting \textit{gradient conflicts} directly in the loss landscape during training. For all experiments, we use fixed weights: $\lambda_{\rm PDE} = \lambda_{\mathcal{R}} = 1$ and $\lambda_{u} = \lambda_{u_{\rm avg}} = 20$ unless stated otherwise. In all our experiments, data is only sampled from the boundary/initial values for solving a given PDE. More information about the data sampling is provided in Appendix \ref{appx:sampling}. All implementations are based on JAX and run on an RTX6000B GPU with 16 CPU cores and 16 GB of system memory. Each experiment is repeated across multiple random seeds drawn from the set $\{12, 123, 1234, 12345, 44, 444, 4444\}$, with three or more seeds selected depending on problem size.

Consider a 2D domain decomposed into two subdomains, $\Omega_1$ and $\Omega_2$. The model accuracy is evaluated using the relative $L_2$ error,
\begin{equation}
    \text{(Relative $L_2$ Error)}_{\Omega_\#} = \frac{\|u^{\Omega_\#}_{\text{pred}} - u^{\Omega_\#}_{\text{exact}}\|_2}{\|u^{\Omega_\#}_{\text{exact}}\|_2},
\end{equation}
computed separately over $\Omega_1$ and $\Omega_2$ (i.e., $\# \in \{1, 2\}$) using each subdomain's own network, where $u^{\Omega_\#}_{\text{pred}}$ denotes the predicted solution from the $\Omega_\#$ network and $u^{\Omega_\#}_{\text{exact}}$ denotes the reference solution over $\Omega_\#$. At the interface, we additionally report a consistency error,
\begin{equation}
    \text{(Relative $L_2$ Error)}_{\Omega_1 \cap \Omega_2} = \frac{\|u^{\Omega_1\cap\Omega_2}_{1,\rm pred} - u^{\Omega_1\cap\Omega_2}_{2,\rm pred}\|_2}{\|u^{\Omega_1\cap\Omega_2}_{\text{exact}}\|_2},
\end{equation}
where $u^{\Omega_1\cap\Omega_2}_{1,\rm pred}$ and $u^{\Omega_1\cap\Omega_2}_{2,\rm pred}$ denote the predictions of the $\Omega_1$ and $\Omega_2$ networks evaluated at the interface points, and $u^{\Omega_1\cap\Omega_2}_{\text{exact}}$ is the reference solution at those points. This metric measures the disagreement between the two subnetworks at the interface, normalized by the reference solution, and is applicable for both overlapping and non-overlapping interfaces. Together, these three metrics separate domain-wise accuracy from interfacial consistency, and the same formulation extends to domain decompositions with more than two subdomains.

To quantify gradient conflicts during training, we utilize the cosine similarity score $\mathcal{S}_c(\mathbf{g}_1, \mathbf{g}_2)$, computed using Equation~\eqref{eq:similarity}, where $\mathbf{g}_1$ and $\mathbf{g}_2$ are selected from the set of gradients corresponding to different loss terms in the composite loss function. The moving averages of different gradient pairs across training iterations are aggregated and represented as a density distribution over $[-1,1]$. Gradient conflicts are characterized based on the following criteria:
\begin{itemize}
    \item $\mathcal{S}_c \in [-1,0)$: gradients are in conflict.
    \item $\mathcal{S}_c \approx 0$: gradients are orthogonal.
    \item $\mathcal{S}_c \in (0,1]$: gradients are aligned and conflict-free.
\end{itemize}
For a domain decomposition with two subdomains, gradient conflicts are analyzed using $\mathcal{S}_c^{\Omega_1}$ and $\mathcal{S}_c^{\Omega_2}$, while for three subdomains, $\mathcal{S}_c^{\Omega_1}$, $\mathcal{S}_c^{\Omega_2}$, and $\mathcal{S}_c^{\Omega_3}$ are used, following the criteria defined above.

The following sections systematically evaluate gradient surgery for domain decomposition across five increasingly complex benchmark problems. Section~\ref{sec:2D_DD} considers a high-frequency 2D Helmholtz equation with two subdomains, introducing multi-grid SPINN (mSPINN) for efficient collocation sampling and comparing initialization sensitivity and PINN-PINN (XPINN) and SPINN-PINN configurations. Section~\ref{sec:2D_DD_2} considers a 2D Poisson equation with two Euclidean subdomains using SPINN-SPINN decomposition and evaluates SSBroyden for mitigating gradient conflicts. Sections~\ref{sec:3DDomain-2} and \ref{sec:3DDomain-3} extend the study to 3D Poisson and nonlinear sine-Gordon equations. To the best of our knowledge, Sections~\ref{sec:3DDomain-2} and \ref{sec:3DDomain-3} provide the first demonstrations of KAN-based domain decomposition for 3D linear and nonlinear PDEs. Gradient surgery remains the primary focus, with quasi-Newton methods evaluated on selected problems for comparison. The domain decomposition configurations are summarized below.

{\setlength{\tabcolsep}{2pt}
\renewcommand{\arraystretch}{1.8}
\begin{table}[H]
\centering
\footnotesize
\begin{tabular}{|c|c|c|c|c|c|c|c|}
\hline
\textbf{Section} & \textbf{PDE} & \textbf{Domain} & \textbf{Interface Type} & \textbf{Interface Losses ($\mathcal{L}_{\rm I}$)} & \textbf{Subdomain} & \textbf{Loss Terms} & \textbf{\# Conflicts} \\
\hline\hline
\multirow{2}{*}{\ref{sec:2D_DD}} & \multirow{2}{*}{Helmholtz} & \multirow{2}{*}{2D} & \multirow{2}{*}{Non-overlapping} & \multirow{2}{*}{\shortstack{Data loss}} & $\Omega_1$ & $\mathcal{L}_{\rm PDE}^{\Omega_1},~ \mathcal{L}_{\rm BC}^{\Omega_1},~ \mathcal{L}_{\rm I}^{\Omega_1\cap\Omega_2}$ & \multirow{2}{*}{\textbf{6 + 1 = 7}} \\ \cline{6-7}
 & & & & & $\Omega_2$ & $\mathcal{L}_{\rm PDE}^{\Omega_2},~ \mathcal{L}_{\rm BC}^{\Omega_2},~ \mathcal{L}_{\rm I}^{\Omega_2\cap\Omega_1}$ & \\
\hline\hline
\multirow{2}{*}{\ref{sec:2D_DD_2}} & \multirow{2}{*}{Poisson} & \multirow{2}{*}{2D} & \multirow{2}{*}{Non-overlapping} & \multirow{2}{*}{\shortstack{Data loss\\+ $1^{\rm st}$-order deriv.}} & $\Omega_1$ & $\mathcal{L}_{\rm PDE}^{\Omega_1},~ \mathcal{L}_{\rm BC}^{\Omega_1},~ \mathcal{L}_{\rm I}^{\Omega_1\cap\Omega_2}$ & \multirow{2}{*}{\textbf{6 + 1 = 7}} \\ \cline{6-7}
 & & & & & $\Omega_2$ & $\mathcal{L}_{\rm PDE}^{\Omega_2},~ \mathcal{L}_{\rm BC}^{\Omega_2},~ \mathcal{L}_{\rm I}^{\Omega_2\cap\Omega_1}$ & \\
\hline\hline
\multirow{3}{*}{\ref{sec:3DDomain-2}} & \multirow{3}{*}{Poisson} & \multirow{3}{*}{3D} & \multirow{3}{*}{Overlapping} & \multirow{3}{*}{\shortstack{Data loss\\+ $1^{\rm st}$-order deriv.}} & $\Omega_1$ & $\mathcal{L}_{\rm PDE}^{\Omega_1},~ \mathcal{L}_{\rm BC}^{\Omega_1},~ \mathcal{L}_{\rm I}^{\Omega_1\cap\Omega_2},~ \mathcal{L}_{\rm I}^{\Omega_1\cap\Omega_3}$ & \multirow{3}{*}{\textbf{12 + 2 = 14}} \\ \cline{6-7}
 & & & & & $\Omega_2$ & $\mathcal{L}_{\rm PDE}^{\Omega_2},~ \mathcal{L}_{\rm BC}^{\Omega_2},~ \mathcal{L}_{\rm I}^{\Omega_2\cap\Omega_1}$ & \\ \cline{6-7}
 & & & & & $\Omega_3$ & $\mathcal{L}_{\rm PDE}^{\Omega_3},~ \mathcal{L}_{\rm BC}^{\Omega_3},~ \mathcal{L}_{\rm I}^{\Omega_3\cap\Omega_1}$ & \\
\hline\hline
\multirow{2}{*}{\ref{sec:3DDomain-3}} & \multirow{2}{*}{\shortstack{Sine-\\Gordon}} & \multirow{2}{*}{3D} & \multirow{2}{*}{Overlapping} & \multirow{2}{*}{\shortstack{Data loss\\+ $1^{\rm st}$-order deriv.}} & $\Omega_1$ & $\mathcal{L}_{\rm PDE}^{\Omega_1},~ \mathcal{L}_{\rm BC}^{\Omega_1},~ \mathcal{L}_{\rm I}^{\Omega_1\cap\Omega_2}$ & \multirow{2}{*}{\textbf{6 + 1 = 7}} \\ \cline{6-7}
 & & & & & $\Omega_2$ & $\mathcal{L}_{\rm PDE}^{\Omega_2},~ \mathcal{L}_{\rm BC}^{\Omega_2},~ \mathcal{L}_{\rm I}^{\Omega_2\cap\Omega_1}$ & \\
\hline\hline
\end{tabular}
\caption{\textbf{Overview of benchmark problems and potential gradient conflicts.} Summary of PDE, domain dimensionality, interface type, interface losses, and per-subdomain loss terms with the total number of potential gradient conflicts. The interface losses comprise a data loss (Eqn~\eqref{eq:loss_uavg}) for solution continuity and, where indicated, $1^{\rm st}$-order derivative matching (Eqn~\eqref{eq:loss_R}) for flux continuity. The within-subdomain conflicts are computed as $\binom{n}{2}$ per subdomain (where $n$ is the number of loss terms) and summed across all subdomains. Additionally, one cross-subdomain conflict is added per shared interface to account for the implicit coupling between the interface gradients of adjacent subdomains. $\mathcal{L}_{\rm BC}$ and $\mathcal{L}_{\rm I}$ denote boundary and interface losses, respectively.}
\label{tab:benchmark_overview}
\end{table}
\renewcommand{\arraystretch}{1.0}}

\subsection{Helmholtz Equation: 2D Domain \texorpdfstring{$\rightarrow$}{->} Non-Euclidean \texorpdfstring{$+$}{+} Euclidean}
\label{sec:2D_DD}

We consider the two-dimensional inhomogeneous Helmholtz equation
\begin{equation}
\nabla^2 u(x,y) + k^2u(x,y) = f(x,y), \qquad (x,y)\in\Omega,
\label{eq00}
\end{equation}
where $\nabla^2u=u_{xx}+u_{yy}$ and $k$ is the wavenumber. We prescribe the manufactured solution
$u_{\text{exact}}(x,y)=\sin(\pi a_1x)\sin(\pi a_2y),$
where $a_1$ and $a_2$ control the oscillation frequencies along the $x$- and $y$-directions. Since
$u_{xx}=-(\pi a_1)^2u_{\text{exact}}, ~~
u_{yy}=-(\pi a_2)^2u_{\text{exact}},$
the corresponding source term is
$$f(x,y)=\left[k^2-\pi^2(a_1^2+a_2^2)\right]
\sin(\pi a_1x)\sin(\pi a_2y).$$

Thus, $u_{\text{exact}}$ satisfies \eqref{eq00} exactly, providing a closed-form benchmark for evaluating the trained network. The spatial domain is $\Omega=[-1,1]^2$, with $k=1$ and $a_1=a_2=10$.

\begin{figure}[H]
\centering
{
\centering
\includegraphics[width=0.7\linewidth, trim={10mm 15mm 5mm 20mm}, clip]{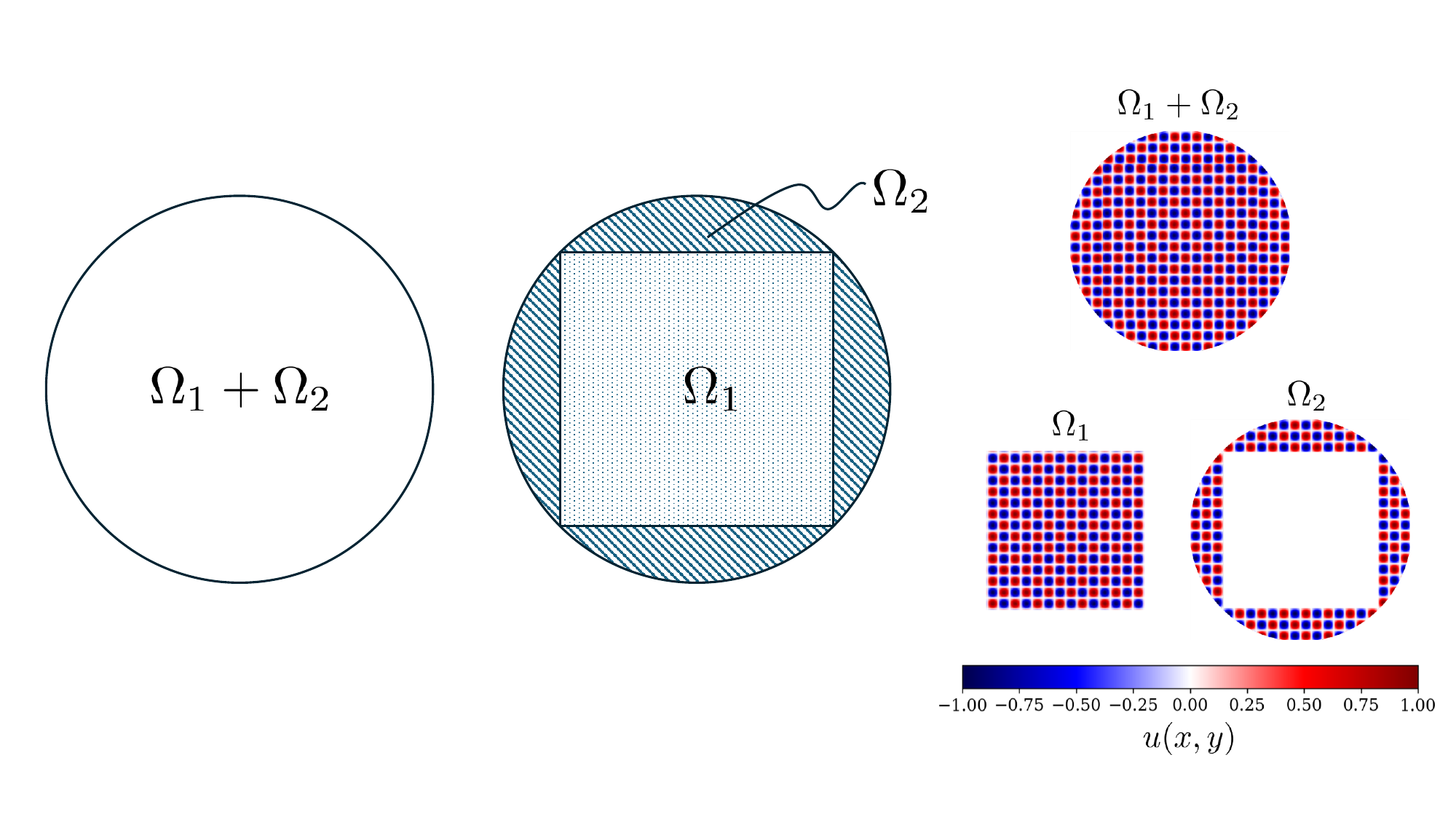}
}
\caption{\textbf{2D Helmholtz -- Domain setup.} Domain decomposition of a non-Euclidean domain for the Helmholtz equation with frequencies $a_1 = 10$ and $a_2 = 10$. The circular domain is decomposed into two non-overlapping subdomains: $\Omega_1$ (Euclidean) and $\Omega_2$ (non-Euclidean).}
\label{fig:2D_Helm}
\end{figure}

Here, we decompose the circular domain into two sub-domains which are Euclidean ($\Omega_1$) and non-Euclidean ($\Omega_2$) as shown in Figure \ref{fig:2D_Helm}. We note that this combination enables the utilization of SPINN \cite{cho2023separable} architecture to solve $\Omega_1$ while leveraging the $\mathcal{O}(Nd)$ computational cost where $d$ is the dimension and $N$ is the number of collocation points. On the other hand, $\Omega_2$ will be solved with minimal utilization of collocation points using the PINN \cite{raissi2019physics} architecture at $\mathcal{O}(N^d)$ computational cost. 

\vspace{0.15cm}
\begin{rmk}
We note that an important step in domain decomposition could be to maximize the coverage of Euclidean domain and minimize the coverage of non-Euclidean domain to make the best utilization of a separable architecture to attain superior accuracy and computational cost.
\end{rmk}

\begin{rmk}
While our earlier work \cite{menon2025anant} demonstrated that separable architectures can solve high-dimensional PDEs with linear computational cost, this work extends their use to non-Euclidean domains through domain decomposition. The proposed hybrid approach improves the efficiency of methods such as XPINN while overcoming the restriction of separable architectures such as SPINN to Euclidean domains.
\end{rmk}

\subsubsection{Better Accuracy using Multi-Grid SPINN (mSPINN)}
Besides using SPINN for domain decomposition, we also propose an improvement to the training procedure for this architecture. SPINN relies on a grid of collocation points that are structured in a tensor-product (Cartesian) fashion within the domain. However, in the current work, we propose a multi-grid strategy to sample multiple staggered collocation grids from the domain as opposed to relying solely on a fixed set of points within a single grid. This ensures that the collocation points are distributed more broadly across the domain and are not limited by the requirement of a Cartesian structure from the SPINN architecture. We define this architecture as multi-grid SPINN (mSPINN), and the difference between these methods by virtue of the sampling strategies is illustrated in Figure \ref{fig:2D_Helm_Collocation}. The pseudo-code for the mSPINN algorithm is provided in Appendix \ref{appx:mspinn}.

Here, we note that mSPINN attains a perfect balance between PINN and SPINN for speed, accuracy, and robustness while solving PDEs within a non-Euclidean domain such as the one considered in Figure \ref{fig:2D_Helm}.

\begin{figure}[H]
\centering
{
\centering
\includegraphics[width=0.6\linewidth, trim={10mm 35mm 10mm 25mm}, clip]{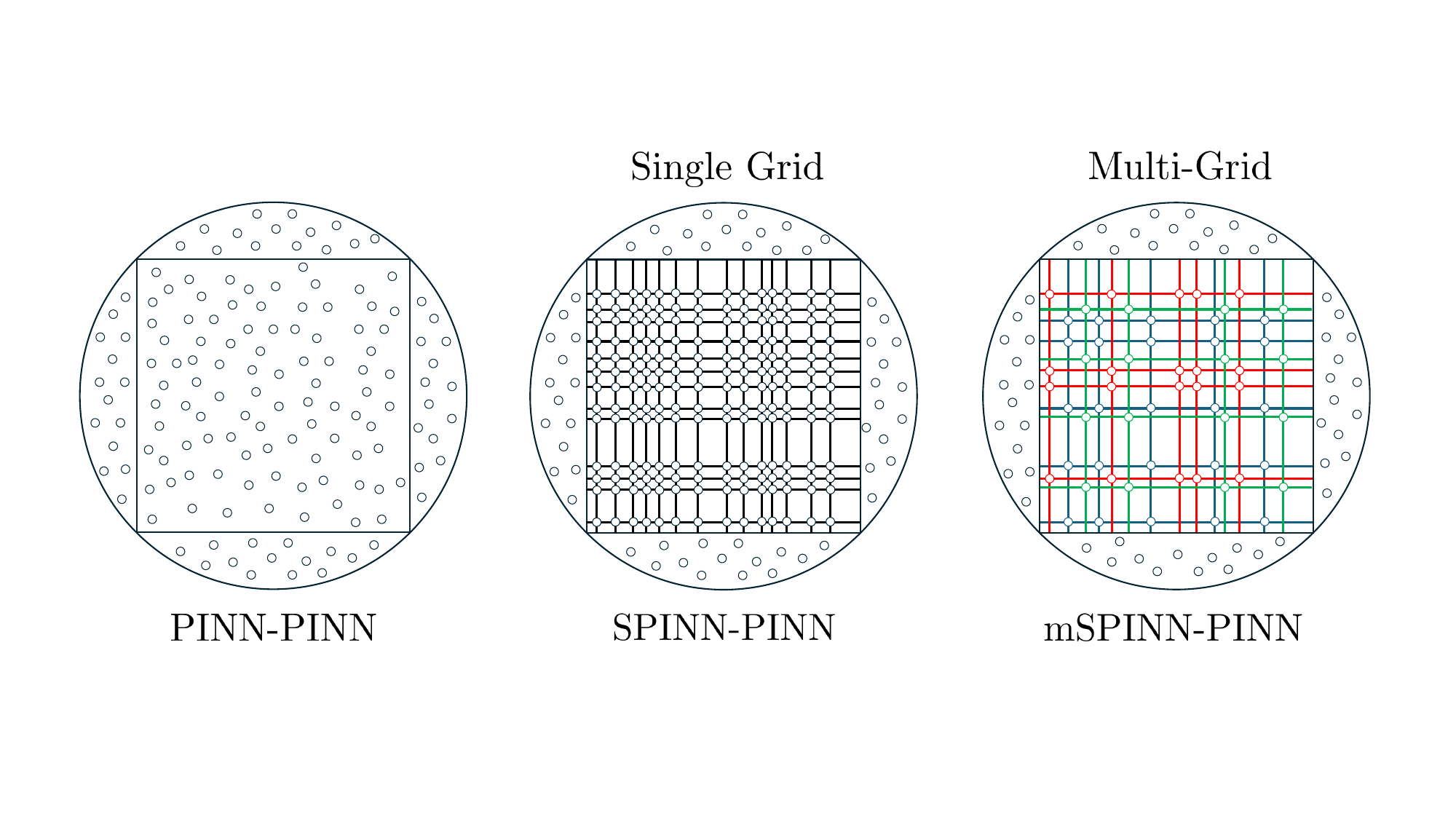}
}
\caption{\textbf{2D Helmholtz -- Collocation strategies.} Collocation sampling methods for domain decomposition of the non-Euclidean domain. With reference to Figure \ref{fig:2D_Helm}, collocation points in $\Omega_1$ can be sampled as a single grid (center) or multi-grid (right), while random sampling (left) is used for $\Omega_2$ handled by PINN.}
\label{fig:2D_Helm_Collocation}
\end{figure}

The comparison of the relative $L_2$ error and computational costs of the PINN-PINN (same as XPINN), SPINN-PINN, and mSPINN-PINN domain decompositions without gradient surgery are given in Table \ref{tab:2D_Helm_multigrid}. It is worth noting that mSPINN requires choosing two additional hyperparameters, namely: the number of batches and the batch size. Here, the number of batches is the total number of staggered collocation grids used for training, while the batch size is the fixed number of random grids chosen at each iteration during training. Also, during each training iteration, the average of the gradients from the grids is used to update the trainable parameters of the model.

\begin{table}[H]
\centering
\begin{tabular}{|c|c||c|c|c||c|c|c|}
\hline
 \multirow{2}{*}{\textbf{Method}} & \multirow{2}{*}{\textbf{\# Collocation Points}} & \multirow{2}{*}{\textbf{Batch Size}} & \multicolumn{2}{c|}{\textbf{Rel. $L_2$ Err.}} & \multirow{2}{*}{\textbf{Time (ms/iter)}} & \multirow{2}{*}{\textbf{Speed-Up}} \\ \cline{4-5}
 & & & \textbf{$\Omega_1$} & \textbf{$\Omega_2$} & & \\ \hline\hline
 PINN-PINN & $16384$ & - & 0.7617 & 0.6729 & 62.44 & \text{Baseline} \\ \cline{1-7}
 SPINN-PINN & $128 \times 128 = 16384$ & - & 1.0291 & 0.9466 & 24.26 & $(+)~2.5 \times$ \\ \cline{1-7}
 \multirow{2}{*}{mSPINN-PINN} & $(16) \times 32 \times 32 = 16384$ & 2 & 0.0733 & 0.2532 & 60.98 & $(+)~1.02 \times$ \\ \cline{2-7}
  & $(16) \times 32 \times 32 = 16384$ & 4 & \textbf{0.0387} & \textbf{0.2300} & 87.91 & $(-)~1.4 \times$  \\  
\hline\hline
\end{tabular}`
\caption{\textbf{2D Helmholtz -- Architectural comparison.} Relative $L_2$ error and computational time for PINN-PINN, SPINN-PINN, and mSPINN-PINN domain decomposition of the non-Euclidean domain shown in Figure \ref{fig:2D_Helm_Collocation}.  For SPINN-PINN, SPINN is assigned to $\Omega_1$ to leverage the computational efficiency of a separable architecture on a Euclidean subdomain.  The results shown in the table does not employ gradient surgery. It is only the demonstration of multi-grid collocation grid sampling. The relative $L_2$ error is represented as mean $\pm$ standard deviation computed from three random realizations. Bold text represents the best results.}
\label{tab:2D_Helm_multigrid}
\end{table} 

Here, utilization of a separable architecture such as SPINN for domain decomposition clearly leads to substantial speed-up compared to using PINN, as shown in Table \ref{tab:2D_Helm_multigrid}. However, it leads to accuracy degradation for a fixed number of collocation points. On the other hand, mSPINN outperforms the baseline with an order of magnitude improvement in accuracy for the Euclidean subdomain ($\Omega_1$) and an overall improvement for the domain as a whole. We observe a minor speed-up using mSPINN with a batch size of 2, which is not as substantial as the acceleration observed while using SPINN. Furthermore, we note that accuracy improves further by increasing the batch size, albeit at a larger computational cost, leading to a slowdown compared to the baseline.

\subsubsection{Handling Gradient Conflicts through Gradient Surgery}

The current section focuses on leveraging gradient surgery to overcome \textit{gradient conflicts} in the domain decomposition of the non-Euclidean domain considered in Figure \ref{fig:2D_Helm}. We specifically compare the different architectural configurations for domain decomposition --- PINN-PINN, SPINN-PINN, and mSPINN-PINN --- in terms of accuracy and computational cost. We then apply different algorithms for gradient surgery, namely PCGrad and ConFIG, to mitigate the conflict between gradients during training, and compare the performance of these existing state-of-the-art algorithms with the proposed Norm-PCGrad algorithm.

\begin{figure}[H]
\centering
{
\centering
\includegraphics[width=0.9\linewidth, trim={10mm 5mm 5mm 10mm}, clip]{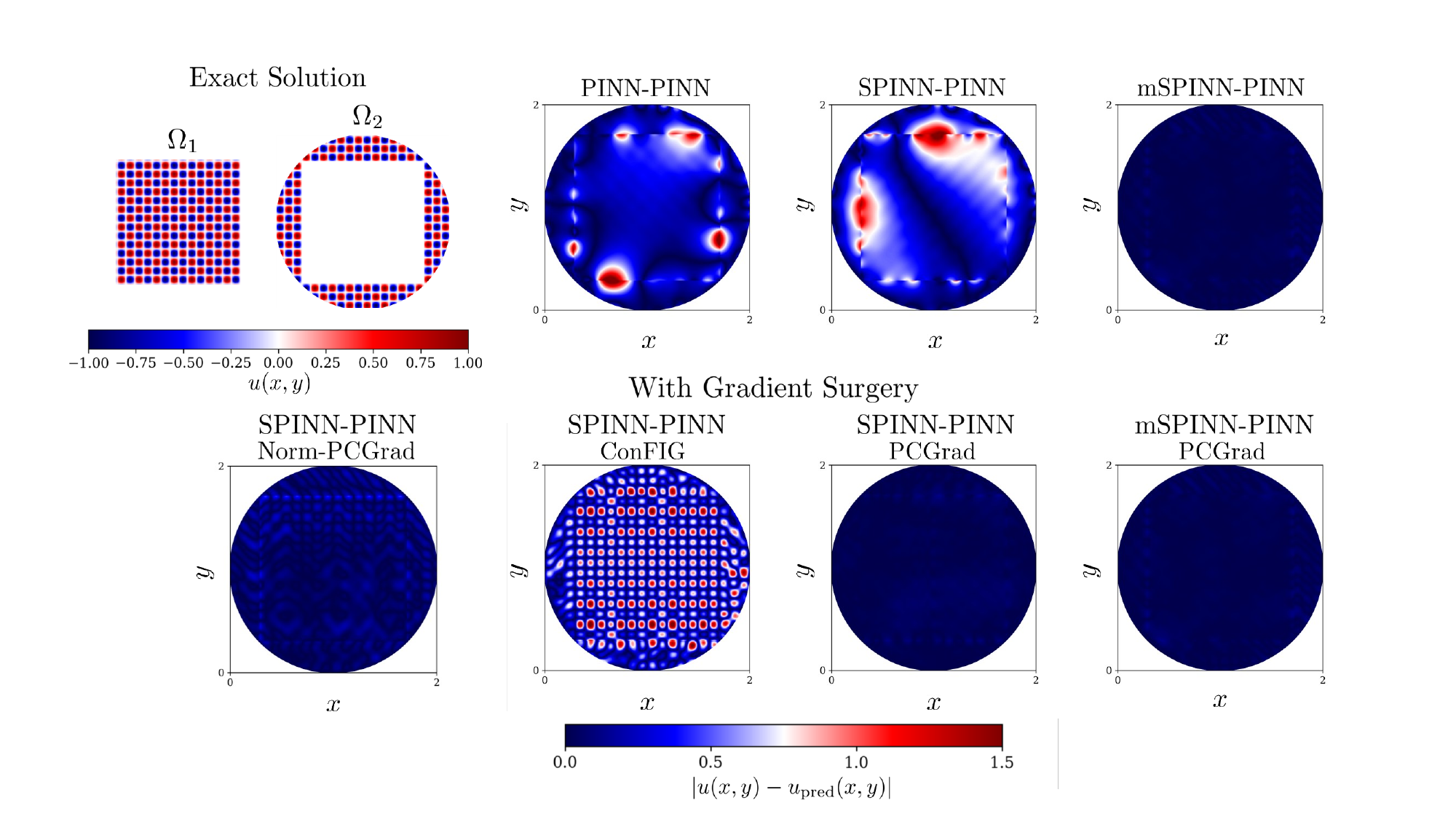}
}
\caption{\textbf{2D Helmholtz -- Absolute error.} Point-wise absolute error for domain decomposition of the non-Euclidean domain with frequencies $a_1 = 10$ and $a_2 = 10$. (Top Row) Without gradient surgery. (Bottom Row) With gradient surgery.} 
\label{fig:2D_Helm_abserror}
\end{figure}

Figure \ref{fig:2D_Helm_abserror} presents the point-wise absolute error while solving the Helmholtz equation through domain decomposition. Specifically, Figure \ref{fig:2D_Helm_abserror} (Top Row) shows the results without gradient surgery. Here, the multi-grid configuration mSPINN-PINN clearly has advantages in solution quality compared to other configurations. However, these gains come at a substantially higher computational cost, consistent with our observations in the previous section. Based on Figure \ref{fig:2D_Helm_abserror} and Table \ref{tab:2D_Helm_gradient_surgery}, SPINN-PINN offers gains in computational efficiency at the expense of accuracy, while mSPINN-PINN achieves superior accuracy but at a considerably higher cost. We then apply gradient surgery to these configurations and observe substantial improvements in accuracy relative to the PINN-PINN baseline. We attribute these improvements to the mitigation of gradient conflicts between the different loss terms within the composite loss function. Figure \ref{fig:2D_Helm_abserror} (Bottom Row) illustrates the point-wise absolute error after implementing gradient surgery using ConFIG, PCGrad, and Norm-PCGrad for SPINN-PINN and mSPINN-PINN domain decompositions. Although ConFIG has demonstrated exceptional performance \cite{liu2025config} while solving PDEs in single domains, it is observed to undergo substantial degradation in its performance for SPINN-PINN domain decomposition. A more recent work \cite{kim2026per} also reported a similar degradation in performance for ConFIG while solving different PDEs in single domains compared to existing algorithms like PCGrad.

\begin{table}[H]
\centering
\begin{tabular}{|c|c||c|c||c|}
\hline
 \multirow{2}{*}{\textbf{Method}} & \multirow{2}{*}{\textbf{Gradient Surgery}} & \multicolumn{2}{c|}{\textbf{Relative $L_2$ Error}} & \multirow{2}{*}{\textbf{Time (ms/iter)}} \\ \cline{3-4}
 & & \textbf{$\Omega_1$} & \textbf{$\Omega_2$} & \\ \hline\hline
 PINN-PINN & \multirow{3}{*}{None} & 1.2090 $\pm$ 0.6386 & 0.9681 $\pm$ 0.8690 & 1.89 \\ \cline{1-1}\cline{3-5}
 SPINN-PINN & & 1.0781 $\pm$ 0.2099 & 0.9288 $\pm$ 0.1944 & 1.29 \\ \cline{1-1}\cline{3-5}
 mSPINN-PINN & & 0.4386 $\pm$ 0.5898 & 0.4792 $\pm$ 0.2820 & 11.73 \\ \cline{1-5}
 \multirow{3}{*}{SPINN-PINN} & ConFIG & 1.1147 $\pm$ 0.1307 & 0.7966 $\pm$ 0.0889 & 2.52 \\ \cline{2-5}
 & Norm-PCGrad & \textbf{0.1312 $\pm$ 0.0314} & \textbf{0.1655 $\pm$ 0.0125} & 3.75 \\ \cline{2-5}
 & \multirow{2}{*}{PCGrad} & 0.4569 $\pm$ 0.6389 & 0.2146 $\pm$ 0.2333 & 3.71 \\ \cline{1-1} \cline{3-5}
 \multirow{2}{*}{mSPINN-PINN} & & \textbf{0.0785 $\pm$ 0.0591} & \textbf{0.0885 $\pm$ 0.0659} & 12.96 \\ \cline{2-5}
 & Norm-PCGrad & 0.1577 $\pm$ 0.0652 & 0.1382 $\pm$ 0.0240 & 12.81 \\
\hline\hline
\end{tabular}
\caption{\textbf{2D Helmholtz -- Gradient surgery.} Relative $L_2$ error (mean $\pm$ std. devi) and computational time for gradient surgery across PINN-PINN, SPINN-PINN, and mSPINN-PINN domain decomposition of the non-Euclidean domain shown in Figure \ref{fig:2D_Helm_Collocation}.  mSPINN-PINN uses a total of 16 grids with a batch size of 4. Bold text represents the best results.}
\label{tab:2D_Helm_gradient_surgery}
\end{table}

In contrast, PCGrad provides substantial accuracy gains with modest computational overhead, as shown in Table~\ref{tab:2D_Helm_gradient_surgery}, while mSPINN-PINN further improves accuracy at significantly higher cost. In contrast, Norm-PCGrad with SPINN-PINN achieves a substantial reduction in relative $L_2$ error at comparable computational cost to PCGrad, whereas combining Norm-PCGrad with mSPINN-PINN offers limited accuracy gains while increasing cost due to multi-grid aggregation. Overall, mSPINN-PINN improves accuracy at higher computational cost, whereas SPINN-PINN with Norm-PCGrad provides the most favorable accuracy--efficiency trade-off, reducing the relative $L_2$ error compared with PINN-PINN at only one-third of the multi-grid computational cost.

\begin{rmk}
    While comparing the area between the two subdomains, it is worth noting that $A_{\Omega_1} \approx 1.75$, and $A_{\Omega_2}$ for the domain shown in Figure \ref{fig:2D_Helm}, where $\Omega_1$ (Euclidean) covers roughly 63.7\% of the circular domain, and where $A_{\Omega_i}$ denotes the area of $\Omega_i$. Since SPINN \cite{cho2023separable} can handle more collocation points at comparatively lower computational cost than PINN by virtue of its separable architecture, we observe better accuracy for $\Omega_1$ compared to $\Omega_2$. The accuracy of $\Omega_2$ may be further improved by sampling more collocation points, albeit at a higher computational cost. Therefore, it is beneficial to maximize the coverage of the Euclidean subdomain over the non-Euclidean subdomain to attain higher accuracy at a lower computational cost.  
\end{rmk}

\begin{figure}[H]
\centering
{
\centering
\includegraphics[width=0.8\linewidth, trim={10mm 14mm 5mm 10mm}, clip]{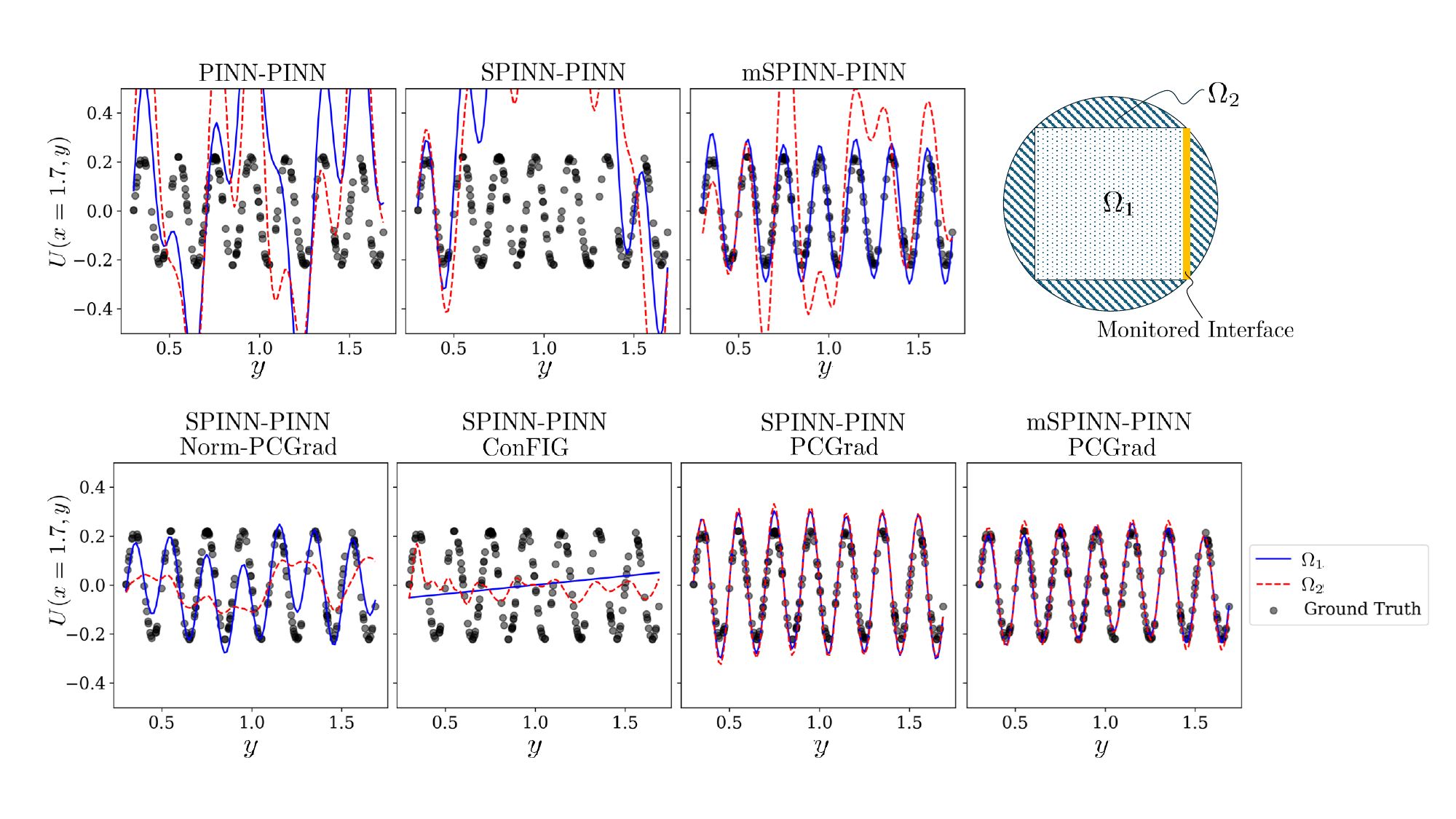}
}
\caption{\textbf{2D Helmholtz -- Interface comparison.} Solution comparison at the interface between $\Omega_1$ and $\Omega_2$ for the Helmholtz equation with frequencies $a_1 = 10$ and $a_2 = 10$. (Top Row) Without gradient surgery. (Bottom Row) With gradient surgery.} 
\label{fig:2D_Helm_interface}
\end{figure}

Table \ref{tab:2D_Helm_gradient_surgery} can be considered as representative of the accuracy for the global solution. However, for domain decomposition, it is important to also evaluate performance locally at the interface region. In the current work, we also evaluate the solution continuity at the interface between the subdomains to clearly observe the advantages of mitigating \textit{gradient conflicts}. Figure \ref{fig:2D_Helm_interface} presents the comparison at the interfaces of the subdomains $\Omega_1$ and $\Omega_2$, with and without gradient surgery. While mSPINN-PINN demonstrates moderate accuracy for the global solution without any gradient surgery as shown in Figure \ref{fig:2D_Helm_abserror}, its accuracy at the interface is noticeably degraded. However, both SPINN-PINN and mSPINN-PINN showcase accurate interface continuity using PCGrad, with the latter slightly better than the former. On the other hand, while Norm-PCGrad has good accuracy at the interface for $\Omega_1$, it struggles with $\Omega_2$. Furthermore, its solution approximation at the interface is bounded: it neither fluctuates vigorously as is the case without gradient surgery, nor drops significantly below the ground truth as in the case of ConFIG. Thus, mSPINN-PINN is justified by its superior global and interface accuracy (Figures~\ref{fig:2D_Helm_abserror} and \ref{fig:2D_Helm_interface}), but not by computational efficiency in the current serial implementation.

\begin{figure}[H]
\centering
{
\centering
\includegraphics[width=\linewidth, trim={0mm 50mm 0mm 50mm}, clip]{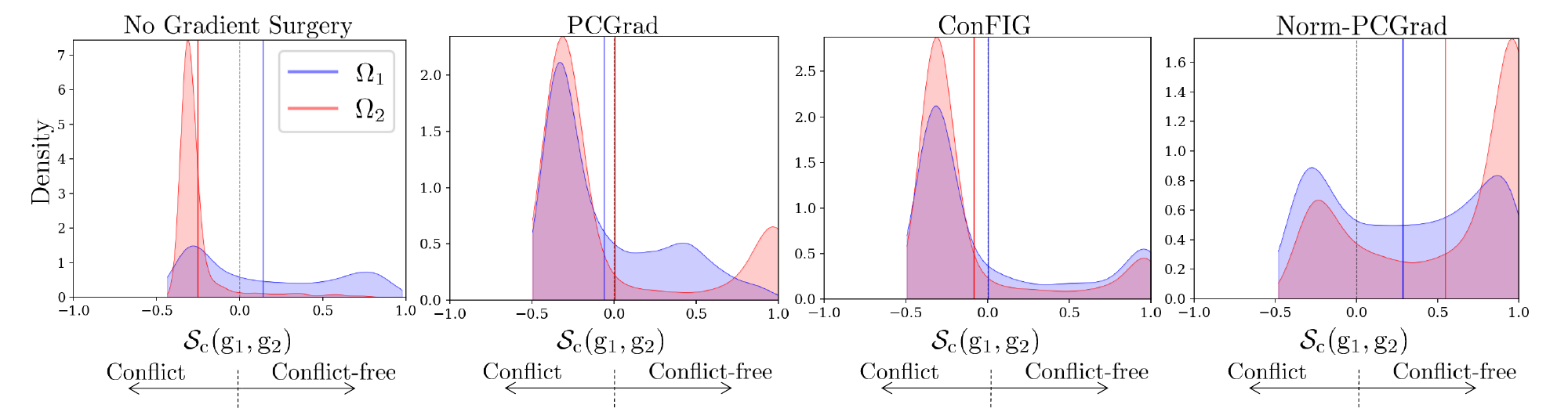}}
\caption{\textbf{2D Helmholtz -- Cosine Similarity.} Density distribution of cosine similarity scores for the two subdomains, $\mathcal{S}_c^{\Omega_1}$ and $\mathcal{S}_c^{\Omega_2}$. Vertical solid lines represent the mean of each distribution ($\mathcal{S}_{c,~\text{mean}}^{\Omega_1}$ and $\mathcal{S}_{c,~\text{mean}}^{\Omega_2}$), while the dashed line marks the zero-line where gradients are orthogonal. If $\mathcal{S}_{c,~\text{mean}}^{\Omega_1}$ and/or $\mathcal{S}_{c,~\text{mean}}^{\Omega_2} < 0$, the gradients were in conflict for the majority of training in $\Omega_1$ and/or $\Omega_2$. On the other hand, if $\mathcal{S}_{c,~\text{mean}}^{\Omega_1}$ and/or $\mathcal{S}_{c,~\text{mean}}^{\Omega_2} > 0$, the gradients were conflict-free for the majority of training in $\Omega_1$ and/or $\Omega_2$. The extent of gradient conflicts (or their suppression) is measured by the distance of $\mathcal{S}_{c,~\text{mean}}^{\Omega_1}$ and $\mathcal{S}_{c,~\text{mean}}^{\Omega_2}$ from the zero-line. Here, the cosine similarity scores between gradient pairs are computed at fixed intervals during training and their distribution is represented as the density distribution.}
\label{fig:2D_Helm_cosine_similarity}
\end{figure}

While solving the Helmholtz equation using an SPINN-PINN domain decomposition, we compare the gradient conflicts across different algorithms of gradient surgery using the cosine similarity score, as shown in Figure \ref{fig:2D_Helm_cosine_similarity}. Without any gradient surgery, we observe that $\mathcal{S}_{c,~\text{mean}}^{\Omega_1} > 0$, denoting conflict-free gradients in $\Omega_1$, while $\mathcal{S}_{c,~\text{mean}}^{\Omega_2} \ll 0$, denoting severe gradient conflicts in $\Omega_2$. On the other hand, PCGrad and ConFIG attain $\mathcal{S}_{c,~\text{mean}}^{\Omega_1}$ and $\mathcal{S}_{c,~\text{mean}}^{\Omega_2} \approx 0$, indicating that their gradients were nearly orthogonal for the majority of training. This means that the gradients have overcome conflicts but are still not aligned with each other. We note that Norm-PCGrad attains conflict-free gradients for the majority of training, since both $\mathcal{S}_{c,~\text{mean}}^{\Omega_1}$ and $\mathcal{S}_{c,~\text{mean}}^{\Omega_2} \gg 0$. These observations further support the trend of relative $L_2$ error for SPINN-PINN domain decomposition shown in Table \ref{tab:2D_Helm_gradient_surgery}.

\subsubsection{Sensitivity to Initialization} 
\begin{figure}[H]
\centering
{
\centering
\includegraphics[width=0.85\linewidth, trim={7mm 10mm 3mm 7mm}, clip]{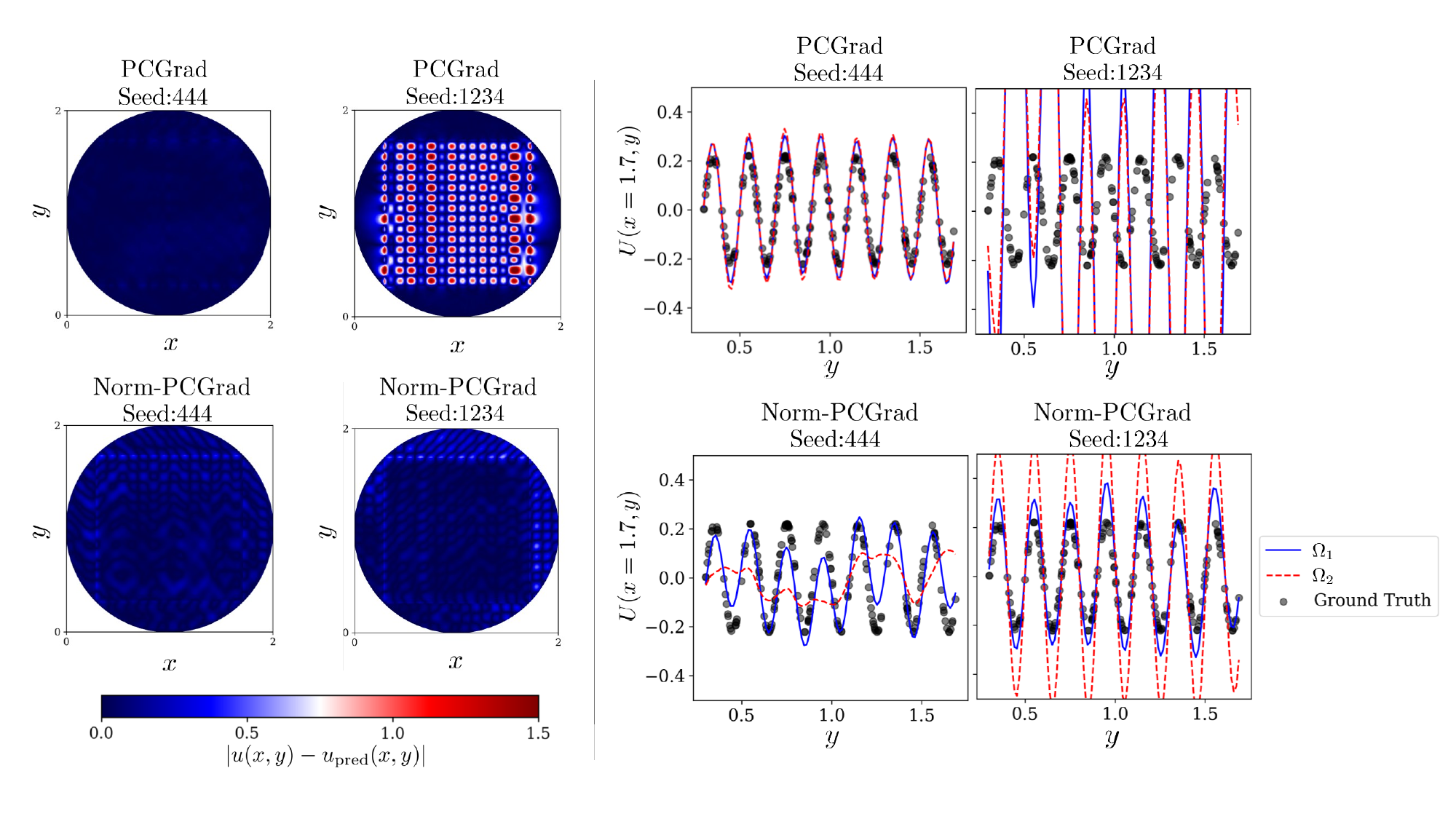}
}
\caption{\textbf{2D Helmholtz -- Initialization sensitivity.} Sensitivity to random initialization using Glorot \cite{glorot2010understanding} for the Helmholtz equation with frequencies $a_1 = 10$ and $a_2 = 10$. (Left) Absolute error. (Right) Interface solution for random seeds 444 and 1234.}
\label{fig:2D_Helm_initialize}
\end{figure}

In spite of the performance gains observed for different algorithms of gradient surgery in Table \ref{tab:2D_Helm_gradient_surgery}, it is worth understanding the sensitivity of these gains to random initialization for a high-frequency Helmholtz equation. Based on the different random seeds used in the current study, we observe large deviations in performance when changing the seeds for the Helmholtz equation with $a_1 = 10$ and $a_2 = 10$, as shown in Figure \ref{fig:2D_Helm_initialize}. However, we also observe a reduction in this sensitivity as the frequencies ($a_1$ and $a_2$) are decreased.

Combining the results from Figure \ref{fig:2D_Helm_initialize} and Table \ref{tab:2D_Helm_gradient_surgery}, we observe that PCGrad incurs a larger relative $L_2$ error due to its poor performance for specific initialization seeds. In contrast, Norm-PCGrad achieves an overall reduction in relative $L_2$ error, confirmed by an order of magnitude improvement compared to PCGrad. While the interface continuity is not perfectly attained for Norm-PCGrad, the deviation from the ground truth remains limited and bounded across different seeds. This demonstrates the robustness of Norm-PCGrad over PCGrad under varying initializations while solving high-frequency Helmholtz equation through domain decomposition.

\subsection{Poisson Equation: 2D Domain \texorpdfstring{$\rightarrow$}{->} Euclidean \texorpdfstring{$+$}{+} Euclidean}
\label{sec:2D_DD_2}
\begin{figure}[H]
\centering
{
\centering
\includegraphics[width=0.8\linewidth, trim={0mm 30mm 0mm 20mm}, clip]{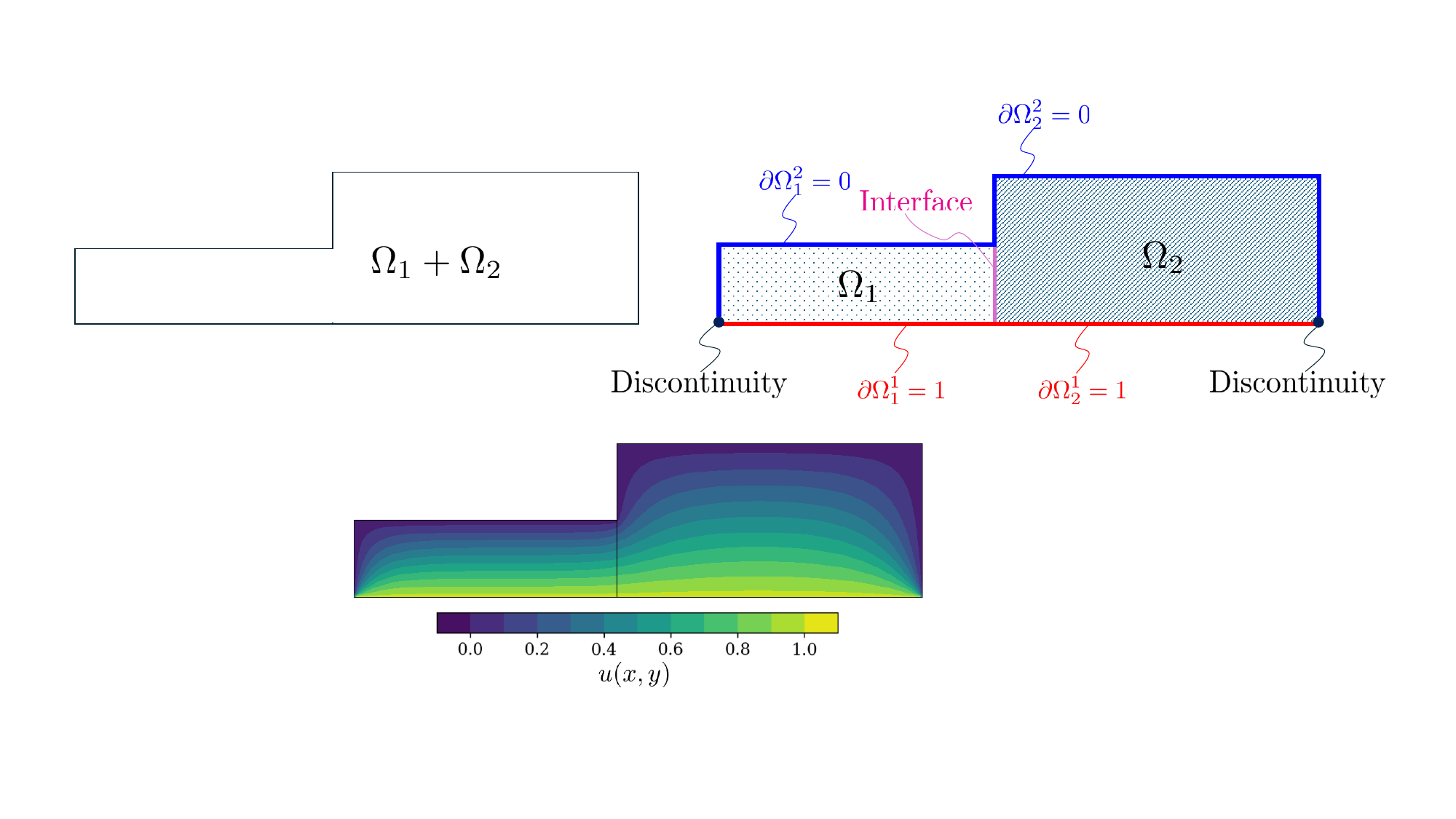}}
\caption{\textbf{2D Poisson -- Domain setup.} Decomposition of a 2D domain into Euclidean subdomains through a non-overlapping interface.}
\label{fig:2D_DD_domain}
\end{figure}

We consider the steady-state Poisson equation in a two-dimensional domain,
\begin{equation}
\Delta u(\mathbf{x}) = q(\mathbf{x}), \qquad \mathbf{x}\in\Omega,
\label{eq:poisson}
\end{equation}
subject to Dirichlet boundary conditions
\begin{align*}
u(\mathbf{x}) &= 1, \qquad \mathbf{x}\in\partial\Omega^1,  \\
u(\mathbf{x}) &= 0, \qquad \mathbf{x}\in\partial\Omega^2,
\end{align*}
where $\Omega=[0,1]^2$, $q$ is the forcing term, and $\partial\Omega^1$ and $\partial\Omega^2$ are distinct boundary portions. In the context of steady-state heat conduction, $u$ represents temperature and $q$ internal heat generation; hence, the imposed conditions correspond to a fixed potential (temperature) difference, with $u|{\partial\Omega^1}>u|{\partial\Omega^2}$, as illustrated in Figure~\ref{fig:2D_DD_domain}. Here, $q=0$, reducing \eqref{eq} to the homogeneous Laplace equation
\begin{equation}
\Delta u(\mathbf{x})=0, \qquad \mathbf{x}\in\Omega,
\label{eq}
\end{equation}
so that, in the absence of internal sources, the solution is driven solely by the imposed boundary potential (temperature) difference.

\subsubsection{PINN Failure: Choosing the Correct Interface}
\begin{figure}[H]
\centering
{
\centering
\includegraphics[width=0.9\linewidth, trim={5mm 40mm 10mm 30mm}, clip]{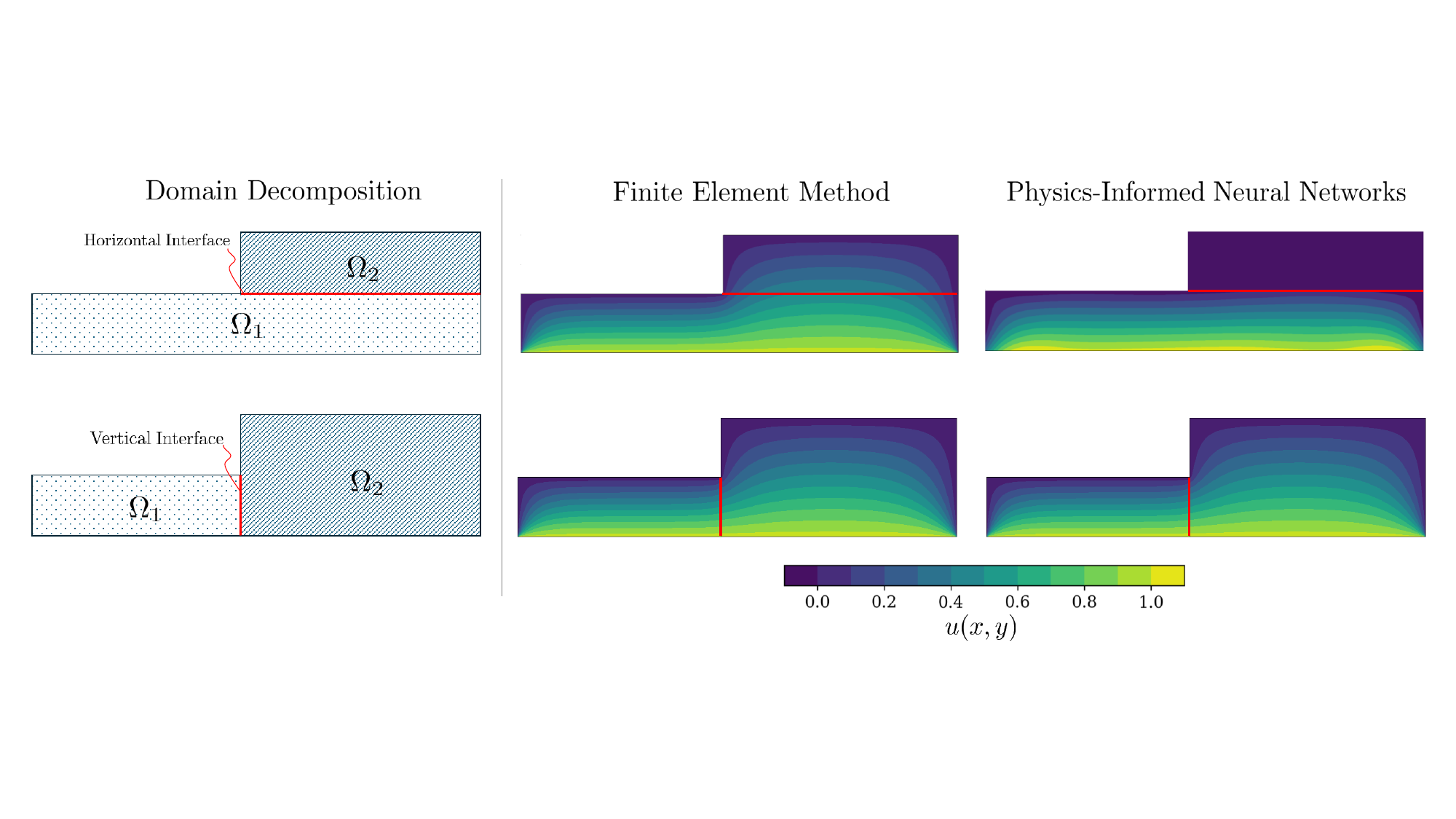}}
\caption{\textbf{2D Poisson -- Interface failure.} Comparison between FEM and PINN solutions for different interface locations, illustrating PINN sensitivity to interface placement.}
\label{fig:2D_DD_failure_interface}
\end{figure}

For the current problem setup, we decompose the domain into two subdomains at the non-overlapping interface as shown in Figure \ref{fig:2D_DD_domain}. However, it is also reasonable to test other interfaces that can lead to a different set of subdomains as shown in Figure \ref{fig:2D_DD_failure_interface}. The boundary conditions for the interface comparison are similar to the boundary conditions described for Figure \ref{fig:2D_DD_domain}.
\noindent
Compared to conventional numerical solvers such as the Finite Element Method (FEM), PINNs are highly sensitive to the interface locations for domain decomposition. We observe from Figure \ref{fig:2D_DD_failure_interface} that vertical interfaces lead to the same solutions for both FEM and PINNs as expected. While the PINN solution is vastly different from the FEM solution for a horizontal interface, the FEM solution remains unchanged for both interfaces.

Both PINN solutions for the vertical and horizontal interfaces satisfy the underlying PDE by minimizing the residual loss and also ensure continuity across the interface between the two subdomains. However, the global solution for the whole domain still deviates largely from the expected true solution obtained from FEM for the horizontal interface. We attribute this PINN failure in the context of domain decomposition to the manner in which the optimization of the loss functions is carried out for the two subdomains connected by a horizontal interface. We make the following observations:
\begin{itemize}
    \item With respect to the boundary conditions shown in Figure \ref{fig:2D_DD_domain}, the solution field is driven by the potential difference between the bottom ($\partial\Omega^1$) and top ($\partial\Omega^2$) boundaries $-$ $u|_{\partial\Omega^1} > u|_{\partial\Omega^2}$.
    \item For a vertical interface, we note that both subdomains have portions of $\partial\Omega^1$ and $\partial\Omega^2$ (denoted as $\partial\Omega_1^1, \partial\Omega_1^2$ for $\Omega_1$ and $\partial\Omega_2^1, \partial\Omega_2^2$ for $\Omega_2$) influencing the optimization of the loss functions for the two subdomains.
    \item However, for a horizontal interface, we note that only $\Omega_1$ has both $\partial\Omega_1^1$ and $\partial\Omega_1^2$ while $\Omega_2$ is strictly limited to $\partial\Omega_2^2$. In other words, the high-potential boundary $\partial\Omega_2^1$ that drives the solution field does not exist for $\Omega_2$.
    \item The optimizer therefore minimizes the loss function constituting the residual loss, the boundary loss, and the interface loss for subdomain $\Omega_2$ with only $\partial\Omega_2^2 = 0$ and an interface between the subdomains. This leads to the trivial solution (\textit{zero everywhere within} $\Omega_2$) that still minimizes all the losses for $\Omega_1$ and $\Omega_2$ in a ``physics-informed'' manner while ensuring continuity across the interface, a clear PINN failure case.
    \item As a result, we observe an unexpected deviation of the PINN solution from the FEM solution for a horizontal interface as shown in Figure \ref{fig:2D_DD_failure_interface}. This in turn means that a converged solution from a PINN model does not necessarily guarantee a physically realizable solution unless a correct interface is chosen for domain decomposition.
\end{itemize}
\noindent

It is also worth noting that the choice of interface location can critically affect the solution quality for domain decomposition using PINNs, independent of any gradient surgery applied. A horizontal interface can lead to a trivial zero solution in one of the subdomains due to the absence of information from the high-potential boundary condition. A detailed investigation of this failure mode across different gradient surgery algorithms is also provided in Appendix \ref{appx:2D_DD_Failure}. Therefore, the aforementioned discussion emphasizes the need to select a suitable interface while solving PDEs through domain decomposition using PINNs. These observations for the 2D domain hold true and extend to 3D domains as well.

\subsubsection{Gradient Surgery to tackle Gradient Conflicts}
For the domain decomposition discussed in the current section, we utilize SPINN-SPINN domain decomposition since we can decompose the current domain into two Euclidean subdomains, instead of the SPINN-PINN domain decomposition discussed in the previous section. As shown in the previous section, utilizing a separable architecture such as SPINN has its computational advantages for domain decomposition, and we leverage this extensively in the current domain.

\begin{figure}[H]
\centering
{
\centering
\includegraphics[width=0.9\linewidth, trim={15mm 20mm 15mm 10mm}, clip]{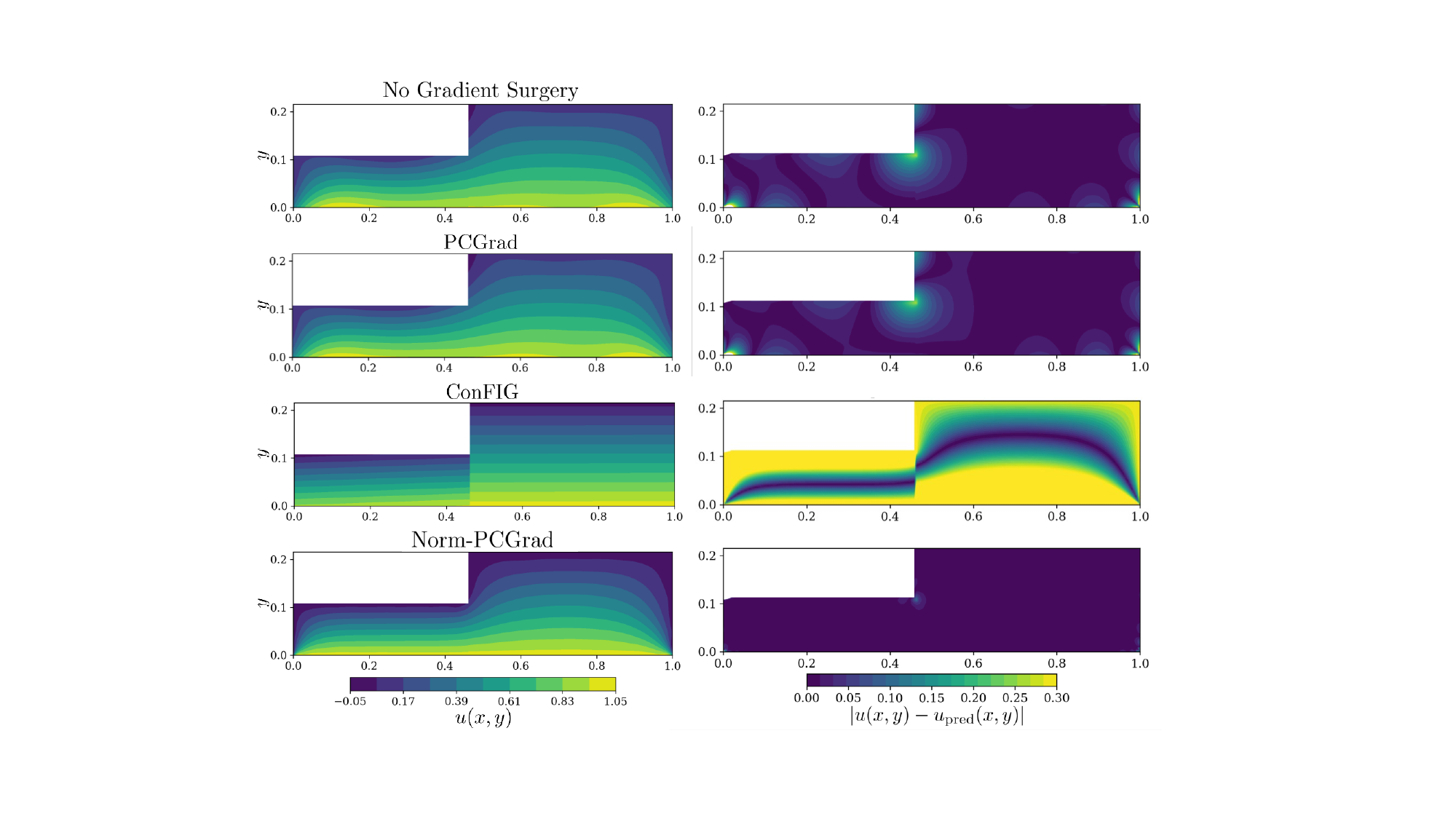}}
\caption{\textbf{2D Poisson -- Absolute error.} Point-wise absolute error for 2D domain decomposition: (a) without gradient surgery, (b) with PCGrad, (c) with ConFIG, and (d) with Norm-PCGrad.}
\label{fig:2D_DD_abserror}
\end{figure}

Here, we test and compare the influence of gradient surgery on the accuracy of the solution from domain decomposition of the domain described in Figure \ref{fig:2D_DD_domain}. It is worth noting that the discontinuity denoted in Figure \ref{fig:2D_DD_domain} emerges at the lower corners of the domain where the two sets of Dirichlet boundary conditions $-$ $\partial\Omega^1 = 1$ and $\partial\Omega^2 = 0$ meet. It is expected for the model to struggle with learning the corner section of the domain accurately if gradient conflicts are present during training. Furthermore, we also investigate different combinations of derivative orders at the interface loss formulation to ensure continuity, and we observe that first-order spatial derivatives facilitate the best solution continuity compared to other combinations that predominantly include second-order derivatives. This is notable despite the direct presence of second-order derivatives in the Poisson equation given by Equation \eqref{eq:poisson}. A detailed comparison of different combinations at the interface is shown in Appendix \ref{appx:2D_DD}.

We implement PCGrad and ConFIG to ensure conflict-free gradients and compare their performance against the baseline without any gradient surgery. Furthermore, the proposed Norm-PCGrad algorithm is also introduced here to test its performance against the existing algorithms for gradient surgery.
Figure \ref{fig:2D_DD_abserror} shows the comparison between the solutions to the Poisson equation using different algorithms for gradient surgery. We examine the predicted solution field and the point-wise absolute error for the different cases, compared against the reference solution from the FEM solver. 

With reference to the global solution field shown in Figure \ref{fig:2D_DD_abserror}, we observe reasonable accuracy for domain decomposition without any gradient surgery. With PCGrad, there is minor improvement in accuracy, reflected by comparable magnitude of the point-wise absolute error. On the other hand, ConFIG substantially degrades the performance, leading to an order of magnitude increase in the error compared to the baseline. However, Norm-PCGrad showcases exceptional performance in predicting an accurate global solution field, evident from the near zero absolute error compared to the baseline.  

With respect to the direction of propagation of the solution field, we observe a smooth propagation represented by a smooth and continuous gradient from the high-potential boundary to the low-potential boundary ($\partial \Omega^1 \rightarrow \partial\Omega^2$) across all cases shown in Figure \ref{fig:2D_DD_abserror}.

\begin{figure}[H]
\centering
{
\centering
\includegraphics[width=0.9\linewidth, trim={0mm 20mm 0mm 10mm}, clip]{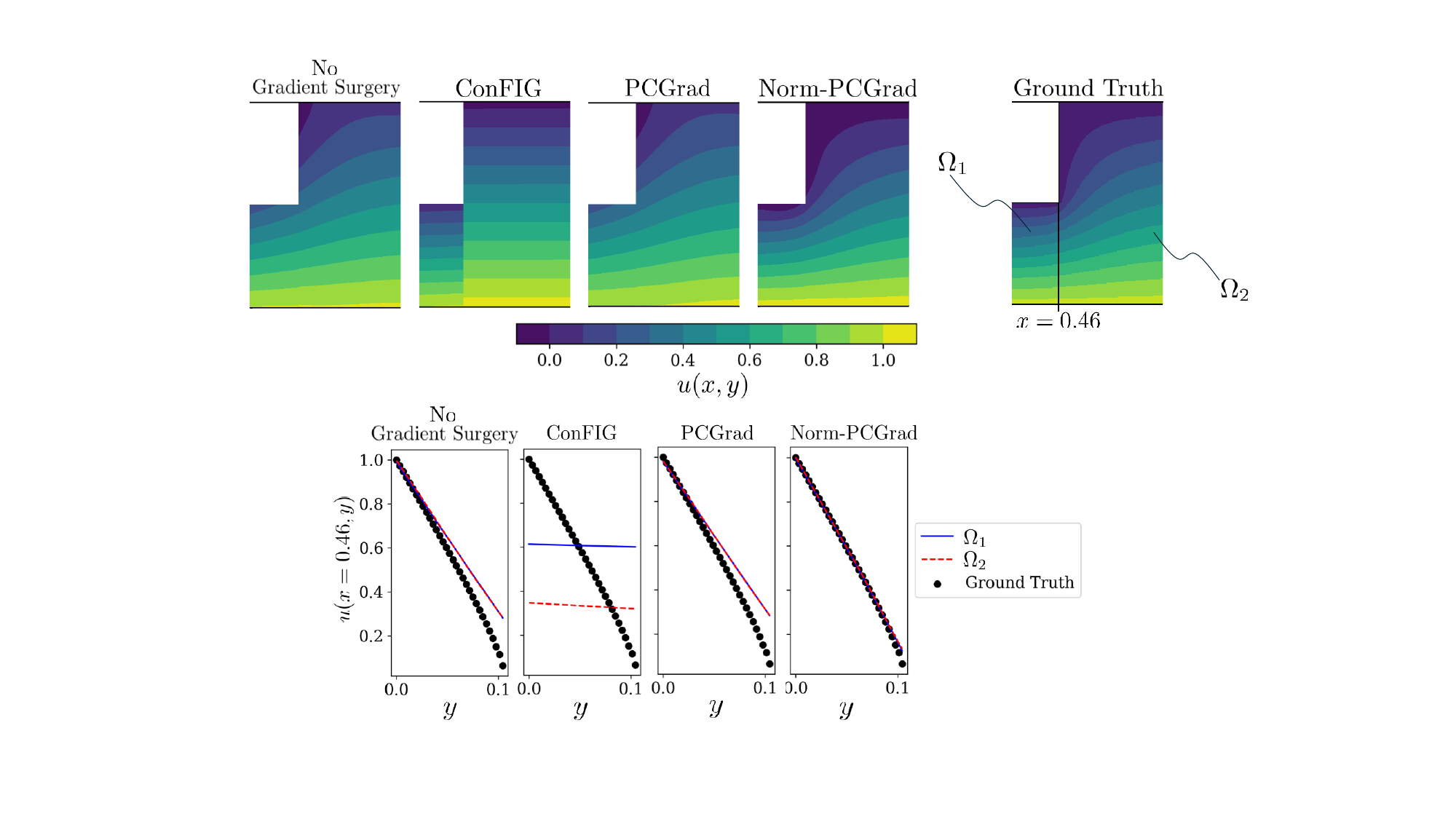}}
\caption{\textbf{2D Poisson -- Interface comparison.} Solution comparison at the interface ($x = 0.46$) of $\Omega_1$ and $\Omega_2$: (a) without gradient surgery, (b) with ConFIG, (c) with PCGrad, (d) with Norm-PCGrad.}
\label{fig:2D_DD_interface}
\end{figure}

Another important criterion for any domain decomposition problem is the interface continuity. Figure \ref{fig:2D_DD_interface} clearly shows the continuity across the interface between the two subdomains for domain decomposition by a vertical interface. It can be observed that Norm-PCGrad outperforms all other cases to attain a continuous and smooth solution field across the interface. Comparing the solution of Norm-PCGrad at the interface with the ground truth reveals near-perfect alignment. On the other hand, the interface continuity for domain decomposition without gradient surgery and with PCGrad is satisfactory. Moreover, their alignment with the ground truth deviates as they move away from the lower boundary $\partial\Omega^1$, as evident from Figure \ref{fig:2D_DD_interface}. Similar to the observations for the Helmholtz equation in the previous section, ConFIG shows discontinuities at the interface also for the Poisson equation considered in this section.

\begin{figure}[H]
\centering
{
\centering
\includegraphics[width=0.9\linewidth, trim={10mm 50mm 10mm 50mm}, clip]{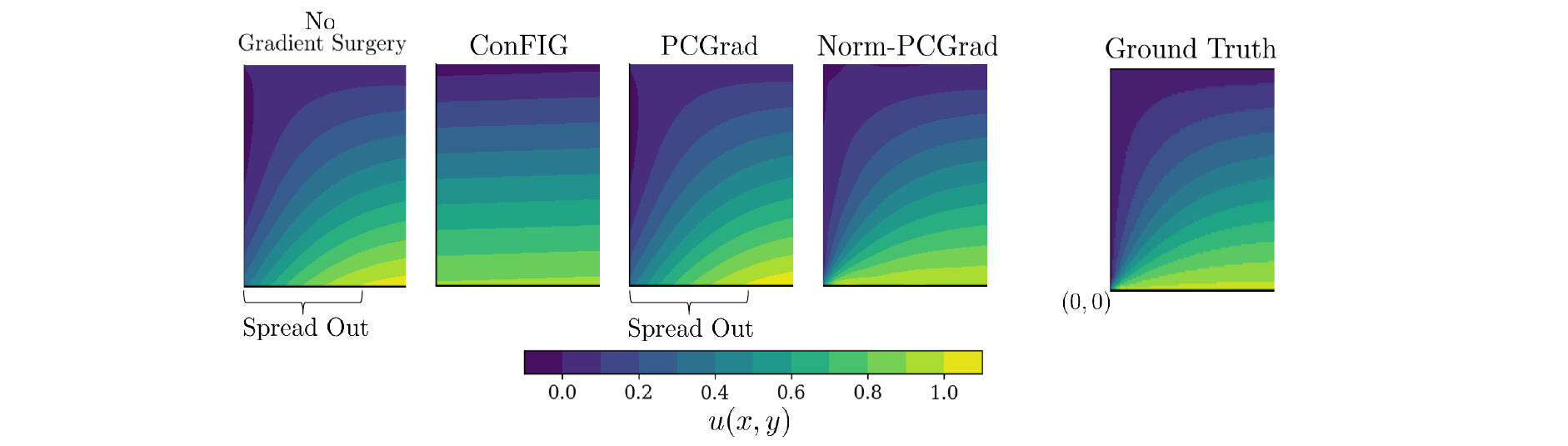}}
\caption{\textbf{2D Poisson -- Corner comparison.} Solution comparison at the corner $(0, 0)$ of $\Omega_1$: (a) without gradient surgery, (b) with ConFIG, (c) with PCGrad, (d) with Norm-PCGrad.}
\label{fig:2D_DD_corner}
\end{figure}

\noindent
Furthermore, by comparing the accuracy of each case at the discontinuities, $(0,0)$ and $(1,0)$, we observe bare minimum error accumulation for Norm-PCGrad. However, all other cases with and without gradient surgery struggle at the corner points, as evidenced by the localized error concentration observed in Figure \ref{fig:2D_DD_abserror}. This is also reflected as a spreading of the error for PCGrad as the solution approaches the corner points from the interior of the subdomains. With reference to Figure \ref{fig:2D_DD_corner}, we observe something interesting. For cases without any gradient surgery and with PCGrad, we observe a spread out of the solution field as we approach the corner, $(0,0)$. On the hand, there is no such spread out observed for Norm-PCGrad at the corners leading to accurate representation of the solution compared to the ground truth.

\begin{figure}[H]
\centering
{
\centering
\includegraphics[width=\linewidth, trim={0mm 50mm 0mm 50mm}, clip]{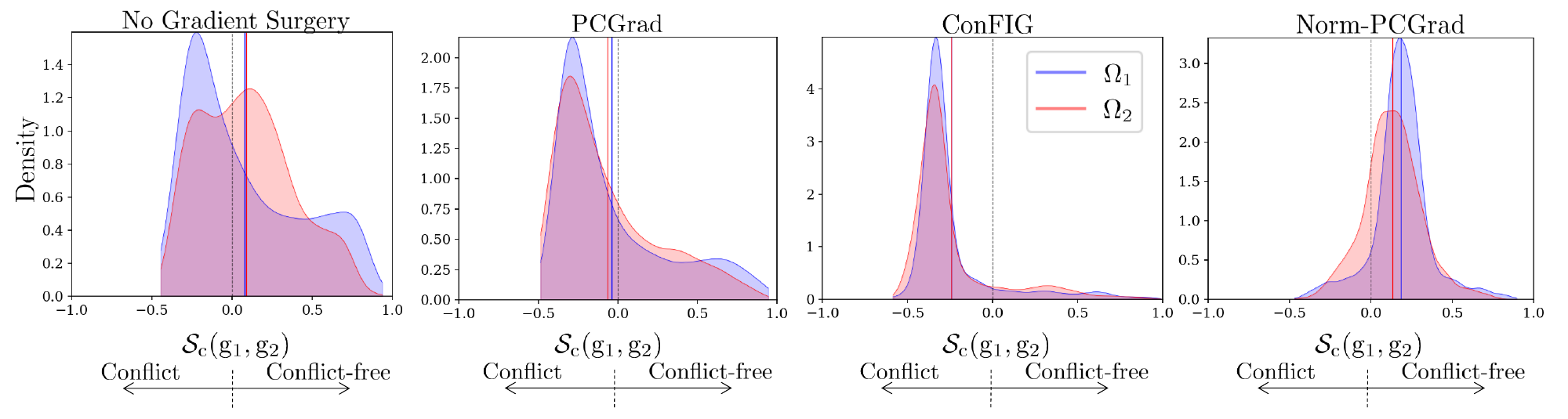}}
\caption{\textbf{2D Poisson -- Cosine Similarity.} Density distribution of cosine similarity scores for the two subdomains, $\mathcal{S}_c^{\Omega_1}$ and $\mathcal{S}_c^{\Omega_2}$. Vertical solid lines represent the mean of each distribution ($\mathcal{S}_{c,~\text{mean}}^{\Omega_1}$ and $\mathcal{S}_{c,~\text{mean}}^{\Omega_2}$), while the dashed line marks the zero-line where gradients are orthogonal. If $\mathcal{S}_{c,~\text{mean}}^{\Omega_1}$ and/or $\mathcal{S}_{c,~\text{mean}}^{\Omega_2} < 0$, the gradients were in conflict for the majority of training in $\Omega_1$ and/or $\Omega_2$. On the other hand, if $\mathcal{S}_{c,~\text{mean}}^{\Omega_1}$ and/or $\mathcal{S}_{c,~\text{mean}}^{\Omega_2} > 0$, the gradients were conflict-free for the majority of training in $\Omega_1$ and/or $\Omega_2$. The extent of gradient conflicts (or their suppression) is measured by the distance of $\mathcal{S}_{c,~\text{mean}}^{\Omega_1}$ and $\mathcal{S}_{c,~\text{mean}}^{\Omega_2}$ from the zero-line. Here, the cosine similarity scores between gradient pairs are computed at fixed intervals during training and their distribution is represented as the density distribution.}
\label{fig:2D_DD_cosine_similarity}
\end{figure}

While solving the 2D Poisson equation using an SPINN-SPINN domain decomposition, we compare the gradient conflicts across different algorithms of gradient surgery using the cosine similarity score, as shown in Figure \ref{fig:2D_DD_cosine_similarity}. Without any gradient surgery, we observe that $\mathcal{S}_{c,~\text{mean}}^{\Omega_1}$ and $\mathcal{S}_{c,~\text{mean}}^{\Omega_2} > 0$, denoting conflict-free gradients in both $\Omega_1$ and $\Omega_2$. On the other hand, PCGrad attains $\mathcal{S}_{c,~\text{mean}}^{\Omega_1}$ and $\mathcal{S}_{c,~\text{mean}}^{\Omega_2} < 0 \text{ or } \approx 0$, indicating that its gradients were slightly conflicting or nearly orthogonal for the majority of training. Applying ConFIG leads to $\mathcal{S}_{c,~\text{mean}}^{\Omega_1}$ and $\mathcal{S}_{c,~\text{mean}}^{\Omega_2} \ll 0$, making its utilization counterproductive. We note that Norm-PCGrad attains stronger gradient alignment for the majority of training, with both $\mathcal{S}_{c,~\text{mean}}^{\Omega_1}$ and $\mathcal{S}_{c,~\text{mean}}^{\Omega_2} \gg 0$, substantially exceeding the baseline case without gradient surgery. These observations further support the trend of relative $L_2$ error for SPINN-SPINN domain decomposition shown in Table \ref{tab:2D_DD_SSBroyden_rl2_Surgery}. A YouTube video showing the solution is available at \href{https://www.youtube.com/watch?v=na3FuxI0XBk}{2D Poisson's equation with conflict-free gradients} or https://www.youtube.com/watch?v=na3FuxI0XBk. 

\subsubsection{Gradient Surgery + Curvature-Aware Optimization to tackle Gradient Conflicts}

In the previous sections, we addressed \textit{gradient conflicts} through gradient surgery with the Adam optimizer; although recent works \cite{kiyani2025optimizing}\cite{wang2026gradient} have explored curvature-aware quasi-Newton optimizers for mitigating such conflicts, their effectiveness for PDE domain decomposition remains unclear. Here, we combine gradient projection with quasi-Newton optimization, using Norm-PCGrad with Adam during the initial warm-up phase followed by the SSBroyden optimizer, providing an intermediate approach between first-order methods such as Adam and computationally expensive second-order Newton methods.

\begin{figure}[H]
\centering
\begin{subfigure}[b]{0.44\linewidth}
    \centering
    \includegraphics[width=\linewidth, trim={0mm 5mm 0mm 0mm}, clip]{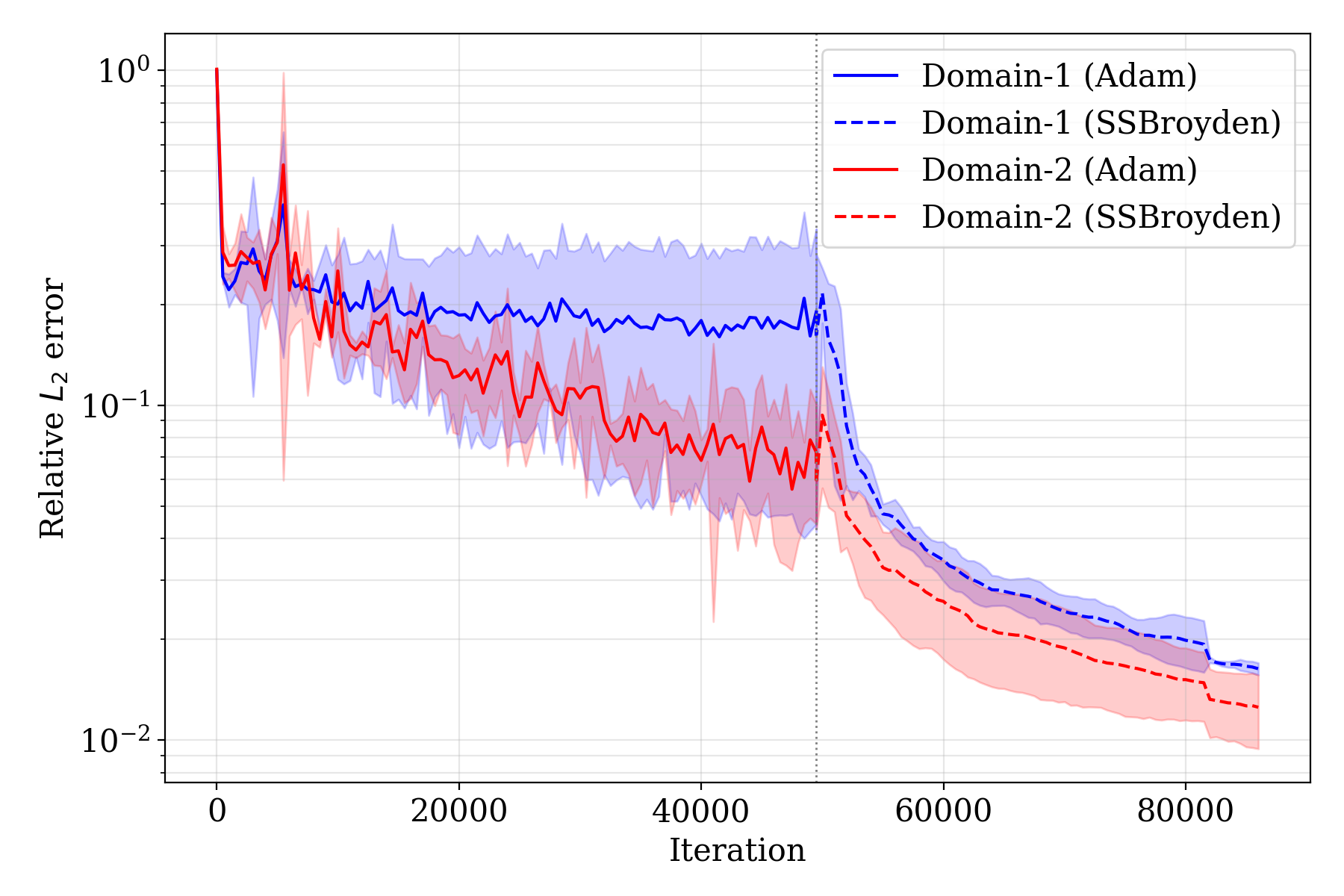}
    \label{fig:2D_DD_SSBroyden_rl2_left}
\end{subfigure}
\begin{subfigure}[b]{0.44\linewidth}
    \centering
    \includegraphics[width=\linewidth, trim={0mm 5mm 0mm 0mm}, clip]{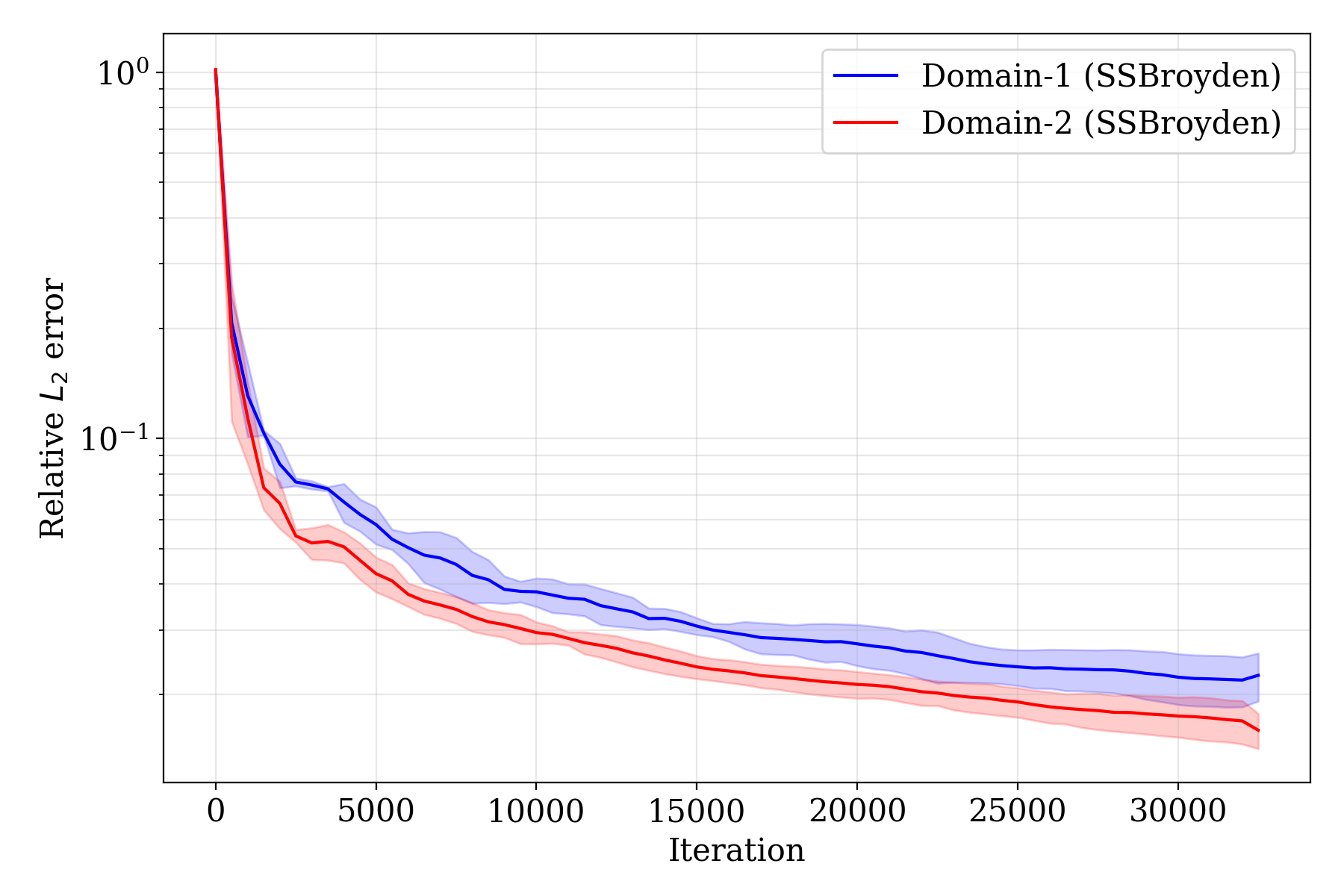}
    \label{fig:2D_DD_SSBroyden_rl2_right}
\end{subfigure}
\caption{\textbf{2D Poisson -- SSBroyden convergence.} Convergence of relative $L_2$ error for 2D domain decomposition using double-precision training. (Left) Adam with Norm-PCGrad followed by SSBroyden. (Right) SSBroyden only, without gradient surgery.  All cases with SSBroyden is evaluated in float64.}
\label{fig:2D_DD_SSBroyden_rl2_Surgery}
\end{figure}

From a practical standpoint, it is worth noting that all cases in Table \ref{tab:2D_DD_SSBroyden_rl2_Surgery} that use the SSBroyden optimizer leverage double-precision training (using float64 datatype) for the entire training phase including the Adam optimizer. This remains a basic requirement for all quasi-Newton optimizers, including the Broyden family, to maintain numerical stability during training.

Here, we utilize the SSBroyden optimizer to test its capability in mitigating gradient conflicts. We specifically use it independently as well as in combination with Adam for solving PDEs through domain decomposition. Figure \ref{fig:2D_DD_SSBroyden_rl2_Surgery} shows the convergence of SSBroyden when used along with Adam after an initial warm-up phase and when used independently without any warm-up. While the final state of convergence appears nearly identical for both cases, Table \ref{tab:2D_DD_SSBroyden_rl2_Surgery} provides better insights on their difference in performance. For all experiments in Table \ref{tab:2D_DD_SSBroyden_rl2_Surgery}, it is worth noting that the total number of iterations is fixed (150000) to ensure fair comparison between the different cases. However, for cases with the SSBroyden optimizer, training may stop early if the total loss reaches the minimum threshold tolerance.

{\setlength{\tabcolsep}{3.5pt}
\begin{table}[H]
\centering
\small
\begin{tabular}{|c|c|c|c|c|c||}
\hline
\multirow{2}{*}{\textbf{Gradient Surgery}} & \multirow{2}{*}{\textbf{Optimizer}} & \multicolumn{2}{c|}{\textbf{Relative $L_2$ Error}} & \multirow{2}{*}{\textbf{Time (ms/iter)}}\\ \cline{3-4}
& & $\Omega_1$ & $\Omega_2$ & \\
\hline\hline
None & Adam [150000] & 0.09595 $\pm$ 0.00808 & 0.06512 $\pm$ 0.00371 & 7.47 \\
\hline
ConFIG & Adam [150000] & 0.51062 $\pm$ 0.00466 & 0.61174 $\pm$ 0.00881 & 6.27 \\
\hline
PCGrad & Adam [150000] & 0.06814 $\pm$ 0.00443 & 0.05617 $\pm$ 0.00359 & 8.19 \\
\hline
Norm-PCGrad & Adam [150000] & \textbf{0.01598 $\pm$ 0.00241} & \textbf{0.01947 $\pm$ 0.00916} & 8.11 \\
\hline\hline
None & SSBroyden [150000] & 0.02256 $\pm$ 0.00338 & 0.01598 $\pm$ 0.00176 & 31 \\
\hline
Norm-PCGrad & Adam [10000] + SSBroyden [140000] & 0.01796 $\pm$ 0.00207 & 0.01487 $\pm$ 0.00837 & 30.78 \\ \hline
Norm-PCGrad & Adam [50000] + SSBroyden [100000] & \textbf{0.01632 $\pm$ 0.00673} & \textbf{0.01252 $\pm$ 0.00311} & 30.71 \\ \hline
\end{tabular}
\caption{\textbf{2D Poisson -- Gradient surgery and curvature-aware optimization.} Relative $L_2$ error (mean $\pm$ std. devi) and computational time for gradient surgery and curvature-aware optimization using SSBroyden for domain decomposition.  Gradient surgery is applied only in the Adam phase and not in the SSBroyden phase. All cases with SSBroyden is evaluated in float64. Bold text represents the best results.}
\label{tab:2D_DD_SSBroyden_rl2_Surgery}
\end{table}}

Table~\ref{tab:2D_DD_SSBroyden_rl2_Surgery} shows that SSBroyden performs best when warmed up with Adam rather than used throughout training, achieving lower relative $L_2$ errors at reduced computational cost, as the initial Adam phase brings the parameters near a local optimum before SSBroyden performs precise convergence. Without gradient surgery, SSBroyden alone achieves substantially lower relative $L_2$ error than Adam but incurs 6--7$\times$ higher computational cost due to approximate Hessian computation in double precision, whereas Adam uses single-precision training without curvature information. Thus, curvature-aware optimization can mitigate gradient conflicts, but at significantly higher computational cost.

\begin{figure}[H]
\centering
{
\centering
\includegraphics[width=0.9\linewidth, trim={20mm 70mm 20mm 50mm}, clip]{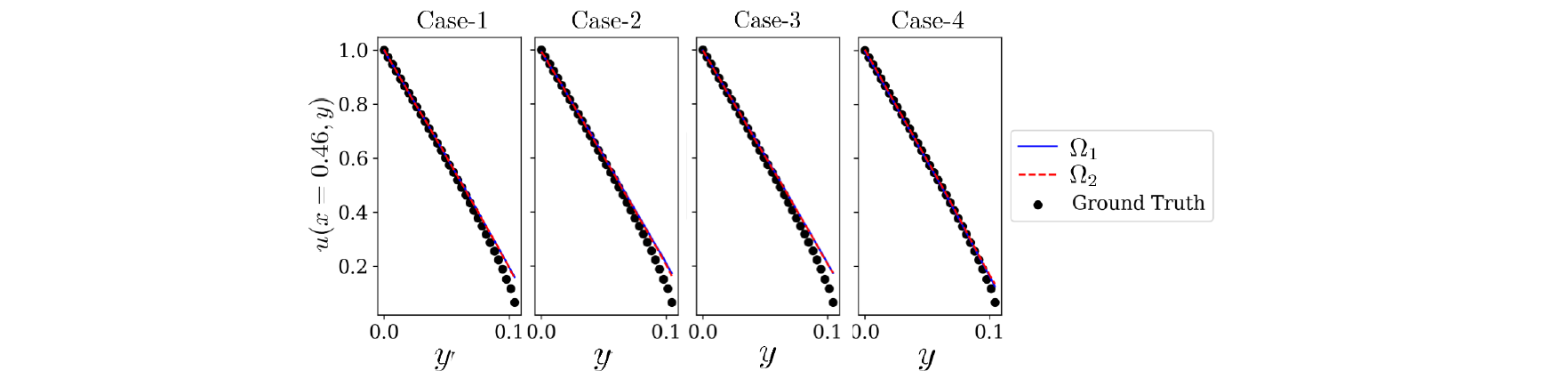}}
\caption{\textbf{2D Poisson -- Interface with curvature-aware optimization.} Solution comparison at the interface ($x = 0.46$) of $\Omega_1$ and $\Omega_2$ using Adam + SSBroyden optimizers. With reference to Table \ref{tab:2D_DD_SSBroyden_rl2_Surgery}: \underline{Case-1}: (Adam + Norm-PCGrad)[50k] + SSBroyden[100k], \underline{Case-2}: (Adam + Norm-PCGrad)[10k] + SSBroyden[140k], \underline{Case-3}: SSBroyden only [150k], \underline{Case-4}: (Adam + Norm-PCGrad)[150k].}
\label{fig:2D_DD_interface_curvatureaware}
\end{figure}

With Norm-PCGrad, Table~\ref{tab:2D_DD_SSBroyden_rl2_Surgery} shows that an Adam warm-up followed by SSBroyden improves convergence, accuracy, and computational efficiency compared with SSBroyden alone; however, no SSBroyden configuration surpasses Adam with Norm-PCGrad. This improvement also extends to the subdomain interface (Figure~\ref{fig:2D_DD_interface_curvatureaware}). Notably, all SSBroyden cases outperform the gradient-surgery methods except Norm-PCGrad, both globally and at the interface (Table~\ref{tab:2D_DD_SSBroyden_rl2_Surgery}, Figures~\ref{fig:2D_DD_interface} and \ref{fig:2D_DD_interface_curvatureaware}). Overall, these results demonstrate that combining gradient surgery with curvature-aware optimization can mitigate gradient conflicts in PINN-based domain decomposition. We apply gradient surgery only to first-order optimizers such as Adam, while SSBroyden operates independently, as combining both substantially increases computational cost and obscures their individual contributions to conflict mitigation.
\subsection{Poisson Equation: 3D Domain \texorpdfstring{$\rightarrow$}{->} Euclidean \texorpdfstring{$+$}{+} Euclidean \texorpdfstring{$+$}{+} Euclidean}
\label{sec:3DDomain-2}
\begin{figure}[H]
\centering
{
\centering
\includegraphics[width=0.9\linewidth, trim={0mm 7.2mm 0mm 6.3mm}, clip]{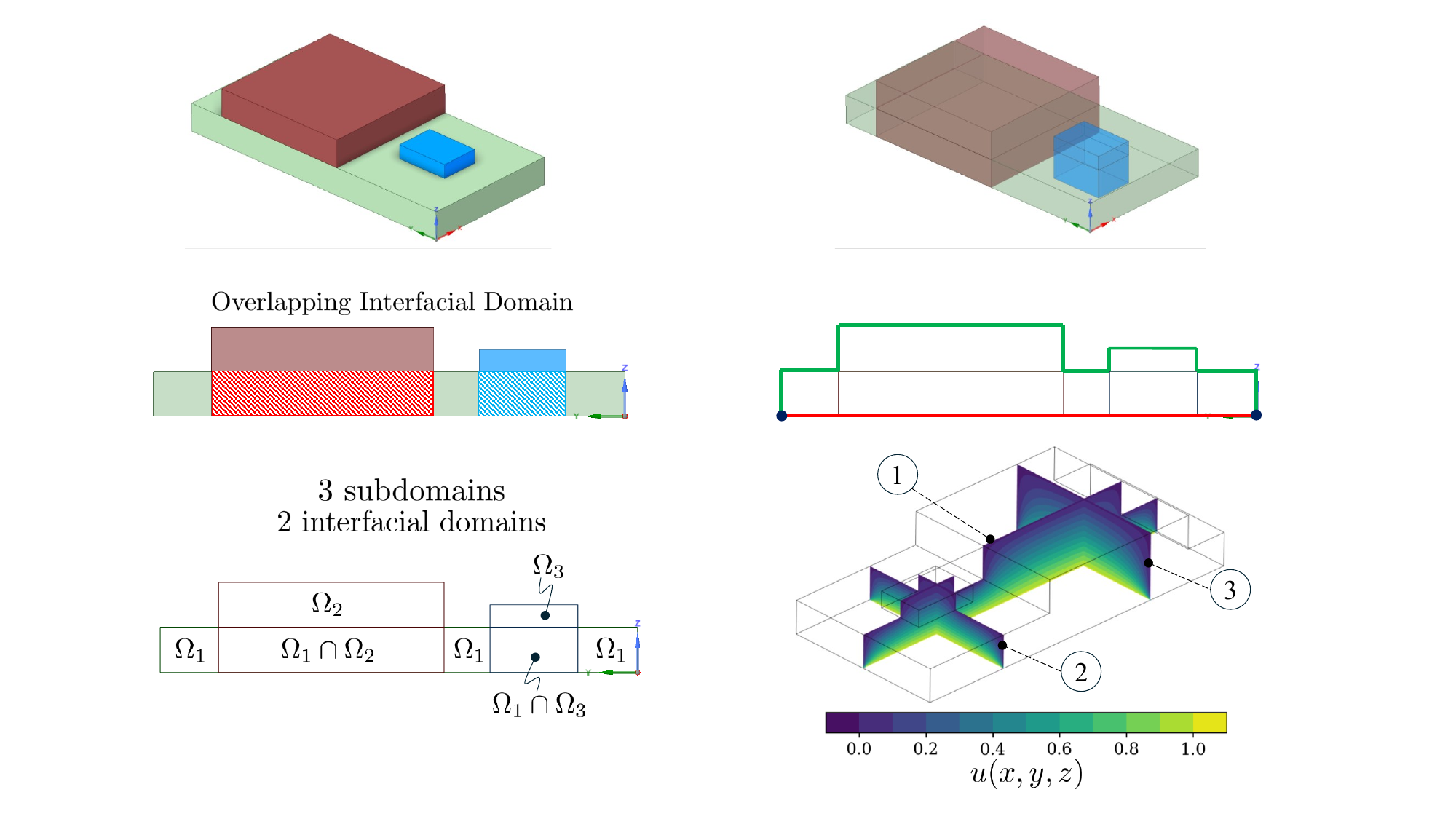}
}
\caption{\textbf{3D Poisson (3 subdomains) -- Domain setup.} Decomposition of a 3D domain into three Euclidean subdomains through multiple overlapping interfacial domains. Interfaces are highlighted in red and blue hatches.}
\label{fig:3D_DD2_domain}
\end{figure}
The domain considered in the current section is decomposed into three subdomains using SPINN, as shown in Figure \ref{fig:3D_DD2_domain}. Owing to the benefits of utilizing an overlapping interfacial domain in combination with SPINN discussed in the previous section, here we decompose the 3D domain through two sets of overlapping interfacial domains as shown in Figure \ref{fig:3D_DD2_domain}.

In the current section, we begin by comparing the different algorithms for gradient surgery, including the proposed Norm-PCGrad, to overcome \textit{gradient conflicts} while decomposing a 3D domain into three Euclidean subdomains. We also include curvature-aware optimization using the Broyden family of optimizers to further mitigate gradient conflicts for domain decomposition of the 3D domain shown in Figure \ref{fig:3D_DD2_domain}. Furthermore, we extend the proposed Norm-PCGrad for solving PDEs through domain decomposition of a 3D domain using Kolmogorov-Arnold Networks (KANs) \cite{liu2025kan}. We note that the literature lacks clarity on the scalability of domain decomposition using KANs to 3D domains, and the results in the current section provide valuable insights in this direction.

\subsubsection{Gradient Surgery to tackle Gradient Conflicts}

\begin{figure}
\centering
{
\centering
\includegraphics[width=0.85\linewidth, trim={15mm 46mm 15mm 45mm}, clip]{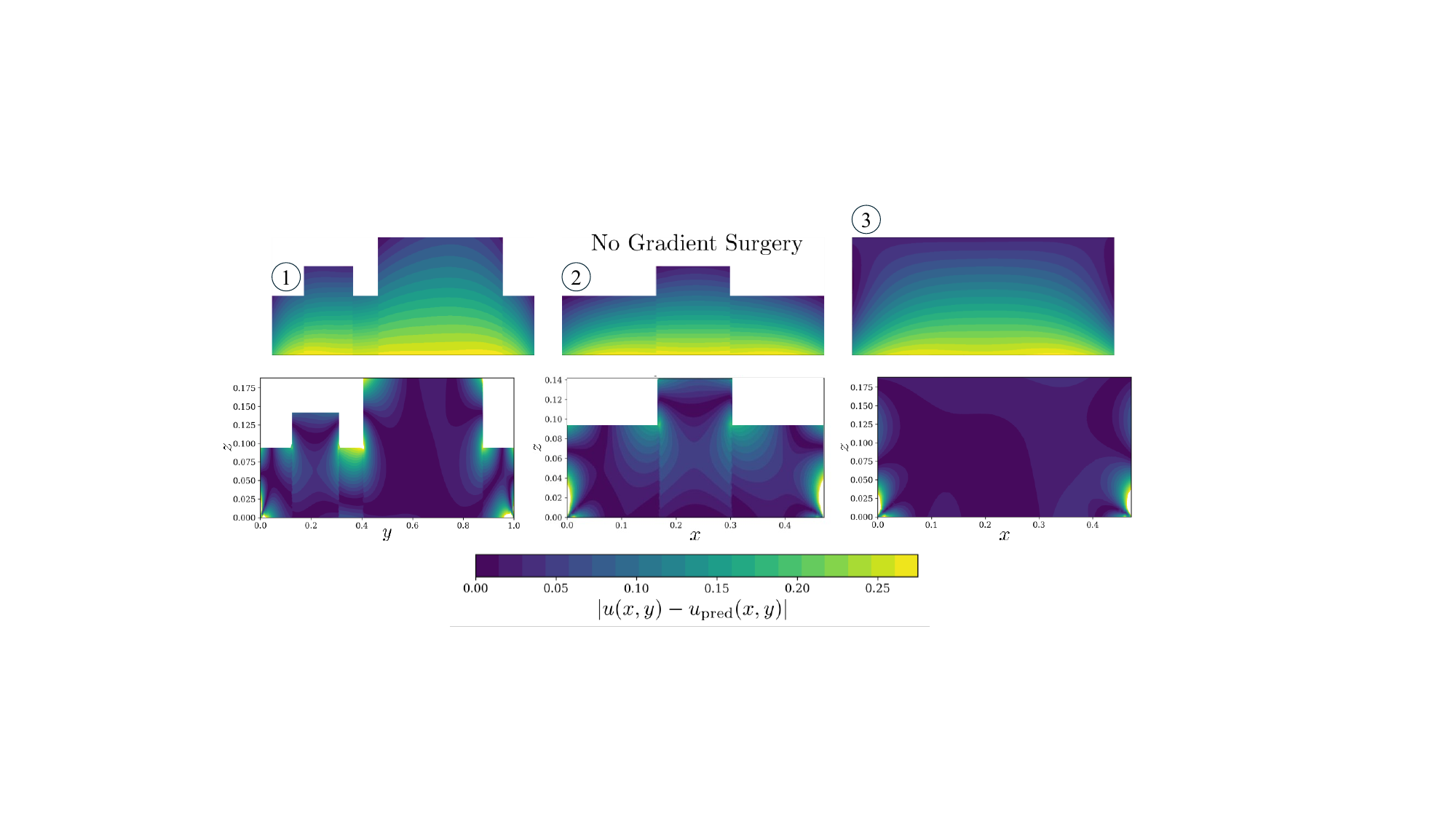}
}
\caption{\textbf{3D Poisson (3 subdomains) -- No gradient surgery.} Solution to the Poisson equation through 3D domain decomposition without gradient surgery. Sections correspond to the slices shown in Figure \ref{fig:3D_DD2_domain}.}
\label{fig:3D_DD2_nosurg}
\end{figure}

\begin{figure}
\centering
{
\centering
\includegraphics[width=0.85\linewidth, trim={15mm 46mm 15mm 45mm}, clip]{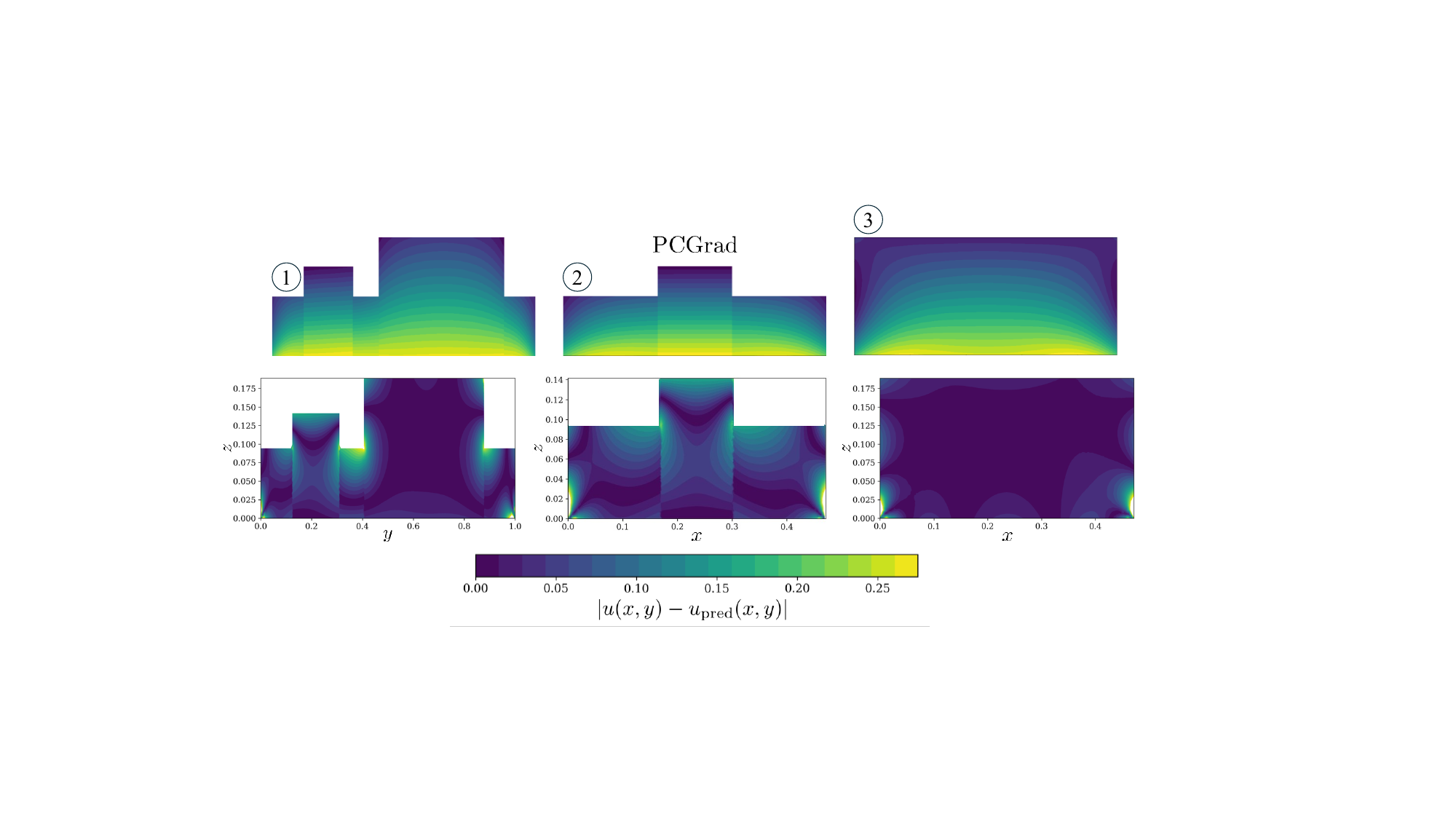}
}
\caption{\textbf{3D Poisson (3 subdomains) -- PCGrad.} Solution to the Poisson equation through 3D domain decomposition using PCGrad. Sections correspond to the slices shown in Figure \ref{fig:3D_DD2_domain}.}
\label{fig:3D_DD2_pcgrad}
\end{figure}

\begin{figure}
\centering
{
\centering
\includegraphics[width=0.85\linewidth, trim={15mm 46mm 15mm 45mm}, clip]{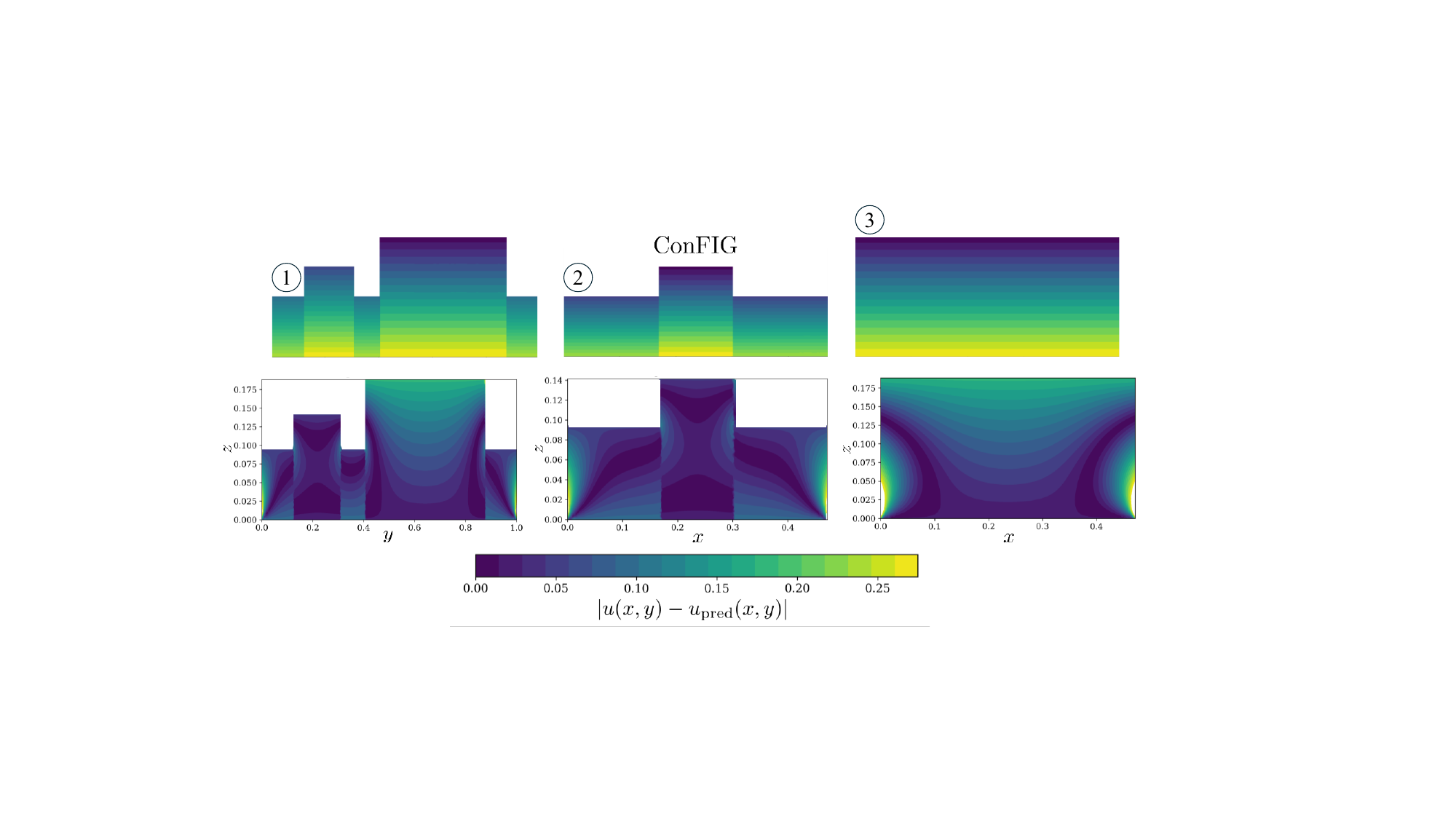}
}
\caption{\textbf{3D Poisson (3 subdomains) -- ConFIG.} Solution to the Poisson equation through 3D domain decomposition using ConFIG. Sections correspond to the slices shown in Figure \ref{fig:3D_DD2_domain}.}
\label{fig:3D_DD2_config}
\end{figure}

\begin{figure}
\centering
{
\centering
\includegraphics[width=0.85\linewidth, trim={15mm 46mm 15mm 45mm}, clip]{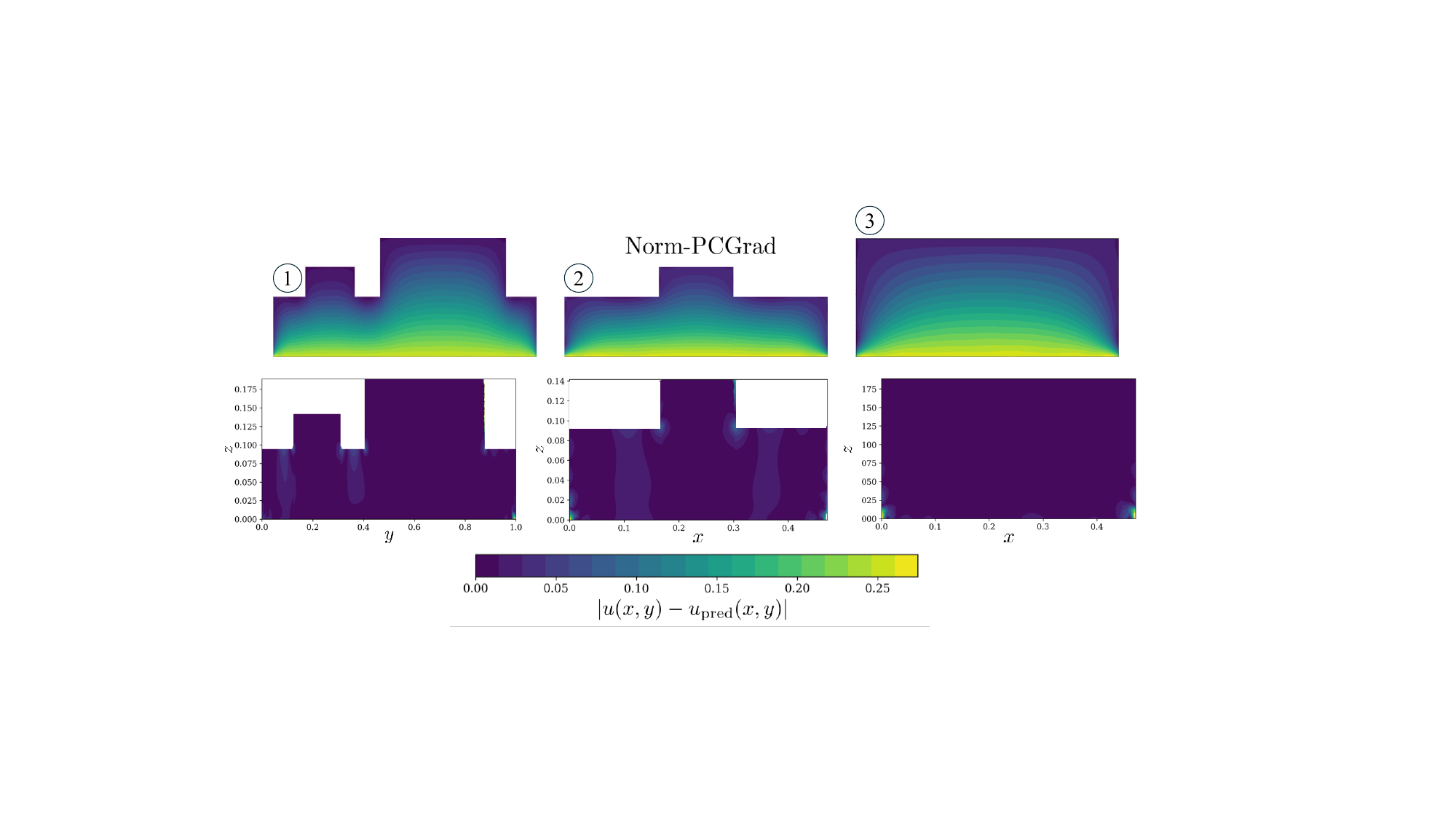}
}
\caption{\textbf{3D Poisson (3 subdomains) -- Norm-PCGrad.} Solution to the Poisson equation through 3D domain decomposition using Norm-PCGrad. Sections correspond to the slices shown in Figure \ref{fig:3D_DD2_domain}.}
\label{fig:3D_DD2_normpcgrad}
\end{figure}

To evaluate the performance of gradient surgery in mitigating conflicting gradients while solving a PDE through 3D domain decomposition, we consider three distinct slices of the 3D domain as shown in Figure \ref{fig:3D_DD2_domain}. We separately predict and evaluate the error incurred by each of the algorithms for gradient surgery in each of these slices as a representative measure of its overall performance in the entire 3D domain. Similar to previous discussions, we can evaluate these based on the (a) global solution field, (b) direction of propagation, (c) interface continuity and (d) the accuracy at the corner regions. To this end, Figure \ref{fig:3D_DD2_nosurg}-\ref{fig:3D_DD2_normpcgrad} shows the solutions to the Poisson equation in the 3D domain (in the Top Row) and the point-wise absolute error (in the Bottom Row).

Compared to the the solution when no gradient surgery is applied (Figure \ref{fig:3D_DD2_nosurg}), we observe a very negligible improvement in the accuracy when PCGrad is utilized (Figure \ref{fig:3D_DD2_pcgrad}). Particularly, we observe that PCGrad struggles with maintaining the continuity across the different interfaces which is evident from the spiking up of the point-wise absolute errors closer to the interface region in Figure \ref{fig:3D_DD2_pcgrad}. However, we observe the least error in the slice \circled{3} with no interfaces and the solution accuracy degrades only at the corner regions. It is also worth noting that, the subdomain, $\Omega_3$ present in slices \circled{1} and \circled{2} has the maximum error and almost all the cases struggle with their accuracy in this region. 

The performance of ConFIG is in alignment with the previous sections: it struggles when it comes to predicting the solution across the whole domain including the interface regions as well as at the corners. While a clear reason for the struggles of these methods are unknown it is worth noting that domain decomposition specifically while dealing with multiple loss terms for PDE residue, boundary conditions and interface conditions are prone to conflicting gradients even more than a regular case that solves a PDE in the whole domain without any interface conditions. There is an inherent coupling across the subdomains within the domain decomposition problem that is facilitated by the interface loss that constitute a PDE loss and a data continuity (or the average) loss at the interface location. Typically, in the context of PINN, these subdomains are assigned with independent models that minimize the same interface loss in two different loss landscapes while maintaining the inherent coupling between them to ensure the interface continuity. To this end, we note that gradient surgery algorithms which are mostly based on gradient projections can at times tamper with the direction of the interface gradients for the two subdomains independently without respecting the inherent coupling between these gradients for the interface loss. This can decouple the gradients of the interface loss function during training, thereby leading to a discontinuity at the interface.

\begin{figure}[H]
\centering
{
\centering
\includegraphics[width=0.76\linewidth, trim={0mm 15mm 0mm 10mm}, clip]{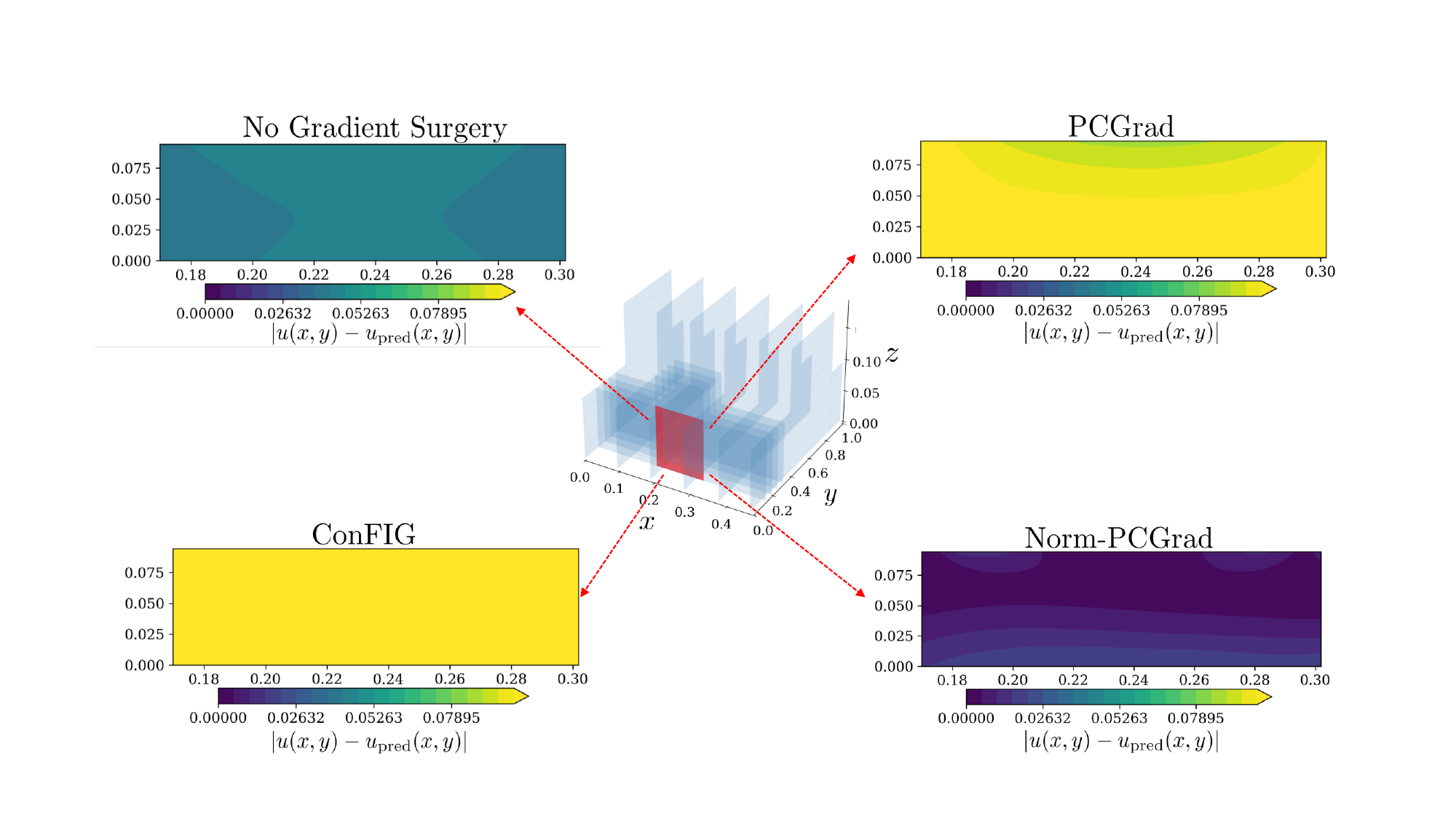}
}
\caption{\textbf{3D Poisson (3 subdomains) -- Interface comparison.} Solution comparison at the interface for 3D domain decomposition. The interface slice is located in the $xz$-plane highlighted in red.  The highlighted red region is a finite slice extracted from the overlapping interfacial domain between the subdomains - $\Omega_1$ and $\Omega_3$.}
\label{fig:3D_DD2_interface}
\end{figure}

Here, we note that Norm-PCGrad overcomes the struggle faced by the other cases and predicts a smooth and continuous solution across all the slices as shown in Figure \ref{fig:3D_DD2_normpcgrad}. Furthermore, we note that the absolute error for Norm-PCGrad is an order of magnitude lower than PCGrad and two orders of magnitude lower than ConFIG. Similar to the previous sections, Norm-PCGrad attains a close-to perfect continuity at the interfaces between the subdomains with only minimal concentration of errors residing at the corner regions of the domain. Unlike PCGrad, we observe a consistent performance for Norm-PCGrad across different domains irrespective of their dimensionality, number of subdomains/interfaces and domain size. This further confirms the potential of Norm-PCGrad as a potential candidate for gradient surgery to mitigate \textit{gradient conflicts} for 3D domain decomposition over the existing algorithms. Based on the results in Figure \ref{fig:3D_DD2_nosurg}-\ref{fig:3D_DD2_normpcgrad}, we observe that the algorithms struggle the most around the subdomain, $\Omega_3$. To this end, we also monitor the continuity at the interface as shown by Figure \ref{fig:3D_DD2_interface}. To this end, we note that Norm-PCGrad attains the smallest interface error, which is two orders of magnitude lower than the error obtained for PCGrad and ConFIG. This further supports the results observed in the global solution field spanning the whole domain in Figure \ref{fig:3D_DD2_nosurg}-\ref{fig:3D_DD2_normpcgrad}. 

\begin{figure}
\centering
{
\centering
\includegraphics[width=\linewidth, trim={0mm 50mm 0mm 50mm}, clip]{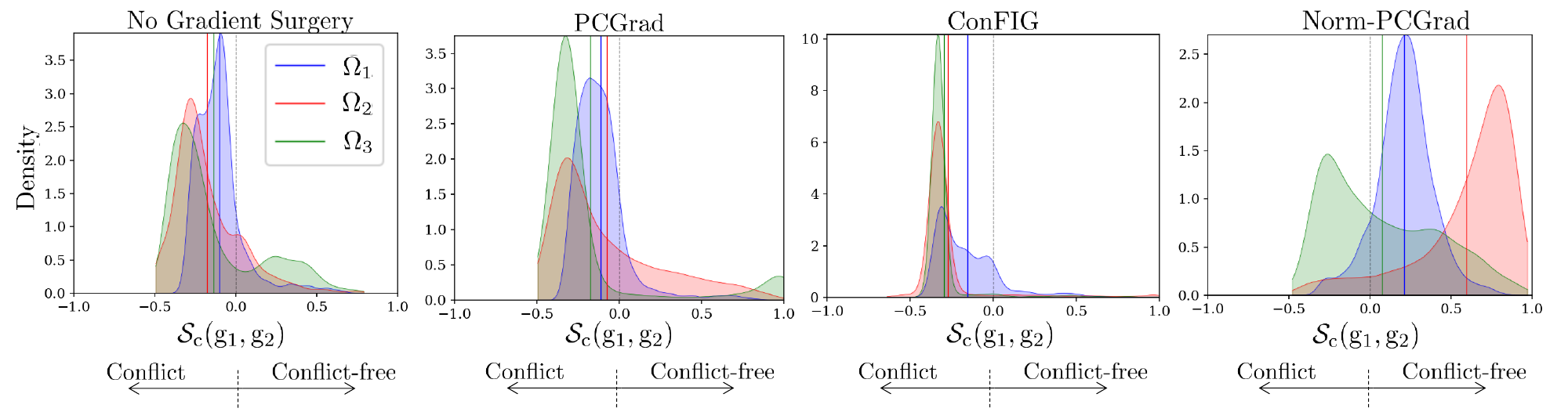}}
\caption{\textbf{3D Poisson (3 subdomains) -- Cosine Similarity.} Density distribution of cosine similarity scores for the three subdomains, $\mathcal{S}_c^{\Omega_1}$, $\mathcal{S}_c^{\Omega_2}$ and $\mathcal{S}_c^{\Omega_3}$. Vertical solid lines represent the mean of each distribution $\mathcal{S}_{c,~\text{mean}}^{\Omega_1}$, $\mathcal{S}_{c,~\text{mean}}^{\Omega_2}$ and $\mathcal{S}_{c,~\text{mean}}^{\Omega_3}$, while the dashed line marks the zero-line where gradients are orthogonal. If ($\mathcal{S}_{c,~\text{mean}}^{\Omega_1}$/$\mathcal{S}_{c,~\text{mean}}^{\Omega_2}$/$\mathcal{S}_{c,~\text{mean}}^{\Omega_3}) < 0$, the gradients were in conflict for the majority of training in $\Omega_1$/$\Omega_2$/$\Omega_3$. On the other hand, if ($\mathcal{S}_{c,~\text{mean}}^{\Omega_1}$/$\mathcal{S}_{c,~\text{mean}}^{\Omega_2}$/$\mathcal{S}_{c,~\text{mean}}^{\Omega_3}) > 0$, the gradients were conflict-free for the majority of training in $\Omega_1$/$\Omega_2$/$\Omega_3$. The extent of gradient conflicts (or their suppression) is measured by the distance of $\mathcal{S}_{c,~\text{mean}}^{\Omega_1}$, $\mathcal{S}_{c,~\text{mean}}^{\Omega_2}$ and $\mathcal{S}_{c,~\text{mean}}^{\Omega_3}$ from the zero-line. Here, the cosine similarity scores between gradient pairs are computed at fixed intervals during training and their distribution is represented as the density distribution.}
\label{fig:3D_DD2_cosine_similarity}
\end{figure}

While solving the 3D Poisson equation using SPINN-SPINN domain decomposition for the domain shown in Figure \ref{fig:3D_DD2_domain}, we compare the gradient conflicts across different algorithms of gradient surgery using the cosine similarity score, as shown in Figure \ref{fig:3D_DD2_cosine_similarity}. Without any gradient surgery, we observe that $\mathcal{S}_{c,~\text{mean}}^{\Omega_1}$, $\mathcal{S}_{c,~\text{mean}}^{\Omega_2}$, and $\mathcal{S}_{c,~\text{mean}}^{\Omega_3} < 0$, indicating conflicting gradients in all three subdomains. With PCGrad, we observe negligible improvement since the gradients were still conflicting for the majority of training. ConFIG is clearly counterproductive, with $\mathcal{S}_{c,~\text{mean}}^{\Omega_2}$ and $\mathcal{S}_{c,~\text{mean}}^{\Omega_3} \ll 0$ and $\mathcal{S}_{c,~\text{mean}}^{\Omega_1} < 0$. We note that Norm-PCGrad attains stronger gradient alignment for the majority of training, with $\mathcal{S}_{c,~\text{mean}}^{\Omega_1}$ and $\mathcal{S}_{c,~\text{mean}}^{\Omega_2} \gg 0$ and $\mathcal{S}_{c,~\text{mean}}^{\Omega_3} > 0$, substantially exceeding the baseline case without any gradient surgery. These observations further support the trend of relative $L_2$ error for SPINN-SPINN domain decomposition shown in Table \ref{tab:3D_DD2_SSBroyden_rl2_Surgery}.

\subsubsection{Gradient Surgery + Curvature-Aware Optimization to tackle Gradient Conflicts}

Here, we tackle conflict for domain decomposition by leveraging the gradient surgery via gradient projections in combination with quasi-Newton methods which offer a middle ground between the first-order optimizers such as Adam and the Newton-based second-order optimizers. Specifically, we use the Norm-PCGrad for gradient surgery along with Adam in the first phase of training and the SSBroyden optimizer in the second phase after the warm-up to showcase mitigation of \textit{gradient conflicts} while performing domain decomposition.

\begin{figure}[H]
\centering
\begin{subfigure}[b]{0.48\linewidth}
    \centering
    \includegraphics[width=\linewidth, trim={0mm 5mm 0mm 0mm}, clip]{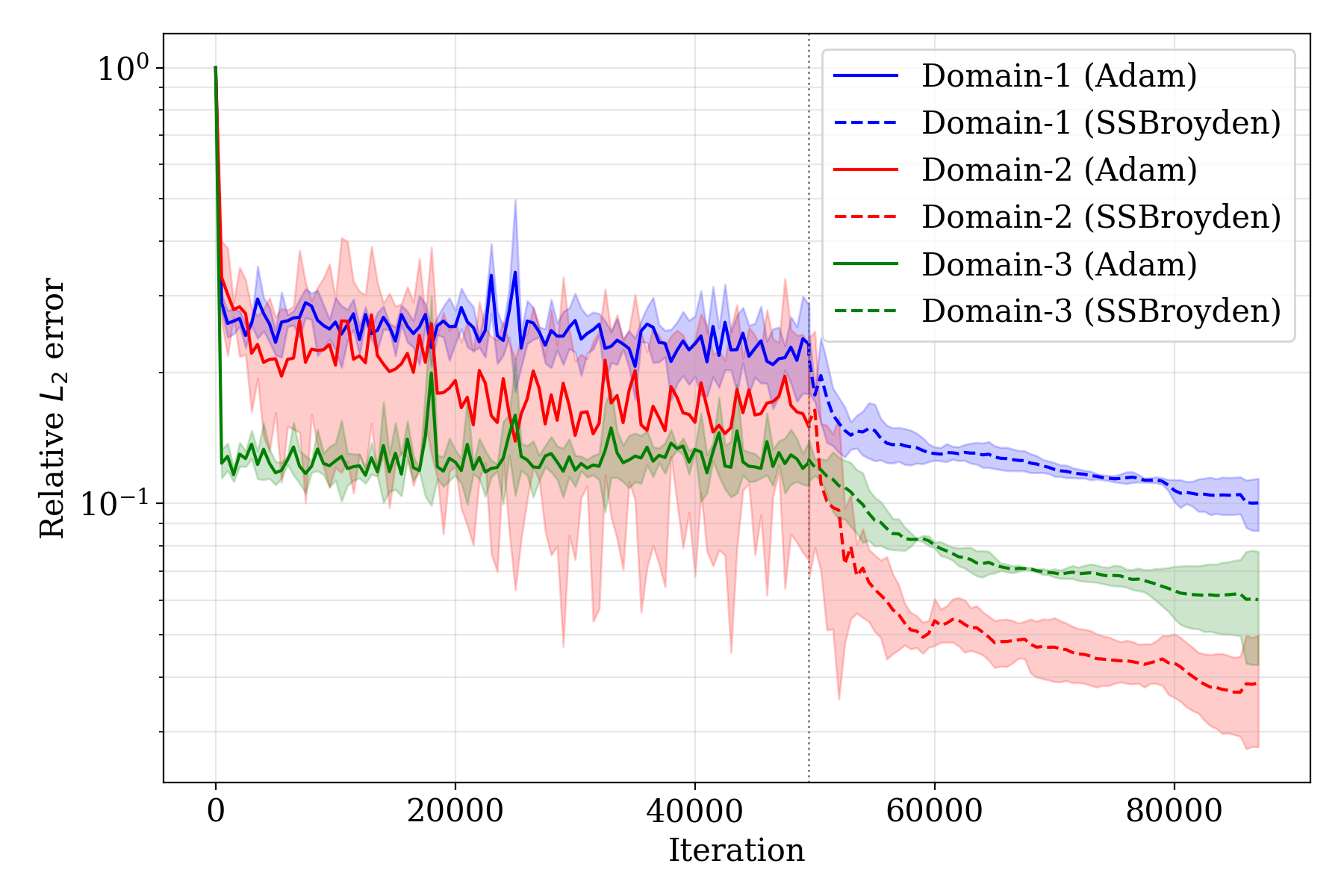}
    \label{fig:3D_DD2_SSBroyden_rl2_left}
\end{subfigure}
\hfill
\begin{subfigure}[b]{0.48\linewidth}
    \centering
    \includegraphics[width=\linewidth, trim={0mm 5mm 0mm 0mm}, clip]{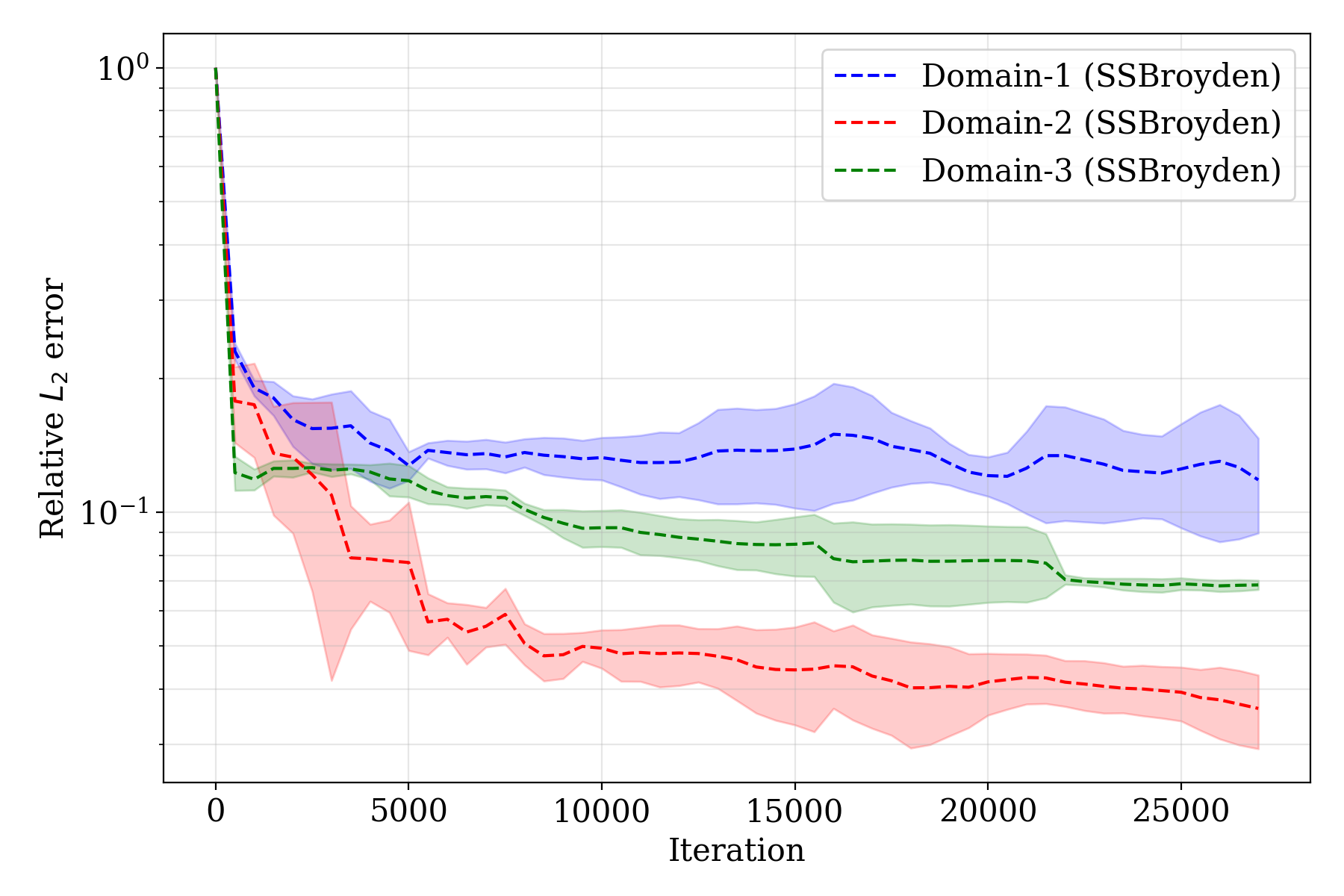}
    \label{fig:3D_DD2_SSBroyden_rl2_right}
\end{subfigure}
\caption{\textbf{3D Poisson (3 subdomains) -- SSBroyden convergence.} Convergence of relative $L_2$ error for 3D domain decomposition using double-precision training with Adam + SSBroyden and Norm-PCGrad.  Gradient surgery is applied only in the Adam phase and not in the SSBroyden phase. Also, the vertical line for Adam + SSBroyden case represents switching from Adam to SSBroyden during training. All cases with SSBroyden is evaluated in float64.}
\label{fig:3D_DD2_SSBroyden_rl2_Surgery}
\end{figure}

In the context of domain decomposition of the 3D domain shown in Figure \ref{fig:3D_DD2_domain}, the training process is integrated with the SSBroyden optimizer in one of the following two ways$-$ either in combination with the Adam optimizer equipped with gradient surgery using Norm-PCGrad or independently as the only optimizer for training. The convergence of the SSBroyden optimizer using the aforementioned two ways are shown in Figure \ref{fig:3D_DD2_SSBroyden_rl2_Surgery} separated into the three subdomains. We observe a clear trend for convergence in all the three subdomains where the $\Omega_3$ attains the least relative $L_2$ error and $\Omega_1$ attains the largest error. It is sensible for $\Omega_1$ to have the largest error among the three subdomains since, it interacts with both $\Omega_2$ and $\Omega_3$ as observed in Figure \ref{fig:3D_DD2_domain}, thereby incurring interface errors from both the subdomains.

{\setlength{\tabcolsep}{2.5pt}
\begin{table}[H]
\centering
\footnotesize
\begin{tabular}{|c|c|c|c|c|c||}
\hline
\multirow{2}{*}{\textbf{Gradient Surgery}} & \multirow{2}{*}{\textbf{Optimizer}} & \multicolumn{3}{c|}{\textbf{Relative $L_2$ Error}} & \multirow{2}{*}{\textbf{Time (ms/iter)}}\\ \cline{3-5}
& & $\Omega_1$ & $\Omega_2$ & $\Omega_3$ & \\
\hline\hline
None & Adam [150000] & 0.1801 $\pm$ 0.0068 & 0.0785 $\pm$ 0.0034 & 0.1030 $\pm$ 0.0051 & 27.29 \\
\hline
ConFIG & Adam [150000] & 0.5564 $\pm$ 0.0203 & 0.5209 $\pm$ 0.0582 & 0.6300 $\pm$ 0.0700 & 27.14 \\
\hline
PCGrad & Adam [150000] &  0.1467 $\pm$ 0.0426 & 0.0718 $\pm$ 0.0868 & 0.1336 $\pm$ 0.0068 & 28.73 \\
\hline
Norm-PCGrad & Adam [150000] & \textbf{0.0749 $\pm$ 0.0127} & \textbf{0.0196 $\pm$ 0.0150} & \textbf{0.0796 $\pm$ 0.1065} & 25.50 \\
\hline\hline
None & SSBroyden [150000] & 0.1182 $\pm$ 0.0285 & 0.0362 $\pm$ 0.0068 & 0.0686 $\pm$ 0.0016 & 102.63 \\ \hline
Norm-PCGrad & Adam [50000] + SSBroyden [100000] & \textbf{0.1048 $\pm$ 0.0101} & \textbf{0.0369 $\pm$ 0.0076} & \textbf{0.0620 $\pm$ 0.0123} & 138 \\ 
\hline
\end{tabular}
\caption{\textbf{3D Poisson -- Gradient surgery and curvature-aware optimization.} Relative $L_2$ error (mean $\pm$ std. devi) and computational time for gradient surgery and curvature-aware optimization using SSBroyden for 3D domain decomposition.  Gradient surgery is applied only in the Adam phase and not in the SSBroyden phase. All cases with SSBroyden is evaluated in float64. Bold text represents the best results.}
\label{tab:3D_DD2_SSBroyden_rl2_Surgery}
\end{table}}

Here, Table \ref{tab:3D_DD2_SSBroyden_rl2_Surgery} describes the comparison between various algorithms for gradient surgery in combination with the choice of optimizer utilized for domain decomposition of the 3D domain shown in Figure \ref{fig:3D_DD2_domain}. As previously described, the curvature-aware optimization using the SSBroyden optimizer is considered as an independent pillar by itself to tackle \textit{gradient conflicts} and therefore treated independently without any gradient surgery even when used in combination with the Adam optimizer.

Based on this broad comparison shown in Table \ref{tab:3D_DD2_SSBroyden_rl2_Surgery}, we can clearly conclude that Norm-PCGrad with the Adam optimizer offers the best performance in terms of accuracy and computational efficiency. It is also, worth noting that SSBroyden when used by itself or in combination with Adam and Norm-PCGrad offer comparable performance at much higher computational costs. This computational bottleneck makes SSBroyden a temporary resort for mitigating \textit{gradient conflicts}. On the other hand, PCGrad with Adam optimizer offers a reasonable performance gain comparable to SSBroyden for both $\Omega_1$ and $\Omega_2$, at a fraction of cost as shown in Table \ref{tab:3D_DD2_SSBroyden_rl2_Surgery}. Interestingly, we observe a similar trend for the 3D domain decomposition as for the 2D domains $-$ the computational costs for all the algorithms for gradient surgery in combination with Adam are comparable. This is a reasonable since SSBroyden utilizes double precision training while Adam relies on single precision training to achieve a comparable performance.

\subsubsection{Kolmogorov-Arnold Networks}

Based on the results from all the previous discussions, we observe clear advantages while utilizing the Adam optimizer integrated with the proposed, Norm-PCGrad algorithm for mitigation of the \textit{gradient conflicts} through domain decomposition. To this end, we leverage the aforementioned combination to solve the Poisson equation in a 3D domain shown in Figure \ref{fig:3D_DD2_domain} using the Kolmogorov-Arnold Networks. While the spline basis is an obvious choice for the architecture, it encounters a computational bottleneck when scaling the architecture to high-dimensional PDEs. Instead, for the current study, we utilize two of most efficient formulations of basis functions which are based on Chebyshev polynomials \cite{ss2024chebyshev} and the radial basis functions (RBF) \cite{abueidda2025deepokan}. Specifically, we utilize the recently proposed Feature-Enriched KAN \cite{menon2026fekan} (FEKAN) architecture that was demonstrated to showcase desirable characteristics such as improved stability for the Chebyshev basis, faster convergence rate and extreme parametric efficiency for better scalability of the architecture. Furthermore, in the context of solving PDEs, PI-FEKAN was demonstrated to showcase orders of magnitude improvement in the accuracy at minimal changes in the computational cost compared to PIKAN \cite{shukla2024comprehensive}.

{\setlength{\tabcolsep}{3.5pt}
\begin{table}[H]
\centering
\small
\begin{tabular}{|c|c|c||c|c|c||c||}
\hline
\multirow{2}{*}{\textbf{Basis}} & \multirow{2}{*}{\textbf{Architecture}} & \multirow{2}{*}{\textbf{GS}} & \multicolumn{3}{c||}{\textbf{Relative $L_2$ Error}} & \multirow{2}{*}{\textbf{Time (ms/iter)}}\\
\cline{4-6}
& & & $\Omega_1$ & $\Omega_2$ & $\Omega_3$ & \\
\hline\hline
\multirow{2}{*}{Chebyshev} & \multirow{2}{*}{$[5, 13, 13, 10] \times (3)$} & \xmark & 0.1935 $\pm$ 0.0077 & 0.10240 $\pm$ 0.0079 & 0.0979 $\pm$ 0.0014 & 67.64 \\ \cline{3-7}
& & \cmark & \textbf{0.0855 $\pm$ 0.0529} & \textbf{0.0290 $\pm$ 0.0183} & \textbf{0.0511 $\pm$ 0.0295} & 71.01 \\ 
\hline\hline
\multirow{2}{*}{RBF} & \multirow{2}{*}{$[5, 7, 7, 10] \times (3)$} & \xmark & 0.2010 $\pm$ 0.0063 & 0.1145 $\pm$ 0.0005 & \textbf{0.1013 $\pm$ 0.0046} & 63.45 \\ \cline{3-7}
& & \cmark & \textbf{0.0874 $\pm$ 0.0176} & \textbf{0.0367 $\pm$ 0.0189} & 0.1806 $\pm$ 0.0290 & 70.15 \\ 
\hline
\end{tabular}
\caption{\textbf{3D Poisson -- PI-FEKAN with gradient surgery.} Relative $L_2$ error (mean $\pm$ std. devi) and computational time for PI-FEKAN using Chebyshev ($k=2$) and RBF ($d=5$) basis functions for 3D domain decomposition.  GS denotes gradient surgery using Norm-PCGrad. The architecture uses 3-body networks with 5 Fourier feature enrichment terms. Parameter counts are comparable: 2400 (Chebyshev) and 2454 (RBF). Bold text represents the best results.}
\label{tab:3D_DD2_fekan}
\end{table}}

\begin{figure}[H]
\centering
{
\centering
\includegraphics[width=0.9\linewidth, trim={15mm 39mm 15mm 37mm}, clip]{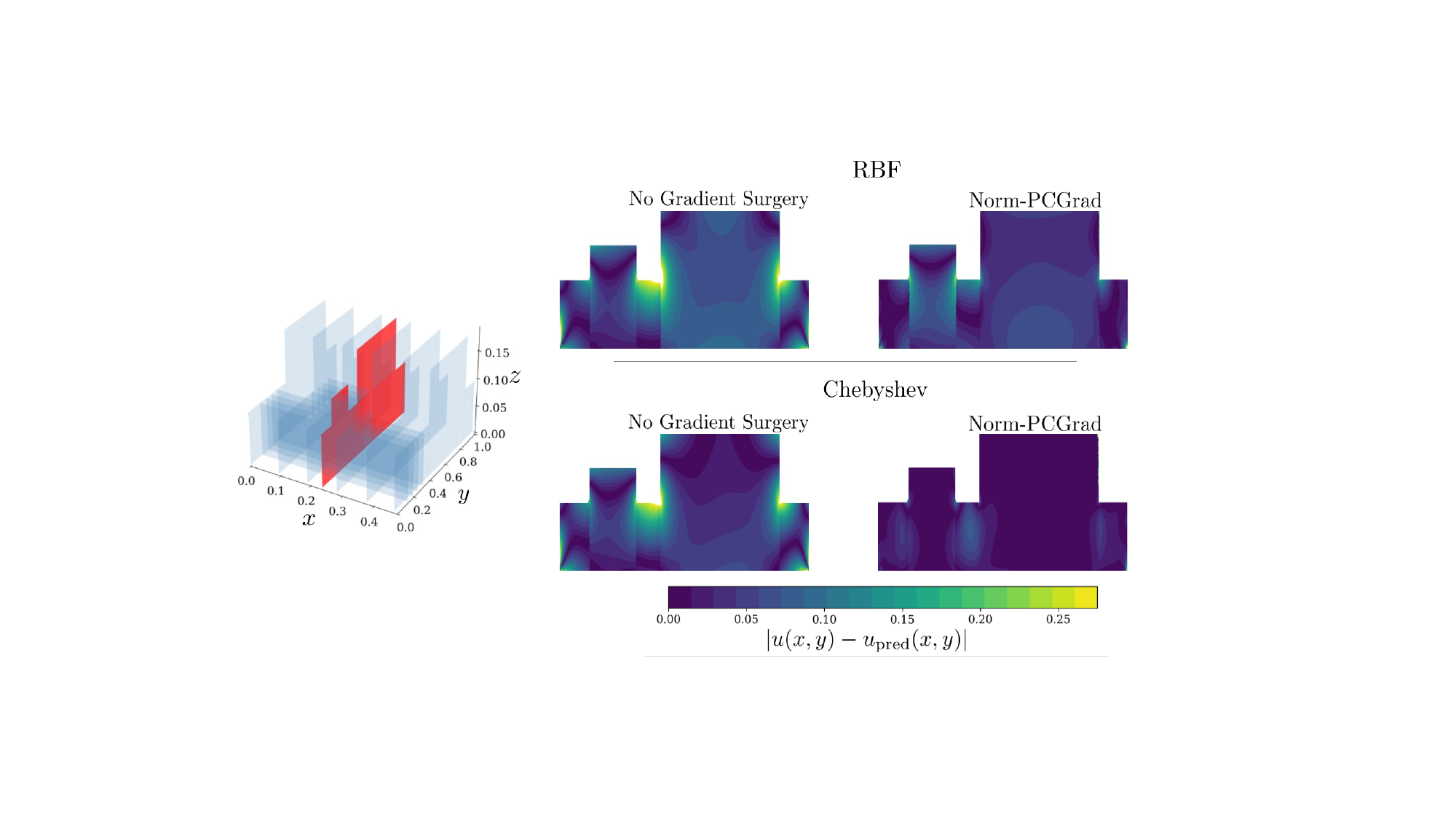}
}
\caption{\textbf{3D Poisson (3 subdomains) -- PI-FEKAN comparison.} Solution to the Poisson equation using 3D domain decomposition (Left) without gradient surgery and (Right) with Norm-PCGrad, for (Top) RBF and (Bottom) Chebyshev basis functions. The 2D slice is located in the $yz$-plane highlighted in red.}
\label{fig:3D_DD2_fekan}
\end{figure}

Unlike MLPs, the KAN architecture has both internal and external parameterizations which can vastly change across different basis functions. To this end in Table \ref{tab:3D_DD2_fekan} we note that, although the number of knots in the architecture changes for the Chebyshev basis function and RBF, the total number of parameters is almost comparable for their fair comparison of performance. Figure \ref{fig:3D_DD2_fekan} provides the comparison between the solutions to the 3D domain decomposition for the different basis functions. With reference to Figure \ref{fig:3D_DD2_domain}, slice \circled{1} is chosen for comparison between the two basis functions utilized by the PI-FEKAN architecture. 

Irrespective of the chosen basis function, we observe an order of magnitude reduction in the relative $L_2$ error as shown in Table \ref{tab:3D_DD2_fekan} at minimal increase in the computational cost and this is also evident from Figure \ref{fig:3D_DD2_fekan}. However, a minor exception is with respect to $\Omega_3$ for RBF where the error increases after the application of gradient surgery while all the other subdomains undergo substantial reduction in the relative $L_2$ error. Similar to the previous discussions, we observe a concentration of the error at the corner regions of the domain and this substantially reduces when gradient surgery was utilized to overcome the \textit{gradient conflicts} for the 3D domain decomposition. Except for the Chebyshev basis function equipped with Norm-PCGrad, there is a higher concentration of the error at the interface for all the other cases. Overall, we can conclude that the Chebyshev basis function showcases the best performance compared to RBF when the total number of parameters is comparable, with or without gradient surgery.

It is also worth noting that, the performance gain in accuracy is substantial for the Chebyshev basis function, since the relative $L_2$ error reduces up to $3\times$ with Norm-PCGrad. The relative $L_2$ error drops up to $\approx2\times$ for RBF with Norm-PCGrad, though not as much. This further confirms the existence of \textit{gradient conflicts} in the context of 3D domain decomposition even while using Kolmogorov-Arnold Networks and leveraging an appropriate method for gradient surgery such as the proposed Norm-PCGrad can substantially lead to performance gains as shown by the set of current results.

\subsection{Sine-Gordon Equation: 3D Domain \texorpdfstring{$\rightarrow$}{->} Euclidean \texorpdfstring{$+$}{+} Euclidean}
\label{sec:3DDomain-3}

We consider a steady-state, nonlinear sine-Gordon equation in a spatial domain $\Omega$, given by
\begin{equation}
    \underbrace{-\Delta u(\mathbf{x})}_{\text{diffusion term}}
    \;+\;
    \underbrace{\sin\!\left(u(\mathbf{x})\right)}_{\text{non-linear term}}
    \;=\;
    \underbrace{f(\mathbf{x})}_{\text{forcing term}},
    \qquad \mathbf{x} \in \Omega,
    \label{eq:sine_gordon}
\end{equation}
subject to appropriate boundary conditions on $\partial\Omega$, where $f$ is the forcing term.

\begin{figure}[H]
\centering
{
\centering
\includegraphics[width=0.7\linewidth, trim={15mm 35mm 15mm 40mm}, clip]{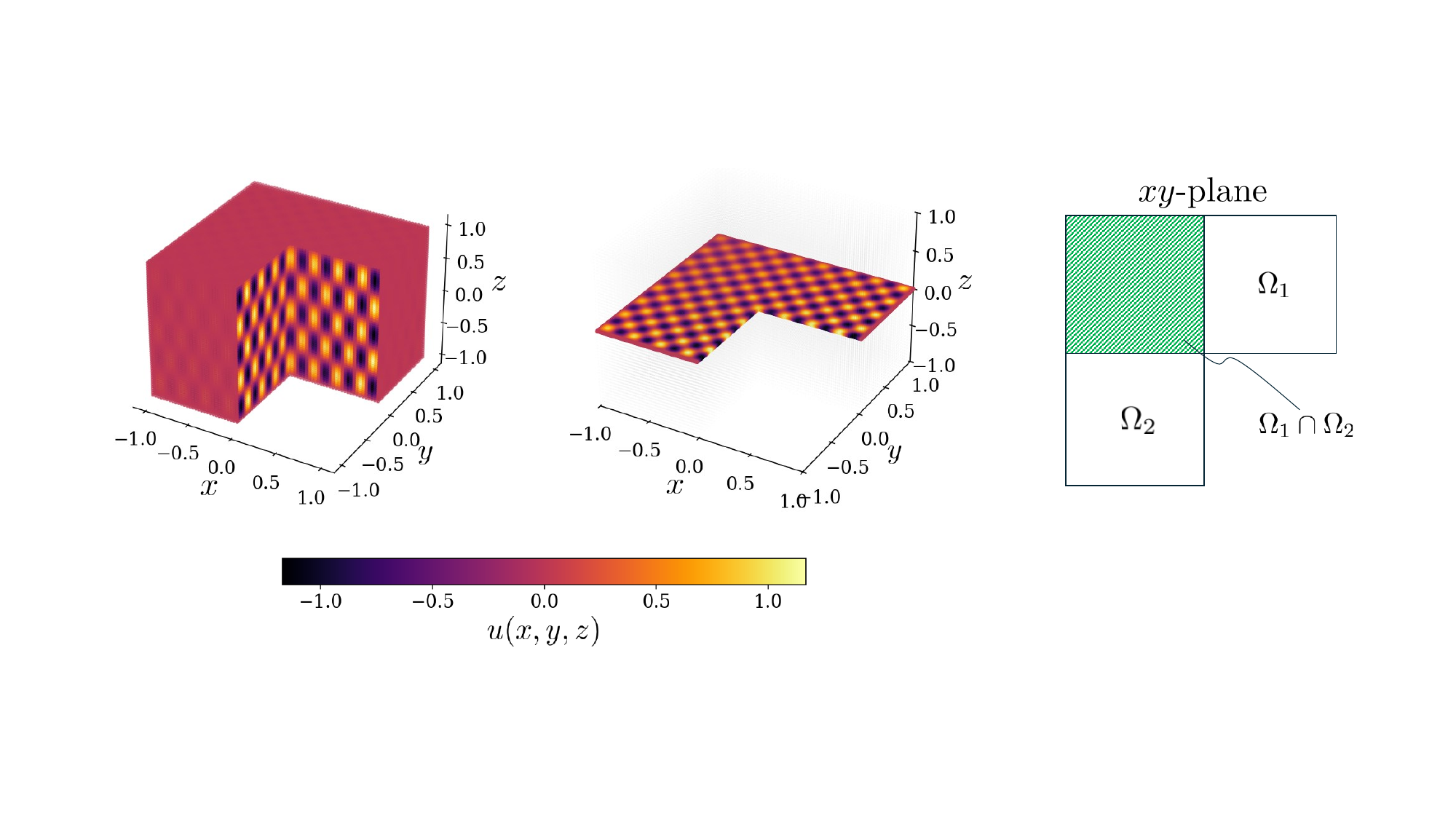}
}
\caption{\textbf{3D sine-Gordon -- Domain setup.} Decomposition of a 3D domain into Euclidean subdomains through an overlapping interfacial domain. The interface is highlighted in green hatches.}
\label{fig:3D_NonLinear}
\end{figure}
\noindent
For the sine-Gordon equation given in Equation~\eqref{eq:sine_gordon}, we prescribe the manufactured solution
\begin{equation}
    u_{\text{exact}}(x,y,z) = \left(1 + 0.3\,x - 0.3\,y\right)
    \cos\!\left(\frac{a_1 \pi x}{2}\right)
    \cos\!\left(\frac{a_2 \pi y}{2}\right)
    \cos\!\left(\frac{a_3 \pi z}{2}\right),
    \label{eq:sg_uexact}
\end{equation}
where $a_1$, $a_2$, and $a_3$ are frequency parameters controlling the rate of oscillation of the solution along the $x$-, $y$-, and $z$-directions, respectively, and the affine prefactor $(1 + 0.3x - 0.3y)$ breaks the symmetry of the solution between the $x$- and $y$-directions.

Substituting $u_{\text{exact}}$ from Equation~\eqref{eq:sg_uexact} into the left-hand side of the governing equation, Equation~\eqref{eq:sine_gordon}, the corresponding source term is obtained directly as
\begin{equation}
    f(x,y,z) = -\Delta u_{\text{exact}}(x,y,z) + \sin\!\left(u_{\text{exact}}(x,y,z)\right),
    \label{eq:sg_source}
\end{equation}
where $\Delta u_{\text{exact}} = \partial_{xx} u_{\text{exact}} + \partial_{yy} u_{\text{exact}} + \partial_{zz} u_{\text{exact}}$. Owing to the transcendental nonlinearity of Equation~\eqref{eq:sg_source}, $f$ is evaluated symbolically rather than expressed in closed form. By construction, $u_{\text{exact}}$ in Equation~\eqref{eq:sg_uexact} satisfies Equation~\eqref{eq:sine_gordon} exactly with the source term given by Equation~\eqref{eq:sg_source}, providing a closed-form benchmark with a known ground-truth solution for evaluating the trained network. The frequency parameters along each direction is assigned with $a_1 = 15$, $a_2 = 15$, and $a_3 = 5$.

\subsubsection{Gradient Surgery to tackle Gradient Conflicts}

\begin{table}[H]
\centering
\begin{tabular}{|c|c|c|c|c||}
\hline
\multirow{2}{*}{\textbf{Gradient Surgery}} & \multicolumn{3}{c|}{\textbf{Relative $L_2$ Error}} & \multirow{2}{*}{\textbf{Time (ms/iter)}}\\ \cline{2-4}
& $\Omega_1$ & $\Omega_2$ & $\Omega_1 \cap \Omega_2$& \\
\hline\hline
None & 0.2154 $\pm$ 0.1765 & 0.2543 $\pm$ 0.2947 & 0.0705 $\pm$ 0.0158 & 5.89 \\
\hline
ConFIG & 0.5092 $\pm$ 0.4421 & 0.3821 $\pm$ 0.2579 & 0.1045 $\pm$ 0.0733 & 3.84 \\
\hline
PCGrad &  0.1646 $\pm$ 0.1064 & 0.2199 $\pm$ 0.2659 & 0.0353 $\pm$ 0.0244 & 6.62 \\
\hline
Norm-PCGrad & \textbf{0.1585 $\pm$ 0.0576} & \textbf{0.1658 $\pm$ 0.1830} & \textbf{0.0296 $\pm$ 0.0111} & 6.64 \\
\hline\hline
\end{tabular}
\caption{\textbf{3D sine-Gordon -- Gradient surgery.} Relative $L_2$ error (mean $\pm$ std. devi) and computational time for gradient surgery algorithms applied to 3D domain decomposition with an overlapping interfacial domain. $\Omega_1 \cap \Omega_2$ denotes the agreement between the two subdomains at the overlapping interface. Bold text represents the best results.}
\label{tab:3D_DD3_rl2_Surgery}
\end{table}

Here, we systematically solve the sine-Gordon equation shown as Equation \eqref{eq:sine_gordon} using PINNs with the support of gradient surgery and compare the performance against the case without any gradient surgery. While the previous discussions on utilization of the SSBroyden optimizer showed performance gains in accuracy, it poses a critical computational bottleneck due to the strict requirement for double-precision training. We have also shown that gradient surgery using a suitable first-order optimizer can attain performance comparable to the quasi-Newton optimizer. Based on these observations, we exclusively utilize gradient surgery with the state-of-the-art algorithms along with the proposed Norm-PCGrad.

\begin{figure}[H]
\centering
{
\centering
\includegraphics[width=\linewidth, trim={20mm 35mm 20mm 30mm}, clip]{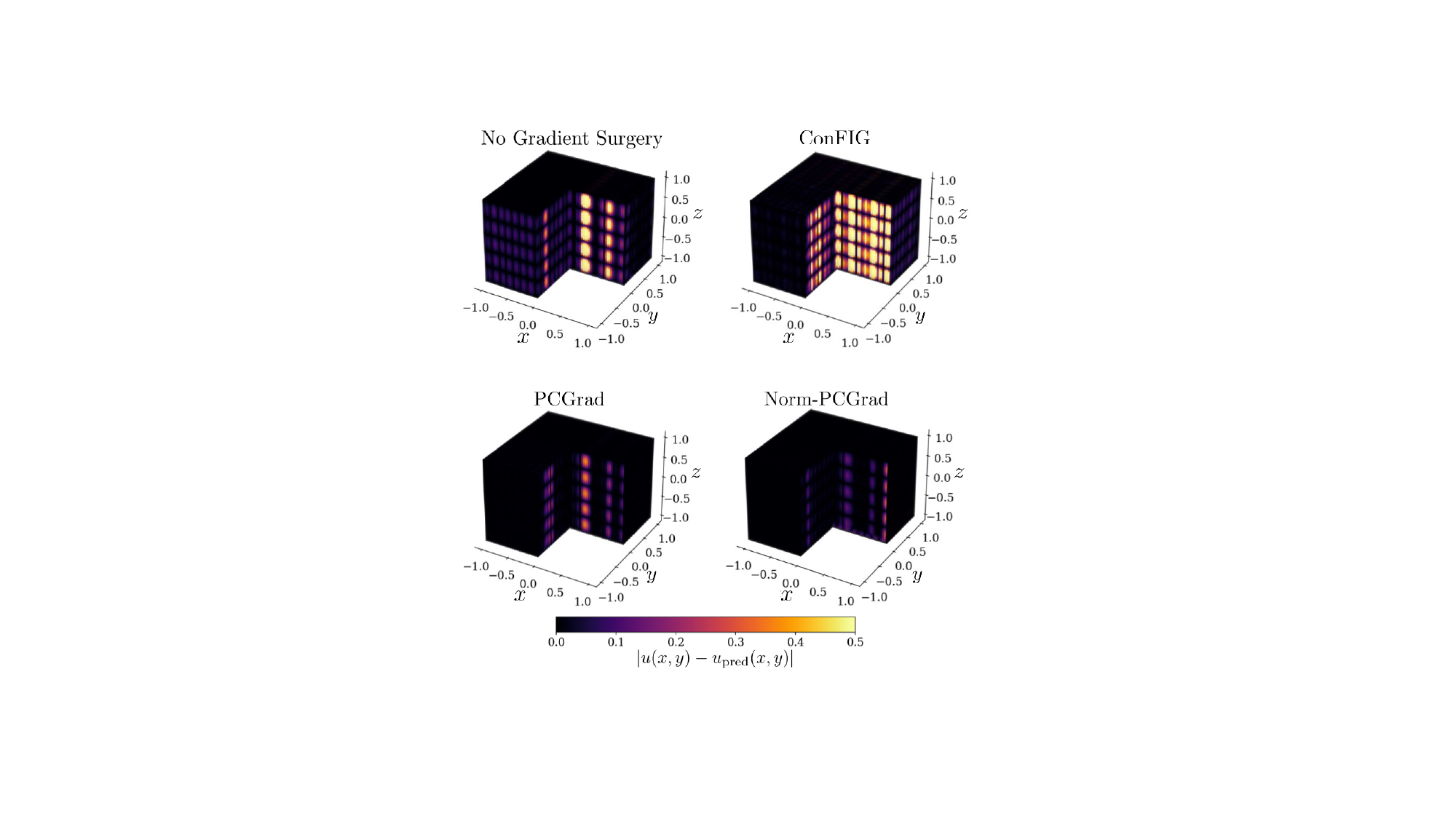}
}
\caption{\textbf{3D sine-Gordon -- Absolute error.} Point-wise absolute error of the solution to the sine-Gordon equation through 3D domain decomposition without gradient surgery and with gradient surgery using ConFIG, PCGrad, and Norm-PCGrad.}
\label{fig:3D_NonLinear_Comparison}
\end{figure}

Table \ref{tab:3D_DD3_rl2_Surgery} compares the performance in terms of the relative $L_2$ error and computational time for different algorithms of gradient surgery. Similar to previous sections, we observe Norm-PCGrad to surpass other algorithms in attaining a lower relative $L_2$ error at a comparable computational cost. It is also worthwhile to note that the best-performing algorithm was determined based on the relative $L_2$ error for the two subdomains.

Additionally, we examine the agreement between the two subdomains at the overlapping interfacial domain ($\Omega_1 \cap \Omega_2$) as an additional criterion for comparing performance across cases. We note that both PCGrad and Norm-PCGrad attain an order of magnitude lower relative $L_2$ error at the interfacial domain compared to ConFIG, and up to $50\%$ reduction in relative $L_2$ error compared to the case without gradient surgery. This difference in accuracy is further confirmed by the comparison of the point-wise absolute error shown in Figure \ref{fig:3D_NonLinear_Comparison}.

While ConFIG attains the best computational efficiency, as observed consistently across sections, it struggles with maintaining accuracy for solutions to PDEs in the context of domain decomposition. It is also worth noting that ConFIG and Norm-PCGrad attain the same two objectives, \textit{i.e.}, removing conflicts between the gradients and normalizing the resulting gradients along their new directions before aggregation, while differing in their process. We note that a given algorithm for gradient surgery can be distinguished based on its ability to respect the implicit coupling at the interface across subdomains, and ConFIG does not appear to respect this coupling. This may be a direction to address in future work.

\begin{figure}
\centering
{
\centering
\includegraphics[width=\linewidth, trim={0mm 50mm 0mm 50mm}, clip]{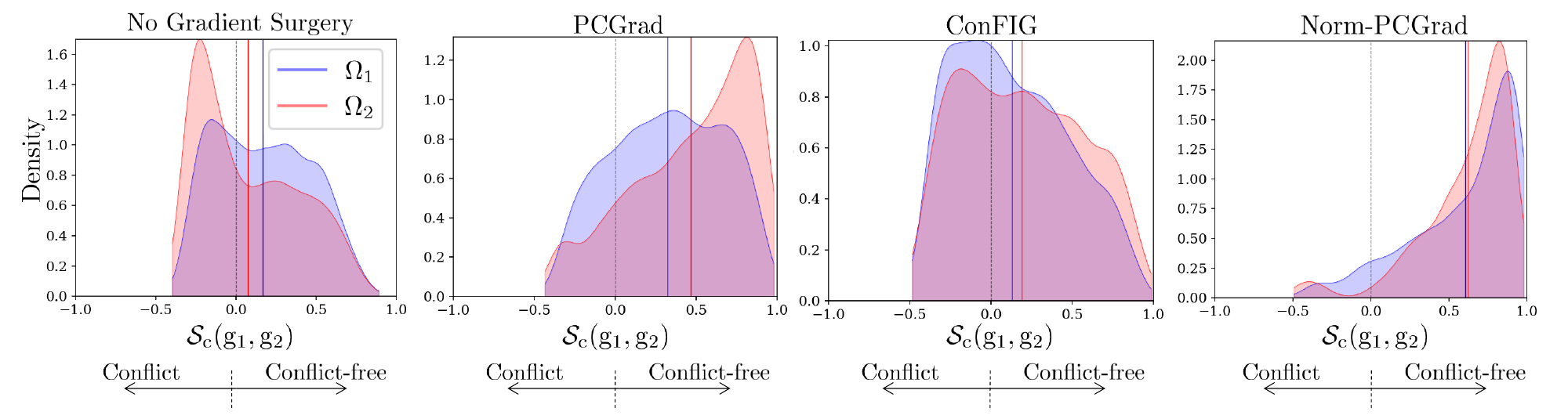}}
\caption{\textbf{3D sine-Gordon -- Cosine Similarity.} Density distribution of cosine similarity scores for the two subdomains, $\mathcal{S}_c^{\Omega_1}$ and $\mathcal{S}_c^{\Omega_2}$. Vertical solid lines represent the mean of each distribution ($\mathcal{S}_{c,~\text{mean}}^{\Omega_1}$ and $\mathcal{S}_{c,~\text{mean}}^{\Omega_2}$), while the dashed line marks the zero-line where gradients are orthogonal. If $\mathcal{S}_{c,~\text{mean}}^{\Omega_1}$ and/or $\mathcal{S}_{c,~\text{mean}}^{\Omega_2} < 0$, the gradients were in conflict for the majority of training in $\Omega_1$ and/or $\Omega_2$. On the other hand, if $\mathcal{S}_{c,~\text{mean}}^{\Omega_1}$ and/or $\mathcal{S}_{c,~\text{mean}}^{\Omega_2} > 0$, the gradients were conflict-free for the majority of training in $\Omega_1$ and/or $\Omega_2$. The extent of gradient conflicts (or their suppression) is measured by the distance of $\mathcal{S}_{c,~\text{mean}}^{\Omega_1}$ and $\mathcal{S}_{c,~\text{mean}}^{\Omega_2}$ from the zero-line. Here, the cosine similarity scores between gradient pairs are computed at fixed intervals during training and their distribution is represented as the density distribution.}
\label{fig:3D_NonLinear_cosine_similarity}
\end{figure}

While solving the 3D sine-Gordon equation using SPINN-SPINN domain decomposition, we compare the gradient conflicts across different algorithms of gradient surgery using the cosine similarity score, as shown in Figure \ref{fig:3D_NonLinear_cosine_similarity}. Interestingly, in the current scenario, all cases exhibit conflict-free gradients for the majority of training. The cases without gradient surgery and with ConFIG have $\mathcal{S}_{c,~\text{mean}}^{\Omega_1}$ and $\mathcal{S}_{c,~\text{mean}}^{\Omega_2} > 0$, indicating that gradients remain conflict-free but only moderately aligned. On the other hand, both PCGrad and Norm-PCGrad attain $\mathcal{S}_{c,~\text{mean}}^{\Omega_1}$ and $\mathcal{S}_{c,~\text{mean}}^{\Omega_2} \gg 0$, indicating strong gradient alignment for the majority of training. However, it is worthwhile to note that Norm-PCGrad remains the best case, since its distribution for both subdomains exhibits a strong left skew with the majority of the density concentrated in the right-half plane ($\mathcal{S}_c > 0$). These observations further support the trend of relative $L_2$ error for SPINN-SPINN domain decomposition shown in Table \ref{tab:3D_DD3_rl2_Surgery}.

\subsubsection{Kolmogorov-Arnold Networks}

\begin{figure}[H]
\centering
{
\centering
\includegraphics[width=0.95\linewidth, trim={25mm 34mm 25mm 25mm}, clip]{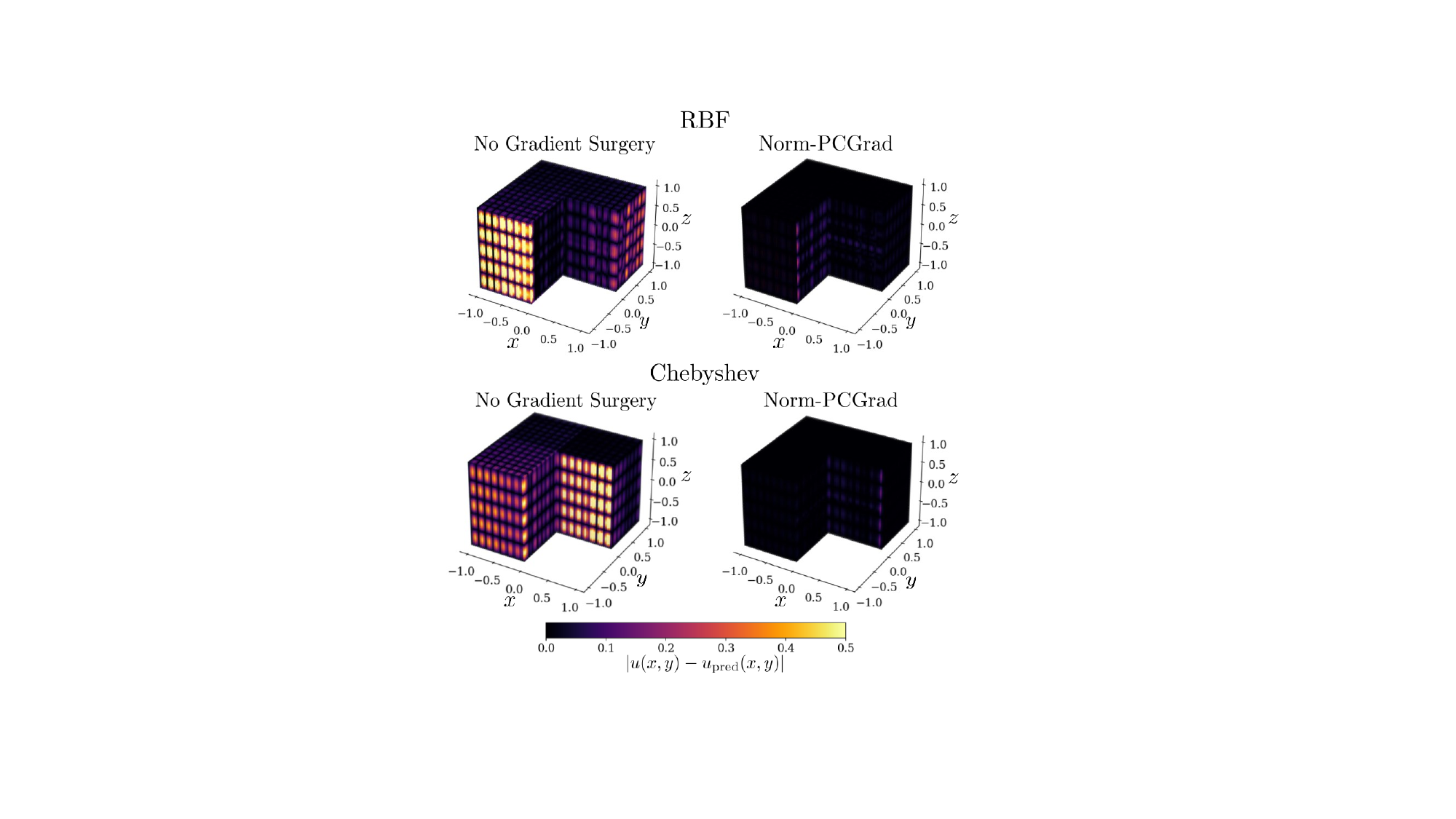}
}
\caption{\textbf{3D sine-Gordon -- Absolute error for PI-FEKAN comparison.} Point-wise absolute error of the solution to the sine-Gordon equation through 3D domain decomposition (Left) without gradient surgery and (Right) with Norm-PCGrad, for (Top) RBF and (Bottom) Chebyshev basis functions.}
\label{fig:3D_DD3_fekan}
\end{figure}

Having established the advantages of the Adam optimizer integrated with the proposed Norm-PCGrad algorithm for mitigation of gradient conflicts in domain decomposition, we leverage this combination to solve the sine-Gordon equation in the 3D domain shown in Figure \ref{fig:3D_NonLinear} using Kolmogorov-Arnold Networks. Unlike MLPs, the KAN architecture has both internal and external parameterizations that can vary substantially across different basis functions. Although the number of knots in the architecture differs between the Chebyshev basis and RBF, the total number of parameters is kept comparable for a fair comparison of performance, as noted in Table \ref{tab:3D_DD3_fekan}. Irrespective of the chosen basis function, we observe an order of magnitude reduction in the relative $L_2$ error at minimal increase in computational cost, as evident from both Table \ref{tab:3D_DD3_fekan} and Figure \ref{fig:3D_DD3_fekan}. For a comparable number of parameters, the Chebyshev basis attains better accuracy than RBF across the subdomains.

{\setlength{\tabcolsep}{3.5pt}
\begin{table}[H]
\centering
\small
\begin{tabular}{|c|c|c||c|c|c||c||}
\hline
\multirow{2}{*}{\textbf{Basis}} & \multirow{2}{*}{\textbf{Architecture}} & \multirow{2}{*}{\textbf{GS}} & \multicolumn{3}{c||}{\textbf{Relative $L_2$ Error}} & \multirow{2}{*}{\textbf{Time (ms/iter)}}\\
\cline{4-6}
& & & $\Omega_1$ & $\Omega_2$ & $\Omega_1 \cap \Omega_2$ & \\
\hline\hline
\multirow{2}{*}{Chebyshev} & \multirow{2}{*}{$[5, 13, 13, 10] \times (2)$} & \xmark & 0.2020 $\pm$ 0.0712 & 0.1778 $\pm$ 0.0393 & 0.2565 $\pm$ 0.0393 & 72.45 \\ \cline{3-7}
& & \cmark & \textbf{0.0829 $\pm$ 0.0677} & \textbf{0.1322 $\pm$ 0.1862} & \textbf{0.0466 $\pm$ 0.0128} & 73.13 \\ 
\hline\hline
\multirow{2}{*}{RBF} & \multirow{2}{*}{$[5, 7, 7, 10] \times (2)$} & \xmark & 0.2719 $\pm$ 0.0961 & 0.2628 $\pm$ 0.0245 & 0.2395 $\pm$ 0.0757 & 61.56 \\ \cline{3-7}
& & \cmark & \textbf{0.1301 $\pm$ 0.0671} & \textbf{0.1422 $\pm$ 0.0870} & \textbf{0.0865 $\pm$ 0.0240} & 62.34 \\ 
\hline
\end{tabular}
\caption{\textbf{3D sine-Gordon -- PI-FEKAN with gradient surgery.} Relative $L_2$ error (mean $\pm$ std. devi) and computational time for PI-FEKAN using Chebyshev ($k=2$) and RBF ($d=5$) basis functions for 3D domain decomposition.  GS denotes gradient surgery using Norm-PCGrad. The architecture uses 3-body networks with 5 Fourier feature enrichment terms. Parameter counts are comparable: 1600 (Chebyshev) and 1636 (RBF). Bold text represents the best results.}
\label{tab:3D_DD3_fekan}
\end{table}}

The performance gain in accuracy is substantial for both basis functions, with the relative $L_2$ error reducing by up to $2\times$ when Norm-PCGrad is applied. This further confirms the existence of gradient conflicts in the context of 3D domain decomposition for nonlinear PDEs, even when using Kolmogorov-Arnold Networks, and demonstrates that leveraging an appropriate method for gradient surgery can lead to substantial performance gains. It is also noteworthy that Norm-PCGrad leads to an order of magnitude reduction in the relative $L_2$ error at the common interfacial domain $\Omega_1 \cap \Omega_2$, while the reduction independently for each subdomain can be relatively modest. This indicates that subdomains can achieve good agreement at the interface while still exhibiting larger errors globally compared to the physically consistent ground truth.

Across the current and previous sections, we have demonstrated the scalability of KANs for solving both linear and nonlinear PDEs through 3D domain decomposition, and confirmed the benefits of gradient surgery across different basis functions for KAN-based architectures.

\section{Conclusions}
This work emphasizes the importance of tackling \textit{gradient conflicts} while solving PDEs using PINNs and KANs through domain decomposition, predominantly using gradient surgery based on projection of conflicting gradients in the loss landscape. We compare existing algorithms for gradient surgery and propose Norm-PCGrad to tackle such conflicts in the context of domain decomposition when existing algorithms tend towards degraded performance. Besides gradient surgery, we also recognize curvature-aware optimization using the Broyden family of optimizers, specifically the SSBroyden optimizer, and use it to address gradient conflicts either independently or in combination with the first-order Adam optimizer. This work also discusses the extension of 3D domain decomposition to Kolmogorov-Arnold Networks. We note that the existing literature is fairly limited to simple 2D domain decomposition using KANs with hard-constrained interfacial modeling using a predefined \textit{ansatz}, whose selection becomes non-trivial for 3D domains. To the best of our knowledge, this is the first work to demonstrate the scalability of KANs for 3D domain decomposition while tackling conflicting gradients using the proposed Norm-PCGrad algorithm. The valuable insights from this work can be summarized as follows:

\begin{itemize}
    \item The current work has demonstrated that training a neural network for solving PDEs can still be performed with fixed or non-tuned weights for the loss terms to achieve superior accuracy and convergence if the gradient conflicts can be eliminated or substantially mitigated from the training process without utilization of any dynamic weighting strategies.
    \item Our results clearly demonstrate the benefits of mitigating conflicting gradients for domain decomposition even while utilizing fixed, non-tuned weights for the loss terms. Specifically, the proposed Norm-PCGrad algorithm attains the best performance even for cases where existing state-of-the-art algorithms such as PCGrad and ConFIG exhibited degraded performance.
    \item The explicit normalization step in Norm-PCGrad normalizes the gradients for the individual loss terms, thereby preventing abrupt changes to the gradient directions while still mitigating conflicts between them. Consequently, Norm-PCGrad respects the implicit coupling between interface gradients across any two subdomains while performing gradient projections. In contrast, PCGrad without normalization can undergo abrupt changes to gradient directions, potentially breaking this coupling and resulting in discontinuous solutions at the interface, as observed in our empirical studies. The performance of ConFIG was the worst among all algorithms tested for both 2D and 3D domain decompositions. This is further supported by observations of recent works \cite{kim2026per} on the degraded performance of ConFIG for solving PDEs in single domains.
    \item We also establish theoretical results showing that Norm-PCGrad provides a magnitude-balanced and geometrically principled framework for gradient aggregation for the case in which the loss function includes two terms.
    \item This work also discusses the potential of curvature-aware optimization using the SSBroyden optimizer for mitigation of gradient conflicts in the context of 2D and 3D domain decomposition. While the empirical results demonstrate that curvature-aware optimization indeed helps in suppressing gradient conflicts, it comes at a higher computational cost than Adam, as observed through empirical studies. Moreover, the performance of SSBroyden was competitive but inferior to the Adam optimizer equipped with gradient surgery using the proposed Norm-PCGrad.
    \item We also extended gradient surgery using Norm-PCGrad for domain decomposition of 3D domains to Kolmogorov-Arnold Networks. We specifically utilized the Feature-Enriched KAN \cite{menon2026fekan} architecture to test performance gains and scalability using efficient formulations of basis functions based on Chebyshev polynomials \cite{ss2024chebyshev} and RBF \cite{abueidda2025deepokan}. Our observations suggested Chebyshev polynomials with Norm-PCGrad as the best combination for 3D domain decomposition using KANs.
    \item While modeling the interface as an overlapping interfacial domain is beneficial, it can reduce the computational efficiency of a \textit{vanilla} PINN. To address this bottleneck, we propose SPINN-PINN domain decomposition, leveraging the separable SPINN architecture \cite{cho2023separable} for a faster XPINN-based domain decomposition. This is demonstrated empirically in Appendix \ref{appx:2D_DD}.
    \item Building on this, we propose the mSPINN algorithm in the context of domain decomposition to solve the Euclidean subdomains using a multi-grid approach by sampling staggered collocation grids, although at a higher computational cost from the current serial implementation. Using mSPINN, we observe orders of magnitude improvement in accuracy compared to domain decomposition using the \textit{vanilla} PINN or the SPINN architecture for the Helmholtz equation. We refrained from further comparison of its performance until mSPINN can be parallelized to attain computational costs comparable to SPINN or PINN architectures, which may be pursued as future work.
\end{itemize}

While the hybrid domain decomposition proposed in this work showcased enhanced computational efficiency compared to standard PINN-PINN decomposition, it is worth noting its limitations so that they can be addressed in future work. The hybrid approach discussed here entails an SPINN-PINN decomposition, where a domain is decomposed into Euclidean and non-Euclidean subdomains so that the Euclidean subdomain can be solved using a separable architecture at linear computational cost. We have also discussed an SPINN-SPINN decomposition for several 2D and 3D domains, demonstrating its performance gains when combined with gradient surgery. However, in applications dealing with purely non-Euclidean domains, we propose the utilization of an SPINN-PINN decomposition, where a non-Euclidean domain can be decomposed into a combination of Euclidean and non-Euclidean subdomains. For domains with extremely complicated topologies (e.g., a helical spring), it may be difficult to achieve such decomposition without drastic geometric simplifications, and in those instances the most suitable approach remains a standard PINN-PINN (XPINN) decomposition. It is also worth noting that all existing works, including the current work, deal with a manageable number of loss terms within the composite loss function. However, handling the pair-wise gradient projections can become a computational bottleneck for cases with a large number of loss terms, and this can pose a major limitation for the algorithms of gradient surgery.

Based on our results and regardless of the underlying architecture, we emphasize the utilization of gradient surgery for domain decomposition in the context of PINNs (or any PDE-constrained training in general), to ensure conflict-free gradients that can lead to superior accuracy, both globally and across the interface, along with improved convergence rates, with negligible overhead in computational cost.

\section*{Acknowledgements}
The second author received through Sandia National Laboratories Laboratory Directed Research and Development (LDRD) program and through the U.S. Department of Energy, Office of Science, Office of Advanced Scientific Computing Research, Mathematical Multifaceted Integrated Capability Centers (MMICCs) program, under Field Work Proposal 22025291 and the Multifaceted Mathematics for Predictive Digital Twins (M2dt) project. 
Sandia National Laboratories is a multi-mission laboratory managed and operated by National Technology and Engineering Solutions of Sandia, LLC., a wholly owned subsidiary of Honeywell International, Inc., for the U.S. Department of Energy’s National Nuclear Security Administration under contract DE-NA0003525.

\appendix

\section{Theoretical Properties of Norm-PCGrad}

We establish three key theoretical properties of Norm-PCGrad for the two-loss case. First, we show that the projected PCGrad direction is conflict-free. Second, we quantify the gradient-magnitude bias induced by PCGrad and show how normalization removes the explicit projected-magnitude weighting. Finally, we establish that Norm-PCGrad is the max--min optimal direction in the normalized projected-gradient space. We further show that this direction is identical to the two-gradient ConFIG direction.

\noindent
Let $\mathbf{g}_1,\mathbf{g}_2\in\mathbb{R}^d$ denote two nonzero
gradients satisfying
\begin{equation}
\mathbf{g}_1^\top\mathbf{g}_2<0.
\label{eq:theory_conflict}
\end{equation}
Define the orthogonalized gradient components
\begin{align}
\mathbf{p}_1
&=
\mathcal{O}(\mathbf{g}_1,\mathbf{g}_2)
=
\mathbf{g}_2
-
\frac{\mathbf{g}_1^\top\mathbf{g}_2}
{\|\mathbf{g}_1\|^2}\mathbf{g}_1,
\label{eq:theory_p1}\\
\mathbf{p}_2
&=
\mathcal{O}(\mathbf{g}_2,\mathbf{g}_1)
=
\mathbf{g}_1
-
\frac{\mathbf{g}_1^\top\mathbf{g}_2}
{\|\mathbf{g}_2\|^2}\mathbf{g}_2.
\label{eq:theory_p2}
\end{align}
We assume that the gradients are non-collinear so that
$\mathbf{p}_1\neq\mathbf{0}$ and $\mathbf{p}_2\neq\mathbf{0}$.

\paragraph{Theorem 1 (Conflict-free property of two-gradient PCGrad).}
\label{thm1:pcgrad_conflict_free}
For two nonzero, non-collinear conflicting gradients, the PCGrad
update
\begin{equation*}
\mathbf{g}_{\mathrm{PCGrad}}
=
\mathbf{p}_1+\mathbf{p}_2
\label{eq:theory_pcgrad}
\end{equation*}
satisfies

\begin{align*}
\mathbf{g}_1^\top\mathbf{g}_{\mathrm{PCGrad}}
&=
\|\mathbf{p}_2\|^2>0,\\
\mathbf{g}_2^\top\mathbf{g}_{\mathrm{PCGrad}}
&=
\|\mathbf{p}_1\|^2>0.
\end{align*}
Therefore, $\mathbf{g}_{\mathrm{PCGrad}}$ is a simultaneous
first-order descent direction for both loss functions.

\paragraph{Proof.}
By construction,
\begin{equation*}
\mathbf{g}_1^\top\mathbf{p}_1=0,
\qquad
\mathbf{g}_2^\top\mathbf{p}_2=0.
\end{equation*}
Moreover,
\begin{equation*}
\mathbf{g}_1^\top\mathbf{p}_2
=
\|\mathbf{p}_2\|^2,
\qquad
\mathbf{g}_2^\top\mathbf{p}_1
=
\|\mathbf{p}_1\|^2.
\end{equation*}
Consequently,
\begin{align*}
\mathbf{g}_1^\top\mathbf{g}_{\mathrm{PCGrad}}
&=
\mathbf{g}_1^\top
(\mathbf{p}_1+\mathbf{p}_2)
\nonumber\\
&=
0+\|\mathbf{p}_2\|^2
=
\|\mathbf{p}_2\|^2>0,
\end{align*}
and similarly,
\begin{align*}
\mathbf{g}_2^\top\mathbf{g}_{\mathrm{PCGrad}}
&=
\mathbf{g}_2^\top
(\mathbf{p}_1+\mathbf{p}_2)
\nonumber\\
&=
\|\mathbf{p}_1\|^2+0
=
\|\mathbf{p}_1\|^2>0.
\end{align*}
Hence, for an update
$\theta^+=\theta-\eta\mathbf{g}_{\mathrm{PCGrad}}$ with sufficiently
small $\eta>0$, both losses decrease to first order.
\hfill$\square$

\paragraph{Theorem 2 (Magnitude bias of PCGrad and normalization).}
\label{thm2:magnitude_bias}
Let
\begin{equation*}
\mathcal{S}_c(\mathbf{g_1},\mathbf{g_2}) =
\frac{\mathbf{g}_1^\top\mathbf{g}_2}
{\|\mathbf{g}_1\|\,\|\mathbf{g}_2\|}
\end{equation*}
denote the cosine similarity between the two gradients. Then the
orthogonalized components satisfy
\begin{align}
\|\mathbf{p}_1\|
&=
\|\mathbf{g}_2\|\sqrt{1-c^2},
\label{eq:p1_magnitude}\\
\|\mathbf{p}_2\|
&=
\|\mathbf{g}_1\|\sqrt{1-c^2}.
\label{eq:p2_magnitude}
\end{align}
Consequently, the relative first-order contribution of the two losses
under PCGrad is
\begin{equation}
\boxed{
\frac{
\mathbf{g}_1^\top\mathbf{g}_{\mathrm{PCGrad}}
}{
\mathbf{g}_2^\top\mathbf{g}_{\mathrm{PCGrad}}
}
=
\frac{\|\mathbf{g}_1\|^2}
{\|\mathbf{g}_2\|^2}.
}
\label{eq:pcgrad_magnitude_bias}
\end{equation}

\noindent
Thus, PCGrad exhibits a quadratic dependence on the relative gradient
magnitudes.
\noindent
In contrast, define the normalized projected components
\begin{equation*}
\mathbf{u}_1
=
\frac{\mathbf{p}_1}{\|\mathbf{p}_1\|},
\qquad
\mathbf{u}_2
=
\frac{\mathbf{p}_2}{\|\mathbf{p}_2\|},
\label{eq:normalized_projected_components}
\end{equation*}
and the Norm-PCGrad update
\begin{equation*}
\mathbf{g}_{\mathrm{Norm}}
=
\mathbf{u}_1+\mathbf{u}_2.
\label{eq:normpcgrad_theory}
\end{equation*}
Then
\begin{align*}
\mathbf{g}_1^\top\mathbf{g}_{\mathrm{Norm}}
&=
\|\mathbf{p}_2\|,
\\
\mathbf{g}_2^\top\mathbf{g}_{\mathrm{Norm}}
&=
\|\mathbf{p}_1\|,
\end{align*}
and therefore

\begin{equation}
\boxed{
\frac{
\mathbf{g}_1^\top\mathbf{g}_{\mathrm{Norm}}
}{
\mathbf{g}_2^\top\mathbf{g}_{\mathrm{Norm}}
}
=
\frac{\|\mathbf{g}_1\|}
{\|\mathbf{g}_2\|}.
}
\label{eq:normpcgrad_magnitude_bias}
\end{equation}

\noindent
Hence, normalization changes the quadratic magnitude dependence of PCGrad to a linear dependence and removes the explicit weighting of the aggregate direction by the magnitudes of the projected gradients.

\paragraph{Proof.}
From Equation~\eqref{eq:theory_p1},
\begin{align*}
\|\mathbf{p}_1\|^2
&=
\left\|
\mathbf{g}_2
-
\frac{\mathbf{g}_1^\top\mathbf{g}_2}
{\|\mathbf{g}_1\|^2}\mathbf{g}_1
\right\|^2
\nonumber\\
&=
\|\mathbf{g}_2\|^2
-
\frac{
(\mathbf{g}_1^\top\mathbf{g}_2)^2
}{
\|\mathbf{g}_1\|^2
}
\nonumber\\
&=
\|\mathbf{g}_2\|^2(1-c^2),
\end{align*}
which gives Equation~\eqref{eq:p1_magnitude}. Similarly,
\begin{align*}
\|\mathbf{p}_2\|^2
&=
\left\|
\mathbf{g}_1
-
\frac{\mathbf{g}_1^\top\mathbf{g}_2}
{\|\mathbf{g}_2\|^2}\mathbf{g}_2
\right\|^2
\nonumber\\
&=
\|\mathbf{g}_1\|^2
-
\frac{
(\mathbf{g}_1^\top\mathbf{g}_2)^2
}{
\|\mathbf{g}_2\|^2
}
\nonumber\\
&=
\|\mathbf{g}_1\|^2(1-c^2),
\end{align*}
which gives Equation~\eqref{eq:p2_magnitude}.

\noindent
By Theorem~1
\begin{equation*}
\mathbf{g}_1^\top\mathbf{g}_{\mathrm{PCGrad}}
=
\|\mathbf{p}_2\|^2,
\qquad
\mathbf{g}_2^\top\mathbf{g}_{\mathrm{PCGrad}}
=
\|\mathbf{p}_1\|^2.
\end{equation*}
Therefore,
\begin{align*}
\frac{
\mathbf{g}_1^\top\mathbf{g}_{\mathrm{PCGrad}}
}{
\mathbf{g}_2^\top\mathbf{g}_{\mathrm{PCGrad}}
}
&=
\frac{\|\mathbf{p}_2\|^2}
{\|\mathbf{p}_1\|^2}
\nonumber\\
&=
\frac{\|\mathbf{g}_1\|^2(1-c^2)}
{\|\mathbf{g}_2\|^2(1-c^2)}
\nonumber\\
&=
\frac{\|\mathbf{g}_1\|^2}
{\|\mathbf{g}_2\|^2}.
\end{align*}

\noindent
For Norm-PCGrad, using
$\mathbf{g}_1^\top\mathbf{p}_1=0$ and
$\mathbf{g}_2^\top\mathbf{p}_2=0$,
\begin{align*}
\mathbf{g}_1^\top\mathbf{g}_{\mathrm{Norm}}
&=
\frac{
\mathbf{g}_1^\top\mathbf{p}_1
}{
\|\mathbf{p}_1\|}
+
\frac{
\mathbf{g}_1^\top\mathbf{p}_2
}{
\|\mathbf{p}_2\|}
\nonumber\\
&=
\frac{\|\mathbf{p}_2\|^2}{\|\mathbf{p}_2\|}
=
\|\mathbf{p}_2\|,
\\
\mathbf{g}_2^\top\mathbf{g}_{\mathrm{Norm}}
&=
\frac{
\mathbf{g}_2^\top\mathbf{p}_1
}{
\|\mathbf{p}_1\|}
+
\frac{
\mathbf{g}_2^\top\mathbf{p}_2
}{
\|\mathbf{p}_2\|}
\nonumber\\
&=
\frac{\|\mathbf{p}_1\|^2}{\|\mathbf{p}_1\|}
=
\|\mathbf{p}_1\|.
\end{align*}
Substitution of Equations~\eqref{eq:p1_magnitude} and
\eqref{eq:p2_magnitude} gives
\begin{align*}
\frac{
\mathbf{g}_1^\top\mathbf{g}_{\mathrm{Norm}}
}{
\mathbf{g}_2^\top\mathbf{g}_{\mathrm{Norm}}
}
&=
\frac{\|\mathbf{p}_2\|}
{\|\mathbf{p}_1\|}
=
\frac{\|\mathbf{g}_1\|}
{\|\mathbf{g}_2\|}.
\end{align*}
\hfill$\square$

\paragraph{Theorem 3 (Max--min optimality of Norm-PCGrad).}
\label{thm3:maxmin_normpcgrad}
Under the assumptions above, let
\begin{equation*}
\mathbf{u}_1
=
\frac{\mathbf{p}_1}{\|\mathbf{p}_1\|},
\qquad
\mathbf{u}_2
=
\frac{\mathbf{p}_2}{\|\mathbf{p}_2\|}.
\end{equation*}
The unit-norm Norm-PCGrad direction
\begin{equation*}
\mathbf{d}_{\mathrm{Norm}}
=
\frac{
\mathbf{u}_1+\mathbf{u}_2
}{
\|\mathbf{u}_1+\mathbf{u}_2\|
}
\label{eq:norm_unit_direction}
\end{equation*}
solves the max--min optimization problem
\begin{equation}
\boxed{
\mathbf{d}_{\mathrm{Norm}}
=
\underset{\|\mathbf d\|=1}{\operatorname{arg\,max}}
\;
\min
\left\{
\mathbf d^\top\mathbf u_1,
\mathbf d^\top\mathbf u_2
\right\}.
}
\label{eq:maxmin_problem}
\end{equation}
Therefore, Norm-PCGrad selects the unit-norm direction that maximizes the worst-case alignment with the two normalized orthogonalized gradient components.

\paragraph{Proof.}
First, the inner product between the two projected components is
\begin{align*}
\mathbf{p}_1^\top\mathbf{p}_2
&=
\left(
\mathbf{g}_2
-
\frac{\mathbf{g}_1^\top\mathbf{g}_2}
{\|\mathbf{g}_1\|^2}\mathbf{g}_1
\right)^\top
\left(
\mathbf{g}_1
-
\frac{\mathbf{g}_1^\top\mathbf{g}_2}
{\|\mathbf{g}_2\|^2}\mathbf{g}_2
\right)
\nonumber\\
&=
-\mathbf{g}_1^\top\mathbf{g}_2
\left(
1-
\frac{
(\mathbf{g}_1^\top\mathbf{g}_2)^2
}{
\|\mathbf{g}_1\|^2\|\mathbf{g}_2\|^2
}
\right)
\nonumber\\
&=
-\mathbf{g}_1^\top\mathbf{g}_2(1-c^2)>0.
\end{align*}
Therefore,
\begin{equation*}
0<\mathbf{u}_1^\top\mathbf{u}_2<1,
\end{equation*}
where the strict upper bound follows from the non-collinearity assumption. In particular,
$\mathbf{u}_1+\mathbf{u}_2\neq\mathbf{0}$.

\noindent
For any unit vector $\mathbf d$, define
\begin{equation*}
q(\mathbf d)
=
\min
\left\{
\mathbf d^\top\mathbf u_1,
\mathbf d^\top\mathbf u_2
\right\}.
\end{equation*}
Using
\begin{equation*}
\min\{a,b\}
=
\frac{a+b-|a-b|}{2},
\end{equation*}
we obtain
\begin{equation*}
q(\mathbf d)
=
\frac{
\mathbf d^\top(\mathbf u_1+\mathbf u_2)
-
\left|
\mathbf d^\top(\mathbf u_1-\mathbf u_2)
\right|
}{2}.
\end{equation*}
Since the second term is nonnegative,
\begin{equation*}
q(\mathbf d)
\leq
\frac{
\mathbf d^\top(\mathbf u_1+\mathbf u_2)
}{2}
\leq
\frac{
\|\mathbf u_1+\mathbf u_2\|
}{2},
\end{equation*}
where the second inequality follows from the Cauchy--Schwarz
inequality.

\noindent
Now choose
\begin{equation*}
\mathbf d
=
\mathbf d_{\mathrm{Norm}}
=
\frac{
\mathbf u_1+\mathbf u_2
}{
\|\mathbf u_1+\mathbf u_2\|
}.
\end{equation*}
For this choice,
\begin{align*}
\mathbf d_{\mathrm{Norm}}^\top\mathbf u_1
&=
\frac{
1+\mathbf u_1^\top\mathbf u_2
}{
\|\mathbf u_1+\mathbf u_2\|
},
\\
\mathbf d_{\mathrm{Norm}}^\top\mathbf u_2
&=
\frac{
1+\mathbf u_1^\top\mathbf u_2
}{
\|\mathbf u_1+\mathbf u_2\|
}.
\end{align*}
Thus the two projections are equal. Moreover,
\begin{equation*}
\|\mathbf u_1+\mathbf u_2\|^2
=
2(1+\mathbf u_1^\top\mathbf u_2),
\end{equation*}
and hence
\begin{equation*}
q(\mathbf d_{\mathrm{Norm}})
=
\frac{
1+\mathbf u_1^\top\mathbf u_2
}{
\|\mathbf u_1+\mathbf u_2\|
}
=
\frac{
\|\mathbf u_1+\mathbf u_2\|
}{2}.
\end{equation*}
This attains the upper bound above. Therefore,
$\mathbf d_{\mathrm{Norm}}$ is a maximizer of the max--min problem.
\hfill$\square$

\paragraph{Corollary (Equivalence of Norm-PCGrad and two-gradient
ConFIG directions).}
\label{cor1:normpcgrad_config}
For two nonzero, non-collinear conflicting gradients, the
Norm-PCGrad direction is identical to the two-gradient ConFIG
direction:
\begin{equation}
\boxed{
\mathbf d_{\mathrm{Norm}}
=
\mathcal{U}
\left[
\mathcal{U}(\mathbf{p}_1)
+
\mathcal{U}(\mathbf{p}_2)
\right]
=
\mathbf g_v.
}
\label{eq:normpcgrad_config_equivalence}
\end{equation}
Hence, PCGrad, Norm-PCGrad, and ConFIG have the same conflict-free direction in the two-loss case only after accounting for the normalization of the projected components. PCGrad differs through magnitude weighting, whereas Norm-PCGrad and ConFIG generate the same unit direction.

\paragraph{Proof.}
By definition,
\begin{equation*}
\mathcal{U}(\mathbf{p}_1)=\mathbf u_1,
\qquad
\mathcal{U}(\mathbf{p}_2)=\mathbf u_2.
\end{equation*}
Therefore,
\begin{align*}
\mathbf g_v
&=
\mathcal{U}
\left[
\mathbf u_1+\mathbf u_2
\right]
\nonumber\\
&=
\frac{
\mathbf u_1+\mathbf u_2
}{
\|\mathbf u_1+\mathbf u_2\|
}
\nonumber\\
&=
\mathbf d_{\mathrm{Norm}}.
\end{align*}
\hfill$\square$

Taken together, our results establish Norm-PCGrad as a principled approach to balancing conflicting objectives. Unlike PCGrad, whose projected gradients retain a quadratic dependence on the original gradient norms, Norm-PCGrad reduces this dependence to linear, mitigating the dominance of large-gradient objectives. Moreover, Norm-PCGrad is the max–min optimal direction in normalized projected-gradient space, maximizing worst-case alignment across objectives. For two gradients, it is exactly equivalent to ConFIG, linking gradient normalization to an independent optimality principle. Thus, Norm-PCGrad provides a magnitude-balanced and geometrically principled approach to gradient aggregation.

\section{Multi-grid SPINN (mSPINN)}
\label{appx:mspinn}
The pseudo-code for the multi-grid SPINN algorithm discussed in Section  \ref{sec:2D_DD}. is described here.

\small \begin{algorithm}[H]
\caption{Multi-grid SPINN (mSPINN)}
\begin{minipage}{\linewidth}
\begin{algorithmic}[1]
\REQUIRE Surrogate $u_\theta$ with parameters $\theta$; grid pool $\mathcal{G} = \{\mathcal{G}_1, \dots, \mathcal{G}_n\}$; batch size $m \ll n$; learning rate $\epsilon$
\FOR{$i = 1$ to \textbf{EPOCHS}}
    \STATE Sample a batch $\mathcal{G}_{\rm batch} \subset \mathcal{G}$ of $m$ grids uniformly at random
    \FOR{$\mathcal{G}_j \in \mathcal{G}_{\rm batch}$}
        \STATE Compute loss $\mathcal{L}(\theta_i; \mathcal{G}_j)$ and gradient $\nabla_\theta \mathcal{L}(\theta_i; \mathcal{G}_j)$
    \ENDFOR
    \STATE // Average the loss and gradient over the $m$ sampled grids
    \STATE $\mathcal{L}_{\rm avg}(\theta_i; \mathcal{G}_{\rm batch}) \leftarrow \dfrac{1}{m} \sum_{j=1}^{m} \mathcal{L}(\theta_i; \mathcal{G}_j)$
    \STATE $\nabla_\theta \mathcal{L}_{\rm avg}(\theta_i; \mathcal{G}_{\rm batch}) \leftarrow \dfrac{1}{m} \sum_{j=1}^{m} \nabla_\theta \mathcal{L}(\theta_i; \mathcal{G}_j)$
    \STATE // Parameter update; ADAM or any other choice of optimizer can replace this plain gradient step in practice
    \STATE $\theta_{i+1} \leftarrow \theta_i - \epsilon \, \nabla_\theta \mathcal{L}_{\rm avg}(\theta_i; \mathcal{G}_{\rm batch})$
\ENDFOR
\end{algorithmic}
\end{minipage}
\end{algorithm}

\section{Sampling Summary}
\label{appx:sampling}
This appendix provides a detailed summary of the collocation points used for each benchmark problem discussed in the current work. Table \ref{tab:samples} reports the number of residual, boundary, and interface points used per subdomain for each problem. All residual and interface points are sampled on implicit structured grids, while boundary points are distributed across the respective faces of each subdomain.

\begin{table}[H]
\centering
\scriptsize
\begin{tabular}{|c|c|c|c|}
\hline
\textbf{Problem} & \textbf{Subdomain} & \textbf{Type} & \textbf{Samples} \\
\hline\hline
\multirow{5}{*}{2D Helmholtz} & \multirow{2}{*}{$\Omega_1$ (SPINN)} & Residual & 16384 \\ \cline{3-4}
 & & Boundary & 4 \\ \cline{2-4}
 & \multirow{2}{*}{$\Omega_2$ (PINN)} & Residual & 5312 \\ \cline{3-4}
 & & Boundary & 2000 \\ \cline{2-4}
 & $\Omega_1 \cap \Omega_2$ & Interface & 512 \\
\hline\hline
\multirow{5}{*}{2D Poisson} & \multirow{2}{*}{$\Omega_1$ (SPINN)} & Residual & 1024 \\ \cline{3-4}
 & & Boundary & 96 \\ \cline{2-4}
 & \multirow{2}{*}{$\Omega_2$ (SPINN)} & Residual & 1024 \\ \cline{3-4}
 & & Boundary & 160 \\ \cline{2-4}
 & $\Omega_1 \cap \Omega_2$ & Interface & 1024 \\
\hline\hline
\multirow{7}{*}{\shortstack{3D Poisson\\(3 subdomains)}} & \multirow{2}{*}{$\Omega_1$ (SPINN)} & Residual & 524288 \\ \cline{3-4}
 & & Boundary & 73728 \\ \cline{2-4}
 & \multirow{3}{*}{$\Omega_2$ (SPINN)} & Residual & 262144 \\ \cline{3-4}
 & & Boundary & 24576 \\ \cline{3-4}
 & & Interface & 262144 \\ \cline{2-4}
 & \multirow{3}{*}{$\Omega_3$ (SPINN)} & Residual & 262144 \\ \cline{3-4}
 & & Boundary & 24576 \\ \cline{3-4}
 & & Interface & 262144 \\
\hline\hline
\multirow{6}{*}{\shortstack{3D Sine-Gordon}} & \multirow{3}{*}{$\Omega_1$ (SPINN)} & Residual & 32768 \\ \cline{3-4}
 & & Boundary & 6144 \\ \cline{3-4}
 & & Interface & 32768 \\ \cline{2-4}
 & \multirow{3}{*}{$\Omega_2$ (SPINN)} & Residual & 32768 \\ \cline{3-4}
 & & Boundary & 6144 \\ \cline{3-4}
 & & Interface & 32768 \\
\hline\hline
\end{tabular}
\caption{\textbf{Sampling points per problem.} Number of samples used for residual, boundary, and interface terms. For 2D Helmholtz: $\Omega_2$ residual includes interior, circular boundary, and interface points folded in. For 2D Poisson: residual points follow an implicit $32\times32$ grid; interface points are shared by both subdomains. For 3D Poisson (3 subdomains): $\Omega_1$ acts as the hub subdomain while $\Omega_2$ and $\Omega_3$ are leaf subdomains. For 3D Sine-Gordon: residual and interface points follow an implicit $32^3$ grid. For a given PDE, the sampling is the same for both MLP and KAN variants. Note that the points sampled at the interface or interfacial domain serve as collocation points and are not measurements of the solution.}
\label{tab:samples}
\end{table}

\section{Poisson Equation: 2D Domain \texorpdfstring{$\rightarrow$}{->} Euclidean \texorpdfstring{$+$}{+} Euclidean}
\label{appx:2D_DD}

\subsubsection*{Interface Continuity}

\begin{figure}[H]
\centering
{
\centering
\includegraphics[width=0.75\linewidth, trim={0mm 5mm 0mm 15mm}, clip]{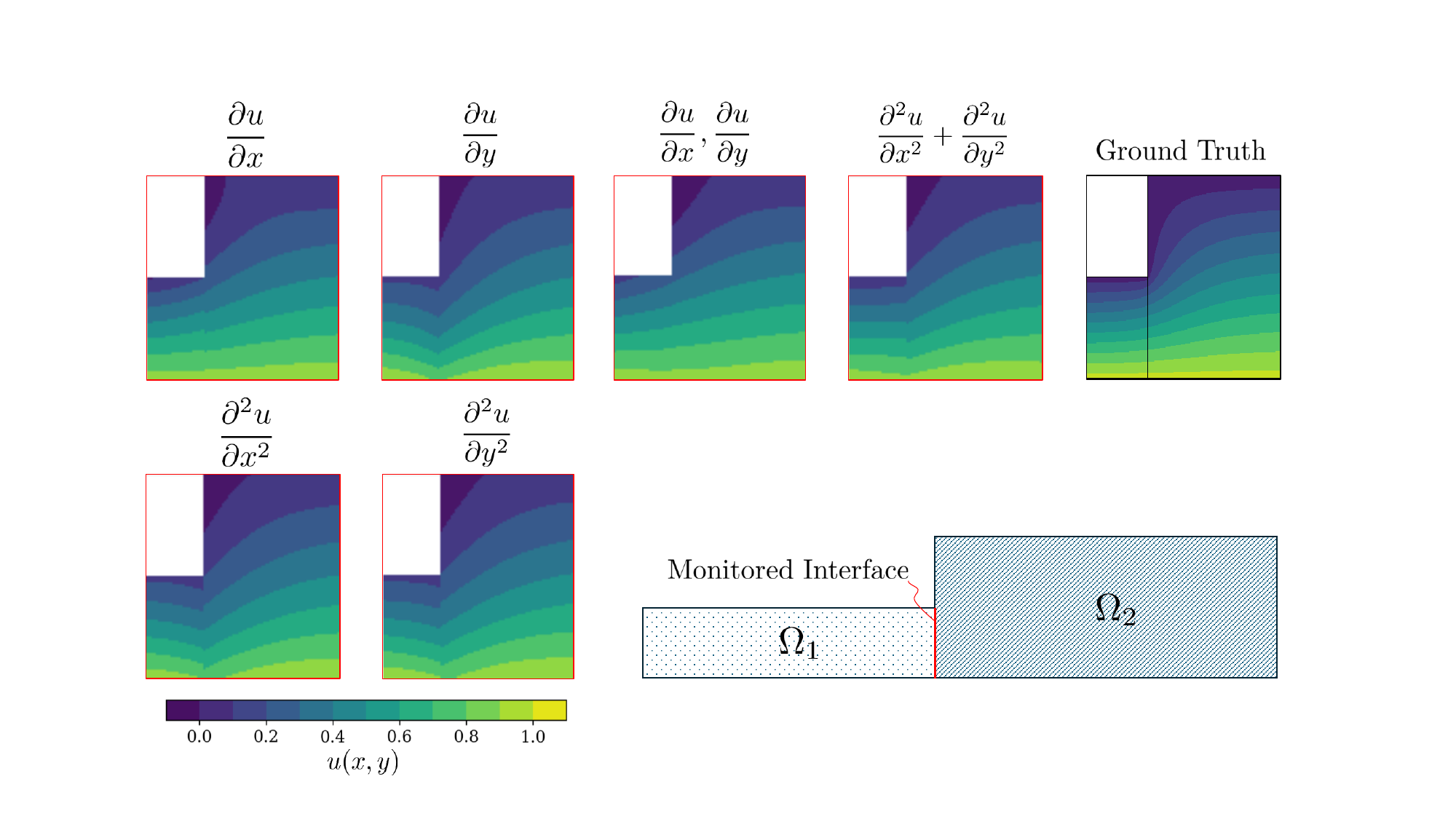}}
\caption{\textbf{Non-overlapping interface.} Comparison of continuity across the interface of subdomains for 2D domain decomposition using different derivative combinations in the interface loss.}
\label{fig:2D_DD_nonoverlapping_interface}
\end{figure}

Here, we evaluate the different combinations of the gradients that can potentially ensure smooth and continuous gradients at the vertical interface between the two subdomains for a 2D domain decomposition problem. Specifically we apply the first and second order gradients along with their combinations to the interface loss function to show the optimal setup for the interface loss function.

\begin{table}[H]
\centering
\begin{tabular}{|c|c|c|c||}
\hline
\multirow{2}{*}{\textbf{Domain Decomposition}} & \multicolumn{2}{c|}{\textbf{Computational Time (ms/iter)}} \\ \cline{2-3}
& \textbf{Non-overlapping} & \textbf{Overlapping} \\
\hline\hline
PINN-PINN & 46.66 & 54.33 \\
\hline
SPINN-SPINN & 3.50 & 3.72 \\
\hline\hline
\end{tabular}
\caption{\textbf{Computational cost of SPINN-SPINN vs.\ PINN-PINN domain decomposition (2D Poisson equation).} Wall-clock training time per iteration (ms/iter), for non-overlapping and overlapping vertical-interface decomposition, with IDENTICAL collocation and boundary point counts across both architectures (verified; see Section~\ref{sec:2D_DD_2}). SPINN-SPINN is $\sim$13$\times$ faster than PINN-PINN in both cases. Introducing the overlapping interface increases cost by 6\% (SPINN-SPINN) and 16\% (PINN-PINN) relative to the non-overlapping baseline.}
\label{tab:2D_DD_Timing}
\end{table}

\begin{figure}[H]
\centering
{
\centering
\includegraphics[width=0.75\linewidth, trim={0mm 5mm 0mm 15mm}, clip]{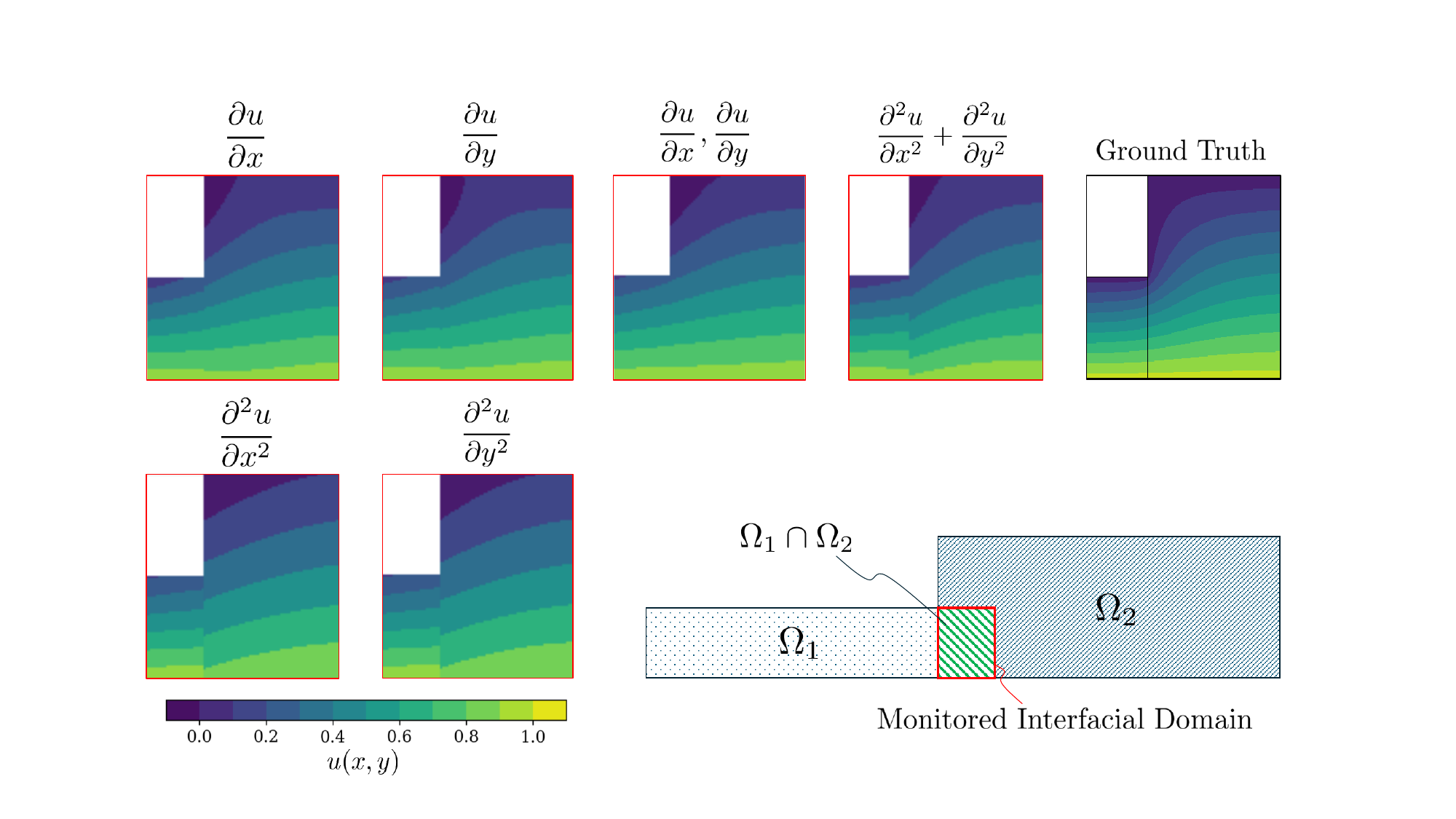}}
\caption{\textbf{Overlapping interfacial domain.} Comparison of continuity across the interface of subdomains for 2D domain decomposition using different derivative combinations in the interface loss.}
\label{fig:2D_DD_overlapping_interface}
\end{figure}

\noindent
We particularity compare for a non-overlapping interface (shown in Figure \ref{fig:2D_DD_nonoverlapping_interface}) and an overlapping interfacial domain (shown in Figure \ref{fig:2D_DD_overlapping_interface}) to show that the first order gradients of the solution field included in the interface loss term attains the best smoothness and continuity across the interface among all the different combinations considered.

\section{PINN Failure: Choosing the Correct Interface}
\label{appx:2D_DD_Failure}

\begin{figure}[H]
\centering
{
\centering
\includegraphics[width=0.8\linewidth, trim={30mm 35mm 30mm 42mm}, clip]{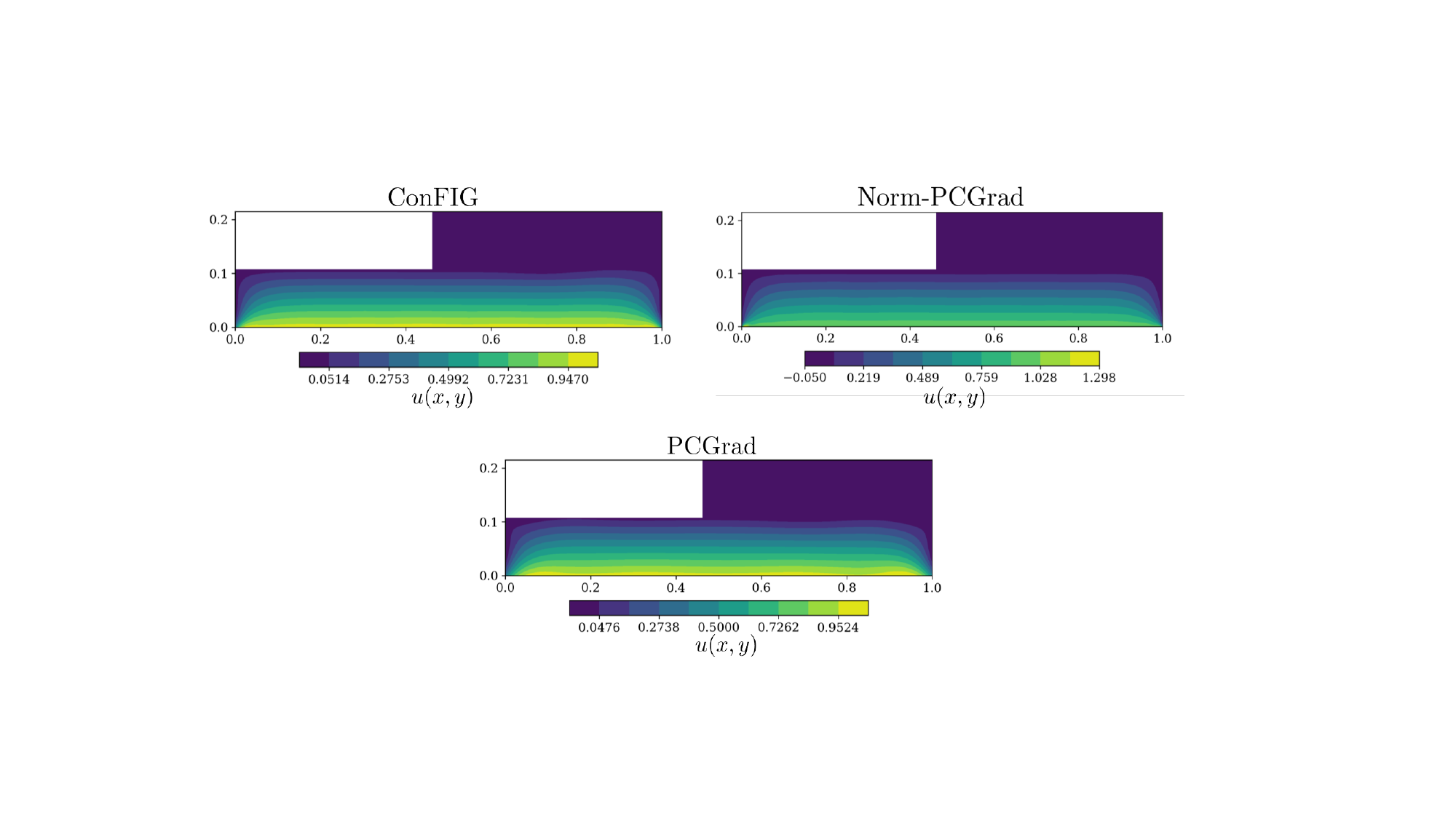}}
\caption{\textbf{2D Poisson (Horizontal Interface) -- Interface failure.} Comparison between solutions from ConFIG, PCGrad and Norm-PCGrad for horizontal interface location.}
\label{fig:2D_DD_failure_horizontal_interface_gs}
\end{figure}

In reference to the discussion in Section \ref{sec:2D_DD_2} on the PINN failure mode for domain decomposition, we note that this failure is independent of any gradient surgery utilized for resolving potential gradient conflicts, as clearly illustrated in Figure \ref{fig:2D_DD_failure_horizontal_interface_gs}.

For a horizontal interface separating the two subdomains, one of the subdomains attains a trivial solution of \textit{zero} across its entirety, since the information from the bottom (high-potential) boundary condition is not transmitted through the interface.

\section{SSBroyden: General Broyden Family Derivation}
\label{appx:ssbroyden}
The general formulation of quasi-Newton methods in the Broyden family can be characterized as follows:
\begin{equation}
\label{eqn:general_quasi}
    \mathrm{B}_{k+1} = \mathrm{B}_{k} - \frac{\mathrm{B}_{k}s_k s_k^\intercal \mathrm{B}_{k}}{s_k^\intercal \mathrm{B}_{k}s_k} + \frac{y_k y_k^\intercal}{y_k^\intercal s_k} + \theta_k (s_k^\intercal \mathrm{B}_k s_k)w_k w_k^\intercal
\end{equation}
Here, $\theta_k$ is a scalar parameter and $\mathrm{H}_k = \mathrm{B}_k^{-1}$ is computed using the gradient information from previous iterations to approximate the inverse Hessian, $\mathrm{H}_k \approx (\nabla^2f_k)^{-1}$, evaluated at $x_k$ in a loss landscape denoted by $f$. Furthermore, $s_k = x_{k+1} - x_k$ represents the change in parameters, while $y_k = \nabla f_{k+1} - \nabla f_{k}$ represents the change in their respective gradients. Once a suitable direction for the update is determined as,
\begin{equation}
    \mathrm{p}_{k} = -\mathrm{H}_k \nabla f_k,
\end{equation}
the next step in the loss landscape is taken as follows,
\begin{equation}
    x_{k+1} = x_k - \alpha_kp_k.
\end{equation}
Here, $\alpha_k$ is computed via a line-search strategy. It is also worthwhile to note that fixing $\theta_k = 0$ recovers the update rule for the BFGS optimizer. Since $\mathrm{H}_k = \mathrm{B}_k^{-1}$, it is infeasible to compute the matrix inverse at every iteration during training. The inverse of Equation \eqref{eqn:general_quasi} is therefore computed to obtain the following update rule,
\begin{equation}
\label{eqn:general_quasi_2}
    \mathrm{H}_{k+1} = \mathrm{H}_{k} - \frac{\mathrm{H}_{k}y_k y_k^\intercal \mathrm{H}_{k}}{y_k^\intercal \mathrm{H}_{k}y_k} + \frac{s_k s_k^\intercal}{y_k^\intercal s_k} + \phi_k (y_k^\intercal \mathrm{H}_k y_k)v_k v_k^\intercal,
\end{equation}
where the parameters are defined as,
\begin{align*}
    v_k &= \frac{s_k}{y_k^\intercal s_k} - \frac{\mathrm{H}_k y_k}{y_k^\intercal \mathrm{H}_k y_k}, \\
    \phi_k &= \frac{1 - \theta_k}{1 + (h_k b_k-1)\theta_k}, \\ 
    b_k &= \frac{s_k^\intercal \mathrm{B}_k s_k}{y_k^\intercal s_k},\\
    h_k &= \frac{y_k^\intercal \mathrm{H}_k y_k}{y_k^\intercal s_k}.
\end{align*}

\bibliographystyle{elsarticle-num} 
\bibliography{reference}
\end{document}